\documentclass[11pt]{article}
\usepackage{booktabs}
\usepackage[final]{acl}
 \usepackage{multirow}
\usepackage{times}
\usepackage{latexsym}

\usepackage[T1]{fontenc}
\usepackage[utf8]{inputenc}

\usepackage{microtype}

\usepackage{inconsolata}
\usepackage{arydshln}

\usepackage{tikz}
\usepackage{pgfplots}
\usepackage{graphicx}
\usepackage{colortbl,xcolor}
\usepackage{enumitem}
\definecolor{lightblue}{RGB}{220,235,250}
\definecolor{skyblue}{RGB}{135,206,235}

\definecolor{poscolor}{RGB}{0, 128, 0}    % 纯深绿
\definecolor{negcolor}{gray}{0.3}         %  深灰 (下降/持平)，几乎接近黑，保证可读

\newcommand{\res}[3]{%
  #1~%
  \ifx+#2%
    \textcolor{poscolor}{\fontsize{8pt}{8pt}\selectfont\textbf{(#2#3)}}%
  \else
    \textcolor{negcolor}{\fontsize{8pt}{8pt}\selectfont(#2#3)}%
  \fi
}

\title{Culture-Aware Machine Translation in Large Language Models: \\Benchmarking and Investigation}

\author{
\textbf{Zekun Yuan\textsuperscript{1}\thanks{These authors contributed equally to this work.}},
\textbf{Yangfan Ye\textsuperscript{1}\footnotemark[1]},
\textbf{Xiaocheng Feng\textsuperscript{1,2}}\thanks{Corresponding author.},
\textbf{Baohang Li\textsuperscript{1}},\\
\textbf{Qichen Hong\textsuperscript{3},}
\textbf{Yunfei Lu\textsuperscript{3},}
\textbf{Dandan Tu\textsuperscript{3},}
\textbf{Bing Qin\textsuperscript{1,2}}
\\
\textsuperscript{1}Harbin Institute of Technology \\
\textsuperscript{2}Peng Cheng Laboratory \\
\textsuperscript{3}Huawei Technologies Co., Ltd \\
\small{
\texttt{\{zkyuan,yfye,xcfeng,qinb\}@ir.hit.edu.cn}
}
}
\begin{document}
\maketitle
\begin{abstract}
Large language models (LLMs) have achieved strong performance in general machine translation, yet their ability in culture-aware scenarios remains poorly understood. 
To bridge this gap, we introduce \textbf{CanMT}, a \textbf{C}ulture-\textbf{A}ware \textbf{N}ovel-Driven Parallel Dataset for \textbf{M}achine \textbf{T}ranslation, together with a theoretically grounded, multi-dimensional evaluation framework for assessing cultural translation quality.
Leveraging \textbf{CanMT}, we systematically evaluate a wide range of LLMs and translation systems under different translation strategy constraints.
Our findings reveal substantial performance disparities across models and demonstrate that translation strategies exert a systematic influence on model behavior. Further analysis shows that translation difficulty varies across types of culture-specific items, and that a persistent gap remains between models’ recognition of culture-specific knowledge and their ability to correctly operationalize it in translation outputs.
In addition, incorporating reference translations is shown to substantially improve evaluation reliability in \textit{LLM-as-a-judge}, underscoring their essential role in assessing culture-aware translation quality. The corpus and code are available at 
\href{https://github.com/zkyuan-scir/CanMT-a-Culture-Aware-Novel-Driven-Parallel-Dataset-for-Machine-Translation}{CanMT}.

\end{abstract}

%%% yangfan %%%
% 1. cultural translation 这个词一听就感觉捞了一个档次
% 2. citep 和 citet
% 3. Intro第一段，不提翻译了，就强调文化的重要性（字数不用多）
% 4. Intro第二段，“翻译”是在多语言和全球化场景下XXX的重要工具和途径。然后衔接下来“普通翻译”在文化场景中的缺陷，现有工作做出了怎样的努力，但大体上还有什么不足？
% 5. Intro第三段，抛出你的工作，需要体现出你要做的这个任务的意义（和普通翻译的区分最核心在哪里？& 创新点在哪里？）
% 6. Intro第四段，你的核心发现
%%%
\section{Introduction}
\begin{table*}[t]
\centering
\small
\renewcommand{\arraystretch}{1.1} % 设置行间距为默认的 1.5 倍
\resizebox{\linewidth}{!}{
\begin{tabular}{llccc}
\toprule
\textbf{Benchmark} & \textbf{Evaluation Focus} & \textbf{Data Source} & \textbf{\# Translation Directions}  \\
\midrule
CulturalRecipes~\citep{cao2023culturaladaptationrecipes}
& Cultural Adaptation in Recipes
& Cooking Recipes
& 2
 \\

MAPS~\citep{wang2025proverbsrunpairsevaluating}
& Proverb Translation (dialogue context)
& Proverbs 
& 8
\\

PoetMT~\citep{chen-etal-2025-benchmarking-llms} 
& Poetry Translation
& Poems
& 1
 \\

DITING~\citep{zhang2025diting}
& Novel Translation 
& Novels
& 1
 \\

% DiscoX~\citep{zhao2025discox}
% & Expert Domains Tranlsation
% & 2
% & Expert \\

CAMT~\citep{yao-etal-2024-benchmarking}
& Cultural-Specific Items (word / phrase-level)
& Wikipedia
& 12
  \\

\midrule
\textbf{CanMT} (Ours)
& Sentences in Diverse Cultural Scenarios
& Novels
& 12
 \\
\bottomrule
\end{tabular}
}
\caption{Comparison of related translation benchmarks.}
\label{tab:cultural-benchmark}
\end{table*}
% Semantic Adequacy \& Text Quality \& Cultural Appropriateness
% 第一段，MT切入
In the era of globalization, Machine Translation (MT) has become a cornerstone technology for cross-lingual communication and global information exchange~\citep{bahdanau2014neural,ye-etal-2024-globesumm,DBLP:journals/corr/abs-2504-01919}. 
Recent advances in large language models (LLMs)~\citep{achiam2023gpt} have further reshaped the MT landscape, positioning LLM-based translation as an increasingly dominant paradigm~\citep{Hendy2023HowGA,jiao2023chatgptgoodtranslatoryes,zhu-etal-2024-multilingual,huang-etal-2024-aligning}.

Despite the impressive progress in general-purpose translation, existing studies on LLM-based MT focus on literal-level translation quality, such as adequacy and fluency. However, language is deeply intertwined with culture, reflecting underlying cultural values, beliefs and social norms. Effective translation therefore often requires more than lexical or syntactic equivalence.

%%% 引用
Prior work has advanced cultural translation evaluation, but most approaches focus on narrowly defined domains such as cooking recipes~\citep{cao2023culturaladaptationrecipes,hu-etal-2024-bridging,zhang-etal-2024-cultural}, proverbs~\citep{wang2025proverbsrunpairsevaluating}, idioms~\citep{li2023translatemeaningsjustwords} and poetry~\citep{chen-etal-2025-benchmarking-llms}. In these datasets, cultural cues are concentrated in a small set of genre-specific features, limiting the evaluation of models in broader, unstructured contexts. \citet{yao-etal-2024-benchmarking} construct parallel corpora from Wikipedia, but the resulting data are largely informational with homogeneous pragmatic and syntactic patterns, limiting the evaluation of models in more diverse, naturalistic contexts. More recently, \citet{zhang2025diting} introduced parallel data from bilingual web novels; however, their corpus is limited to Zh$\rightarrow$En, which prevents a thorough evaluation of cultural transfer across cultures (Table~\ref{tab:cultural-benchmark}).

To bridge this gap, we propose a \textbf{C}ulture-\textbf{A}ware \textbf{N}ovel-Driven Parallel Dataset for \textbf{M}achine \textbf{T}ranslation (\textbf{CanMT}), a parallel corpus constructed from a diverse set of literary novels and their high-quality professional translations. For evaluation, we define assessment dimensions grounded in established translation studies theories, capturing translation quality from multiple perspectives, including contextual accuracy ~\citep{halliday1978language}, cultural adaptation ~\citep{venuti2008translator}, functional equivalence ~\citep{nida1964toward}, fidelity~\citep{newmark1988textbook}, and naturalness~\citep{newmark1988textbook}.

Building on \textbf{CanMT}, we conduct a series of experiments to systematically study culture-aware translation. First, we evaluate and compare the translation performance of a range of representative models and translation systems. Second, we investigate how different translation strategies (Communicative Translation and Semantic Translation) affect translation quality~\citep{newmark1981approaches}.
Our results indicate that culture-aware translation performance generally improves with model scaling up. Furthermore, ``Test-time Scaling Reasoning'' enables consistent and incremental gains. In terms of translation strategies, communicative strategy tends to enhance fluency and functional adequacy, whereas semantic strategy prioritizes accurate meaning transfer during translation.

In addition to the main experiments, we conducted several analyses. First, regarding strategy preference, similarity analysis reveals a systemic bias toward semantic translation in default translation settings, suggesting that LLMs tend to adopt a translation strategy that is closer to a semantic-oriented approach; Second, across culture-specific items (CSIs)~\citep{newmark1988textbook} categories, we identify a clear difficulty hierarchy: models excel at geographic and ecological terms but struggle significantly with nuanced linguistic-symbolic items; Third, by probing the relationship between knowledge and performance, we find that while correct cultural knowledge generally aids translation, a persistent "knowledge-application gap" remains; Finally, we demonstrate that reference translations are vital for calibrating automatic evaluators, as their inclusion consistently improves alignment with human judgment across cultural dimensions.

Overall, this work introduces the \textbf{CanMT} benchmark for investigating culture-aware translation in LLMs, enabling a systematic evaluation of model behavior in culturally grounded translation settings. Our empirical analysis provides detailed insights into models’ behavior in culture-aware translation, and establishes a reliable reference point for future research on cultural adaptability.

\section{Related Works}

\paragraph{Culture-aware Machine Translation.}
Despite the rapid progress of LLMs in general multilingual capabilities~\citep{qin2024multilingual,ye-etal-2025-cc,ye2025langgps}, their ability to accurately convey culture-specific meanings remains insufficiently understood. Culture-aware machine translation aims to preserve culturally grounded meanings across languages. Existing work can be broadly grouped into two main categories. The first focuses on specific culturally relevant linguistic phenomena, such as recipes~\citep{cao2023culturaladaptationrecipes,hu-etal-2024-bridging,zhang-etal-2024-cultural}, proverbs~\citep{wang2025proverbsrunpairsevaluating}, idioms~\citep{li2023translatemeaningsjustwords}, novels~\citep{zhang2025diting}, and poetry~\citep{chen-etal-2025-benchmarking-llms}. The second utilizes parallel corpora enriched with cultural information to construct benchmarks~\citep{yao-etal-2024-benchmarking,singh-etal-2024-translating} and \citet{ye2024exploring} recently proposed $\mathcal{X}$Transplant for cross-lingual complementarity in culture scenarios. Among them, \citet{yao-etal-2024-benchmarking} introduced the CAMT dataset, which is primarily derived from Wikipedia. While Wikipedia provides well-aligned parallel data, its predominantly factual content offers limited coverage of the narrative structures and stylistic variation characteristic of literary texts. In contrast, \textbf{CanMT} leverages novels originating from the target culture, encompassing richer cultural expressions and more diverse linguistic styles, and thereby enabling a more comprehensive and systematic evaluation.

\begin{figure*}[t]
  \centering
  \includegraphics[width=0.95\textwidth]{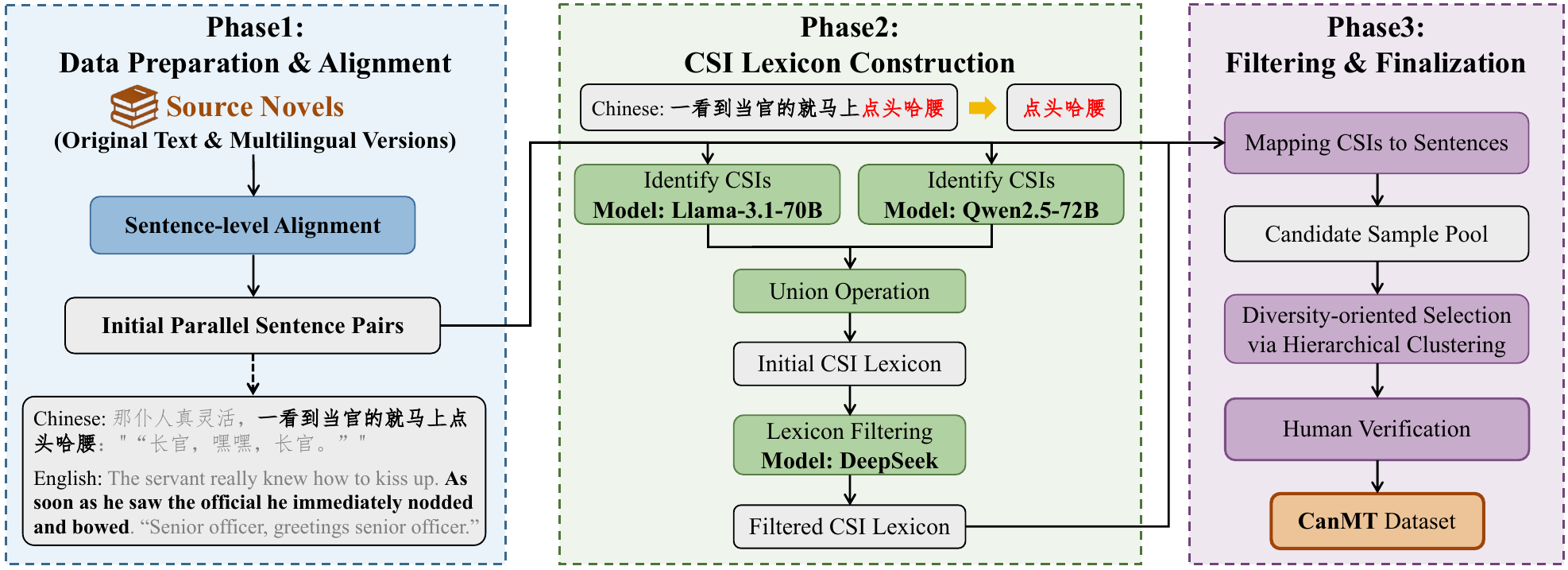}
  \caption{Overview of the \textbf{CanMT} dataset construction pipeline, including data preparation, sentence alignment, CSI-based sentence selection, and diversity-oriented  sample selection via clustering and human verification.}
  \label{fig:dataprocess}
\end{figure*}

\paragraph{LLM-as-a-Judge.}
Due to the complexity and diversity of cultural translation, we employ an LLM-based approach to evaluate models’ translation across multiple dimensions. Recently, the use of large language models for machine translation evaluation has gained increasing popularity. \citet{kocmi-federmann-2023-large} proposed GEMBA, demonstrating the potential of GPT-4 in assessing MT quality. Building on this, \citet{kocmi-federmann-2023-gemba} combined GEMBA with the MQM evaluation framework, using error span detection to systematically evaluate MT outputs. \citet{feng-etal-2025-mad} leverage a multi-agent debate mechanism to assess translation quality from multiple dimensions using error detection strategies. Our approach adopts a simpler, single-judge, multi-dimensional evaluation setup that is cost-efficient and more controlled: we provide the LLM with a detailed scoring rubric in the prompt,  which allows a more focused analysis of evaluation dimensions and translation objectives in cultural translation.

\begin{figure}[t]
  \centering
  \includegraphics[width=0.45\textwidth]{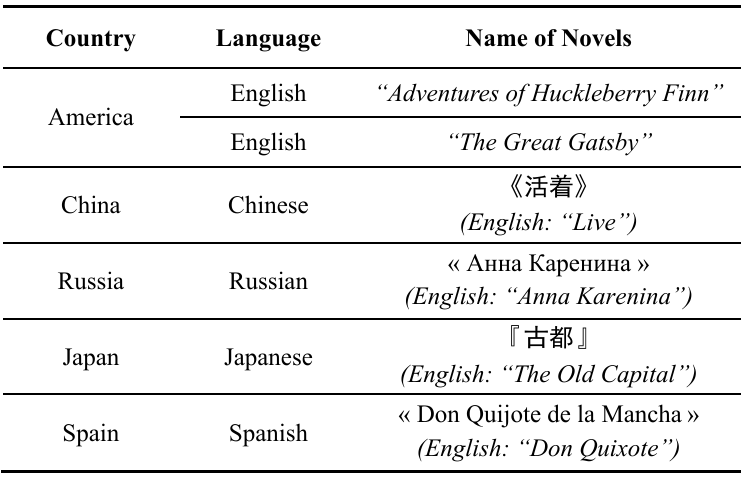}
  \caption{Representative novels from diverse cultural backgrounds used as sources for building \textbf{CanMT}.}
  \label{fig:novels}
\end{figure}

\section{Benchmark Construction}

\subsection{\textbf{CanMT} Dataset}

We construct a parallel corpus based on literary novels, which exhibit intrinsic cultural specificity often absent in general-domain corpora. To ensure cultural representativeness, we select classic novels from different countries that have multiple translations, facilitating the extraction of parallel sentences. The novels and their corresponding language are listed in Figure~\ref{fig:novels}.

During data processing, we first perform sentence-level alignment on each chapter using the Vecalign tool~\citep{thompson2019vecalign} to obtain an initial set of parallel sentence pairs. From these aligned pairs, we employ LLaMA-3.1-70B-Instruct~\citep{dubey2024llama} and Qwen2.5-72B-Instruct~\citep{qwen_qwen25_2025} to identify CSIs, taking the union of their outputs to construct an initial CSI lexicon. To ensure its reliability, the lexicon is filtered by DeepSeek~\citep{liu2024deepseek}.

\begin{figure*}[t]
  \centering
  \includegraphics[width=\textwidth]{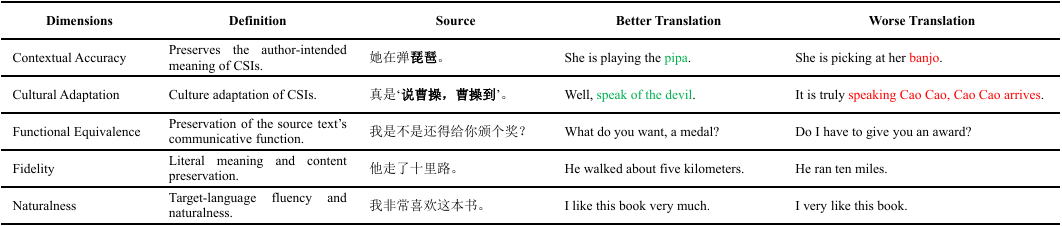}
  \caption{Overview of the evaluation dimensions adopted in this study, including their definitions, illustrative sources, and representative examples of better and worse translations for each dimension.}
  \label{fig:dim_intro}
\end{figure*}

To ensure cultural diversity, we first collect all sentences containing each CSI and limit the candidate pool to at most five sentences per CSI. We then encode these sentences into embeddings using LaBSE~\citep{feng2022language} and apply hierarchical clustering on the embeddings to identify representative sentences, selecting 200 in total from the candidate pool. Finally, human annotators review these selected sentences to remove non-parallel or low-quality translations, resulting in the final dataset. The dataset statistics are summarized in Table~\ref{tab:direction_stats}, with more detailed statistics and the manual filtering procedure provided in Appendix~\ref{app:dataset}.
% \begin{table}[t]
% \centering
% \tiny
% \begin{tabular}{lcccccccccccc}
% \toprule
% \textbf{Direction} 
% & en$\rightarrow$es 
% & en$\rightarrow$ja 
% & en$\rightarrow$zh 
% & es$\rightarrow$en 
% & es$\rightarrow$zh 
% & ja$\rightarrow$en 
% & ja$\rightarrow$zh 
% & ru$\rightarrow$zh 
% & zh$\rightarrow$en 
% & zh$\rightarrow$es 
% & zh$\rightarrow$ja 
% & zh$\rightarrow$ru \\
% \midrule
% \textbf{samples} 
% & 106 & 103 & 111 & 134 & 166 & 116 & 106 & 125 & 125 & 146 & 137 & 95 \\
% \textbf{CSIs Counts} 
% & 143 & 138 & 148 & 300 & 368 & 255 & 221 & 226 & 228 & 272 & 258 & 168 \\
% \bottomrule

\begin{table}[t]
\centering
% \tiny
\resizebox{\columnwidth}{!}{
\begin{tabular}{lcccccc}
\toprule
\textbf{Direction} 
& En$\rightarrow$Es 
& En$\rightarrow$Ja 
& En$\rightarrow$Zh 
& Zh$\rightarrow$Ja 
& Zh$\rightarrow$Ru 
& Zh$\rightarrow$Es \\

\midrule
\textbf{\# Samples} 
& 106 & 103 & 111 & 137 & 95 & 146 \\

\textbf{\# CSIs} 
& 143 & 138 & 148 & 258 & 168 & 272 \\

\toprule
\textbf{Direction} 
& Es$\rightarrow$En 
& Ja$\rightarrow$En 
& Zh$\rightarrow$En 
& Ja$\rightarrow$Zh 
& Ru$\rightarrow$Zh 
& Es$\rightarrow$Zh \\
\midrule
\textbf{\# Samples}  & 134 & 116 & 125 & 106 & 125 & 166 \\
\textbf{\# CSIs} & 300 & 255 & 228 & 221 & 226 & 368 \\

\bottomrule
\end{tabular}
}
\caption{Statistics of the \textbf{CanMT} across translation directions, reporting the number of samples and CSIs.}  
\label{tab:direction_stats}
\end{table}

% To guarantee cultural diversity, we map each CSI to the set of sentences containing it and retain at most five candidate sentences per CSI. Sentence embeddings are then computed using LaBSE~\citep{feng2022language}, followed by hierarchical clustering to select 200 representative sentences from the candidate pool. Finally, human annotators review the selected sentences to remove non-parallel or low-quality translations, producing the final dataset.

\subsection{Evaluation via Multi-Dimensions}
\label{sec:Evaluation}

To systematically evaluate culture-aware translation, we propose a multi-dimensional assessment framework grounded in classical translation theories. Our framework measures translation quality along five dimensions: contextual accuracy~\citep{halliday1978language}, cultural adaptation~\citep{venuti2008translator}, functional equivalence~\citep{nida1964toward}, fidelity~\citep{newmark1988textbook}, and naturalness~\citep{newmark1988textbook}. The definitions of these evaluation dimensions are provided in Figure~\ref{fig:dim_intro}.

For each dimension, we employ a 7-point Likert scale. Detailed scoring rubrics for all dimensions are provided in Appendix~\ref{app:scoring_rubrics}. The overall translation score is calculated as the arithmetic mean of the individual dimension scores, allowing for a unified yet fine-grained assessment.

\begin{equation}
S = \frac{1}{|D|} \sum_{d}^{D} s_{d}
\end{equation}

where $D$ denotes the total number of evaluation dimensions, and $s_{d}$ represents the score assigned to the $d$-th dimension.

\begin{table}[t]
\centering
\small
\setlength{\tabcolsep}{6pt}
\begin{tabular}{lcc}
\toprule
\textbf{Dimension} & \textbf{H--H $\tau$} & \textbf{M--H $\tau$} \\
\midrule
Contextual Accuracy    & 0.4536 & 0.4455 \\
Cultural Adaptation    & 0.4202 & 0.3891 \\
Functional Equivalence & 0.4416  & 0.4503 \\
Fidelity               & 0.4937  & 0.4625 \\
Naturalness            & 0.4168  & 0.4716 \\
\bottomrule
\end{tabular}
\caption{Evaluation consistency across dimensions. H--H $\tau$ denotes inter-annotator agreement measured by Kendall’s $\tau$, while M--H $\tau$ measures the rank correlation between machine predictions and human judgments.}
\label{tab:consistency}
\end{table}

For scalability and consistency, we employ GPT-5-nano as the automatic evaluator to assign scores for each dimension, with the full evaluation prompts provided in Appendix~\ref{app:eval_prompts}. To validate the scoring procedure, we conduct a controlled human evaluation and report both inter-annotator agreement and model–human rank correlation.

Specifically, for each translation direction, we select 100 translation pairs and recruit two professional bilingual annotators to independently score across all evaluation dimensions. To quantify human–human agreement, we compute Kendall’s $\tau$
between the two sets of human scores. For model–human comparison, we retain only those instances for which the absolute difference between the two human scores does not exceed 2. For the retained instances, we use the averaged human score to compute Kendall’s $\tau$
 between GPT’s predictions and human judgments. Table~\ref{tab:consistency} reports evaluation consistency across different dimensions. To further assess the robustness of our evaluation, we examine cross-model consistency across multiple LLM judges as well as repeated scoring stability. The detailed results are reported in Appendix~\ref{app:judge_robustness}.

\section{Experimental Setup}

To fully evaluate the effectiveness of MT systems on culture-aware translation and the influence of different translation paradigms on model outputs, we compare multiple MT systems under our evaluation framework (§~\ref{sec:model-comparison}) and investigate how explicit semantic and communicative translation constraints affect their translation behavior (§~\ref{sec:translation-strategies}).

\subsection{Models \& Systems}
\label{sec:model-comparison}

Our experiments cover a wide range of models or systems:

\begin{itemize}[leftmargin=*]
\setlength{\parsep}{0pt}
\setlength{\parskip}{0pt}
    \item \textbf{Open-source LLMs}: We evaluate a set of representative open-source large language models, including the LLaMA3 ~\citep{dubey2024llama}, Qwen2.5~\citep{qwen_qwen25_2025}, Qwen3~\citep{yang2025qwen3}, and Mixtral~\citep{Jiang2024MixtralOE}. In addition, we consider more recent models, namely DeepSeek-V3.2~\citep{liu2025deepseek}, DeepSeek-R1~\citep{deepseekai2025deepseekr1incentivizingreasoningcapability}.
    
    \item \textbf{Proprietary Models}: We evaluate a set of widely used proprietary models, including GPT-4o, GPT-4~\citep{achiam2023gpt}, Gemini-2.5-Flash-Lite~\citep{team2023gemini}, and Grok-4.1.

    \item \textbf{Specialized MT Models}: We evaluate dedicated machine translation systems, including NLLB-200~\citep{costa2022no}, Seed-X~\citep{cheng2025seedxbuildingstrongmultilingual} and LLaMAX3~\citep{lu-etal-2024-llamax} which are specifically designed for translation. 
    
    \item \textbf{Production Systems}: We further incorporate widely deployed industrial translation engines, namely Google Translate and Youdao Translate, as real-world production-level baselines.
\end{itemize}

All open-source models and specialized MT models, except for the DeepSeek series, are decoded using greedy decoding. Detailed decoding settings and translation prompts for all systems are provided in Appendix~\ref{app:Experimental Setting}.

\begin{table*}[t]
\centering
\small
\resizebox{\textwidth}{!}{
\begin{tabular}{lccccccccccccc}
\toprule
\textbf{Model} & \textbf{En$\rightarrow$Es} & \textbf{En$\rightarrow$Ja} & \textbf{En$\rightarrow$Zh} & \textbf{Es$\rightarrow$En} & \textbf{Es$\rightarrow$Zh} & \textbf{Ja$\rightarrow$En} & \textbf{Ja$\rightarrow$Zh} & \textbf{Ru$\rightarrow$Zh} & \textbf{Zh$\rightarrow$En} & \textbf{Zh$\rightarrow$Es} & \textbf{Zh$\rightarrow$Ja} & \textbf{Zh$\rightarrow$Ru} & \textbf{Avg} \\

\midrule
\rowcolor{gray!20}  % 20% 灰色背景
\multicolumn{14}{c}{\textbf{Proprietary LLMs}} \\

GPT-4 & 4.88 & 5.16 & 5.33 & 5.24 & 4.91 & 5.18 & 5.23 & 4.79 & 5.52 & 5.21 & 5.34 & 4.88 & 5.14 \\
GPT-4o & 4.74 & 4.99 & 5.21 & 5.07 & 4.66 & 5.03 & 4.93 & 4.75 & 5.35 & 4.92 & 5.11 & 4.75 & 4.96 \\
Gemini-2.5-Flash-Lite & 4.97 & 5.08 & 5.11 & 5.05 & 4.63 & 4.96 & 4.72 & 4.84 & 5.40 & 5.16 & 5.07 & 4.97 & 5.00 \\
Grok-4.1 & 4.99 & 5.34 & 5.22 & 5.22 & 5.03 & 5.04 & 5.17 & 5.00 & 5.43 & 5.25 & 5.17 & 5.30 & 5.18 \\

\midrule
\rowcolor{gray!20}  % 20% 灰色背景
\multicolumn{14}{c}{\textbf{Open-source LLMs}} \\

LLaMA-3-8B-Instruct-262k     & 4.40 & 3.36 & 3.89 & 4.19 & 3.20 & 3.41 & 3.29 & 3.50 & 4.33 & 3.81 & 3.38 & 3.32 & 3.67 \\
LLaMA-3.3-70B-Instruct       & 4.87 & 4.83 & 5.00 & 5.04 & 4.57 & 4.44 & 4.43 & 4.61 & 5.14 & 4.80 & 4.81 & 4.80 & 4.78 \\

Mixtral-8x7B-Instruct-v0.1   & 4.47 & 3.00 & 3.15 & 4.81 & 2.90 & 3.51 & 3.15 & 2.96 & 4.68 & 4.19 & 3.12 & 3.73 & 3.64 \\

Qwen2.5-7B-Instruct          & 4.13 & 3.52 & 4.66 & 4.61 & 4.04 & 3.82 & 4.28 & 4.31 & 4.97 & 3.93 & 3.65 & 3.30 & 4.10 \\
Qwen2.5-14B-Instruct         & 4.46 & 4.08 & 5.00 & 4.89 & 4.50 & 4.32 & 4.66 & 4.59 & 5.26 & 4.53 & 4.19 & 3.66 & 4.51 \\
Qwen2.5-32B-Instruct         & 4.71 & 4.40 & 5.06 & 4.99 & 4.64 & 4.60 & 4.82 & 4.69 & 5.22 & 4.74 & 4.35 & 4.05 & 4.69 \\
Qwen2.5-72B-Instruct         & 4.89 & 4.84 & 5.24 & 5.01 & 4.72 & 4.67 & 4.93 & 4.85 & 5.37 & 5.02 & 4.74 & 4.70 & 4.91 \\

Qwen3-4B          & 4.01 & 3.76 & 4.57 & 4.43 & 4.10 & 3.52 & 4.24 & 4.24 & 4.83 & 3.82 & 3.89 & 3.46 & 4.07 \\
Qwen3-8B      & 4.47 & 4.25 & 4.91 & 4.50 & 4.50 & 4.02 & 4.65 & 4.61 & 5.03 & 4.38 & 4.42 & 4.05 & 4.48 \\
Qwen3-14B         & 4.69 & 4.65 & 5.07 & 4.85 & 4.62 & 4.24 & 4.73 & 4.83 & 5.31 & 4.78 & 4.73 & 4.27 & 4.73 \\
Qwen3-32B     & 4.66 & 4.47 & 5.12 & 4.92 & 4.64 & 4.39 & 4.77 & 4.83 & 5.19 & 4.71 & 4.86 & 4.34 & 4.74 \\

DeepSeek-R1                  & 5.01 & 5.08 & 5.24 & 5.16 & 4.65 & 4.97 & 4.84 & 4.70 & 5.28 & 5.05 & 5.07 & 5.22 & 5.02 \\
DeepSeek-V3.2                & 5.05 & 5.21 & 5.31 & 5.10 & 4.82 & 4.88 & 5.07 & 4.86 & 5.41 & 5.10 & 5.15 & 5.00 & 5.08 \\

\midrule
\rowcolor{gray!20}  % 20% 灰色背景
\multicolumn{14}{c}{\textbf{Specialized MT Models}} \\

Seed-X-PPO-7B  & 4.78 & 4.69 & 5.14 & 4.86 & 4.60 & 4.21 & 4.28 & 4.69 & 5.37 & 5.07 & 4.45 & 4.96 & 4.76 \\
NLLB-200-3.3B                & 4.04 & 3.15 & 3.07 & 3.96 & 2.83 & 2.67 & 2.49 & 3.24 & 3.20 & 2.93 & 2.71 & 2.86 & 3.10 \\
LLaMAX3-8B-Alpaca            & 4.07 & 3.90 & 4.12 & 4.45 & 3.61 & 3.70 & 3.94 & 3.81 & 4.46 & 3.89 & 3.77 & 3.69 & 3.95 \\

\midrule
\rowcolor{gray!20}  % 20% 灰色背景
\multicolumn{14}{c}{\textbf{Production Systems}} \\
Google Translate & 4.45 & 4.64 & 4.70 & 4.54 & 4.06 & 4.35 & 4.02 & 4.54 & 5.00 & 4.61 & 4.33 & 4.72 & 4.50 \\
Youdao Translate & 3.59 & 3.72 & 4.89 & 4.11 & 3.46 & 3.61 & 4.67 & 3.28 & 5.14 & 3.35 & 4.35 & 3.55 & 3.98 \\
\bottomrule
\end{tabular}
}
\caption{Overall translation performance across language directions. For the Qwen3 series and Seed-X models, only the non-reasoning variants are included.}
\label{tab:overall_results}
\end{table*}

\subsection{Translation Strategies}
\label{sec:translation-strategies}

In addition, we explore the impact of translation paradigm on culture-aware translation quality. Inspired by classical translation theory~\citep{newmark1988textbook}, we design two strategy-constrained prompts and evaluate model performance under these paradigms.

\paragraph{Semantic Translation Constraint.} Emphasizing preserving the meaning and core informational content of the source text, discouraging paraphrasing or the introduction of unstated information.

\paragraph{Communicative Translation Constraint.} Emphasizing producing translations that are natural and culturally appropriate for target-language readers, while fulfilling the intended communicative function of the source text.

\section{Results and Analysis}
\label{sec:results}

\subsection{Translation Ability across Models}

\begin{table}[t]
\centering
\small
\begin{tabular}{lccc}
\toprule
% \rowcolor{gray!20}  % 20% 灰色背景
\textbf{Model} & \textbf{w/o think} & \textbf{w/ think} & \\
\midrule
Qwen3-4B  & 4.07 & 4.25 & $\uparrow$0.18 \\
Qwen3-8B  & 4.48 & 4.63 & $\uparrow$0.15 \\
Qwen3-14B & 4.73 & 4.82 & $\uparrow$0.09 \\
Qwen3-32B & 4.74 & 4.89 & $\uparrow$0.15 \\
Seed-X-PPO-7B & 4.76 & 4.85 & $\uparrow$0.09 \\
\bottomrule
\end{tabular}
\caption{A comparative analysis of model performance with and without the "think" mode.}
\label{tab:think_improvement}
\end{table}

Table~\ref{tab:overall_results} presents the overall culture-aware translation performance of all evaluated models across 12 language directions.\footnote{Detailed results across dimensions are reported in Appendix~\ref{app:Dimension-level-result}.}

\paragraph{Model-wise Comparison.} Proprietary LLMs and state-of-the-art open-source models consistently outperform traditional production systems across the benchmark. Notably, the performance gap between top-tier open-source and proprietary models has marginally narrowed. Furthermore, the specialized Seed-X exhibits strong efficiency; despite its smaller parameter footprint, it competitively rivals much larger general-purpose models.

\paragraph{Scaling Effects in Open-source Models.}
Among open-source systems, translation performance exhibits a clear positive correlation with model scale. Taking the Qwen2.5 series as an example, scores increase monotonically from 7B to 14B, 32B, and 72B in nearly all language directions. This trend indicates that larger open-source models consistently achieve stronger culture-aware translation performance, highlighting the benefits of scaling for cross-lingual and cultural tasks.

\paragraph{Effect of ``Test-time Scaling Reasoning''.} As shown in Table~\ref{tab:think_improvement}, the Qwen3 variants and Seed-X exhibit consistent gains from "think" mode, indicating that incorporating reasoning mechanisms can contribute to enhanced culture-aware translation performance across multiple language pairs.

\begin{figure}[t]
 \centering
  \includegraphics[width=\columnwidth]{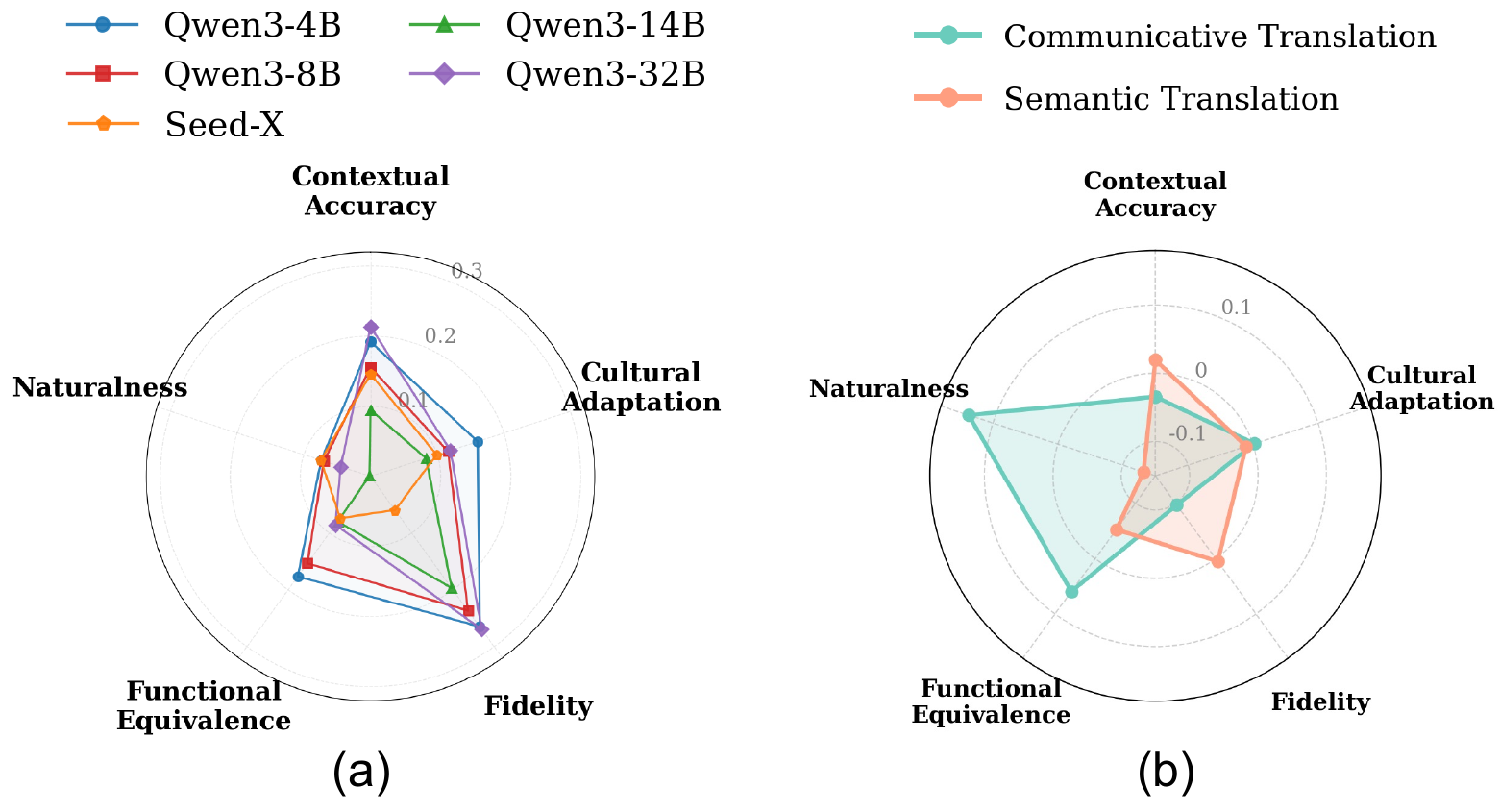}
  \caption{a) Improvement from ``Test-time Scaling Reasoning'' across evaluation dimensions. b) Comparison of strategies across evaluation dimensions. }
  \label{fig:both_rdar}
\end{figure}

To identify the specific drivers of this performance boost, we present a dimension-wise breakdown in Figure~\ref{fig:both_rdar} (a). While reasoning strategies generally yield positive shifts across all metrics, the distribution of gains is non-uniform. Specifically, the \textit{think} models exhibit the most pronounced enhancement in Fidelity, indicating that the reasoning process primarily aids in strictly adhering to the source content's meaning. In contrast, the improvement in Naturalness is the least significant, suggesting that current reasoning mechanisms focus more on semantic transfer than on stylistic fluency.

\subsection{Effect of Translation Strategy Constraints}

\begin{table*}[t]
\centering
% \scriptsize % 如果不用 resizebox，可以开启这个
\setlength{\tabcolsep}{2.5pt} % 极度压缩列间距

\resizebox{\linewidth}{!}{ % 强制适应宽度
\begin{tabular}{lccccc|ccccc|ccccc}
\toprule
% 表头结构
\multirow{2}{*}{\textbf{Model}} & \multicolumn{5}{c}{\textbf{Default Translation}} & \multicolumn{5}{c}{\textbf{Communicative Translation}} & \multicolumn{5}{c}{\textbf{Semantic Translation}} \\
\cmidrule(lr){2-6} \cmidrule(lr){7-11} \cmidrule(lr){12-16}
 & \multicolumn{1}{c}{CTX} & \multicolumn{1}{c}{CAD} & \multicolumn{1}{c}{FEQ} & \multicolumn{1}{c}{FID} & \multicolumn{1}{c}{NAT} 
 & \multicolumn{1}{c}{CTX} & \multicolumn{1}{c}{CAD} & \multicolumn{1}{c}{FEQ} & \multicolumn{1}{c}{FID} & \multicolumn{1}{c}{NAT} 
 & \multicolumn{1}{c}{CTX} & \multicolumn{1}{c}{CAD} & \multicolumn{1}{c}{FEQ} & \multicolumn{1}{c}{FID} & \multicolumn{1}{c}{NAT} \\

\midrule

% --- deepseek-v3.2 ---
DeepSeek-V3.2
& 5.17 & 5.30 & 5.07 & 5.07 & 4.80
& \res{4.95}{-}{.21} & \res{5.20}{-}{.10} & \res{4.95}{-}{.12} & \res{4.72}{-}{.35} & \res{4.71}{-}{.09}
& \res{5.20}{+}{.04} & \res{5.32}{+}{.02} & \res{5.04}{-}{.04} & \res{5.14}{+}{.07} & \res{4.58}{-}{.22} \\

% --- gemini-2.5 ---
Gemini-2.5-Flash-Lite
& 5.04 & 5.22 & 4.97 & 5.08 & 4.68
& \res{5.01}{-}{.03} & \res{5.24}{+}{.02} & \res{5.05}{+}{.08} & \res{5.00}{-}{.09} & \res{4.83}{+}{.15}
& \res{5.10}{+}{.06} & \res{5.20}{-}{.02} & \res{4.83}{-}{.14} & \res{5.07}{-}{.02} & \res{4.43}{-}{.24} \\

% --- gpt-4o ---
GPT-4o
& 5.03 & 5.17 & 4.90 & 5.10 & 4.61
& \res{5.02}{-}{.01} & \res{5.19}{+}{.02} & \res{5.07}{+}{.17} & \res{4.94}{-}{.16} & \res{4.94}{+}{.33}
& \res{5.16}{+}{.13} & \res{5.24}{+}{.07} & \res{5.05}{+}{.05} & \res{5.18}{+}{.09} & \res{4.52}{-}{.09} \\

% --- Llama-3-8B ---
Llama-3-8B-Ins-262k
& 3.61 & 4.14 & 3.60 & 3.51 & 3.50
& \res{3.60}{-}{.01} & \res{4.09}{-}{.04} & \res{3.60}{ }{.00} & \res{3.50}{-}{.02} & \res{3.61}{+}{.11}
& \res{3.61}{ }{.00} & \res{4.15}{+}{.01} & \res{3.53}{-}{.08} & \res{3.47}{-}{.04} & \res{3.44}{-}{.07} \\

% --- Llama-3.3-70B ---
Llama-3.3-70B-Ins
& 4.81 & 5.03 & 4.72 & 4.84 & 4.50
& \res{4.68}{-}{.13} & \res{4.99}{-}{.04} & \res{4.80}{+}{.08} & \res{4.65}{-}{.19} & \res{4.67}{+}{.17}
& \res{4.84}{+}{.03} & \res{4.95}{-}{.07} & \res{4.69}{-}{.03} & \res{4.81}{-}{.03} & \res{4.35}{-}{.15} \\

% --- Mixtral-8x7B ---
Mixtral-8x7B-Ins-v0.1
& 3.71 & 4.12 & 3.50 & 3.48 & 3.38
& \res{3.79}{+}{.08} & \res{4.21}{+}{.09} & \res{3.47}{-}{.03} & \res{3.34}{-}{.14} & \res{3.44}{+}{.06}
& \res{3.67}{-}{.04} & \res{4.08}{-}{.04} & \res{3.33}{-}{.17} & \res{3.38}{-}{.10} & \res{3.19}{-}{.19} \\

% --- Qwen2.5-14B ---
Qwen2.5-14B-Ins
& 4.53 & 4.78 & 4.49 & 4.49 & 4.27
& \res{4.53}{ }{.00} & \res{4.86}{+}{.07} & \res{4.70}{+}{.21} & \res{4.52}{+}{.03} & \res{4.49}{+}{.23}
& \res{4.56}{+}{.02} & \res{4.85}{+}{.07} & \res{4.53}{+}{.04} & \res{4.58}{+}{.10} & \res{4.15}{-}{.12} \\

% --- Qwen2.5-32B ---
Qwen2.5-32B-Ins
& 4.73 & 5.00 & 4.64 & 4.72 & 4.34
& \res{4.63}{-}{.11} & \res{4.93}{-}{.08} & \res{4.75}{+}{.10} & \res{4.55}{-}{.17} & \res{4.58}{+}{.23}
& \res{4.74}{+}{.01} & \res{4.93}{-}{.08} & \res{4.60}{-}{.04} & \res{4.68}{-}{.04} & \res{4.24}{-}{.10} \\

% --- Qwen2.5-72B ---
Qwen2.5-72B-Ins
& 4.95 & 5.12 & 4.89 & 4.97 & 4.64
& \res{4.91}{-}{.05} & \res{5.14}{+}{.02} & \res{4.95}{+}{.06} & \res{4.89}{-}{.07} & \res{4.83}{+}{.19}
& \res{4.99}{+}{.04} & \res{5.11}{-}{.02} & \res{4.87}{-}{.02} & \res{5.04}{+}{.08} & \res{4.57}{-}{.07} \\

% --- Qwen2.5-7B ---
Qwen2.5-7B-Ins
& 4.13 & 4.49 & 4.06 & 4.00 & 3.84
& \res{4.17}{+}{.04} & \res{4.56}{+}{.07} & \res{4.14}{+}{.08} & \res{4.02}{+}{.03} & \res{3.98}{+}{.14}
& \res{4.21}{+}{.08} & \res{4.55}{+}{.06} & \res{4.11}{+}{.05} & \res{4.09}{+}{.10} & \res{3.86}{+}{.02} \\

% --- Qwen3-8B (w/o think) ---
Qwen3-8B(w/o think)
& 4.46 & 4.75 & 4.50 & 4.47 & 4.25
& \res{4.41}{-}{.04} & \res{4.73}{-}{.02} & \res{4.51}{+}{.02} & \res{4.43}{-}{.03} & \res{4.29}{+}{.04}
& \res{4.42}{-}{.04} & \res{4.72}{-}{.03} & \res{4.40}{-}{.10} & \res{4.35}{-}{.12} & \res{4.10}{-}{.15} \\

% --- Qwen3-8B (with think) ---
Qwen3-8B(w/ think)
& 4.61 & 4.87 & 4.65 & 4.70 & 4.32
& \res{4.67}{+}{.06} & \res{4.88}{+}{.01} & \res{4.71}{+}{.06} & \res{4.69}{-}{.01} & \res{4.41}{+}{.10}
& \res{4.51}{-}{.10} & \res{4.78}{-}{.09} & \res{4.49}{-}{.17} & \res{4.67}{-}{.04} & \res{4.11}{-}{.21} \\
\bottomrule
\end{tabular}
}
\caption{Effect of translation strategy constraints across three strategies on five evaluation dimensions: Contextual Accuracy (CTX), Cultural Adaptation (CAD), Functional Equivalence (FEQ), Fidelity (FID), and Naturalness (NAT). Default Translation refers to model-generated translations produced without any explicit strategy constraints. Differences relative to the Default Translation are indicated in parentheses.}

\label{tab:strategy_comparison}
\end{table*}

This section analyzes the impact of different translation strategy constraints on translation quality, by comparing model performance under each strategy across multiple evaluation dimensions.\footnote{Detailed results are reported in Appendix~\ref{app:lang-level-res}.}

\paragraph{Communicative vs. Default Translation.}
As shown in Table~\ref{tab:strategy_comparison}, compared to the default translation, the communicative constraint leads to consistent improvements in Naturalness and Functional Equivalence across most models. This indicates that the communicative constraint primarily focuses on facilitating target readers’ comprehension.

\paragraph{Semantic vs. Default Translation.}
The semantic constraint leads to limited improvement across evaluation dimensions, and in some cases is associated with slight performance declines.

To better understand how different constraints shape translation behavior, we analyze the directional preferences induced by the two constrained translation strategies. As shown in Figure~\ref{fig:both_rdar} (b), the communicative constraint consistently yields improvements in Naturalness, Functional Equivalence, and Cultural Adaptation across most models, suggesting that the communicative strategy encourages translations that are more fluent, functionally appropriate, and readily comprehensible to target-language readers.

In contrast, the semantic constraint demonstrates more conservative and asymmetric trade-offs across dimensions. Under this strategy, Fidelity and Contextual Accuracy remain consistently more stable than other dimensions, indicating a clear prioritization of semantic faithfulness and contextual precision. These results suggest that the semantic translation strategy is more effective at preserving the original author’s intended meaning, favoring accurate transmission of source semantics over target-oriented adaptation.

\section{Discussion and Analysis}

% \subsection{Similarity Across Strategies}
\subsection{Default Translation Preference Across Different Strategies}
\label{sec:Default Translation Preference Across Different Strategies}
% \begin{table}[t]
% \centering
% \small
% \begin{tabular}{lcc}
% \toprule
% \textbf{Model} & \textbf{Sim$_{B,Com}$} & \textbf{Sim$_{B,Sem}$} \\
% \midrule
% Llama-3-8B & 0.90 & 0.90 \\
% Llama-3.3-70B & 0.90 & 0.94 \\
% Mixtral-8x7B & 0.75 & 0.74 \\
% Qwen2.5-7B & 0.90 & 0.91 \\
% Qwen2.5-14B & 0.90 & 0.91 \\
% Qwen2.5-32B & 0.90 & 0.93 \\
% Qwen2.5-72B & 0.93 & 0.95 \\
% Qwen3-8B (w/o think) & 0.93 & 0.94 \\
% Qwen3-8B (with think) & 0.90 & 0.91 \\
% GPT-4o & 0.91 & 0.95 \\
% \bottomrule
% \end{tabular}
% \caption{Similarity between baseline translations (B) and translations under semantic (Sem) and communicative (Com) constraints.}
% \label{tab:sim_results}
% \end{table}

\begin{figure}[t]
  \includegraphics[width=\columnwidth]{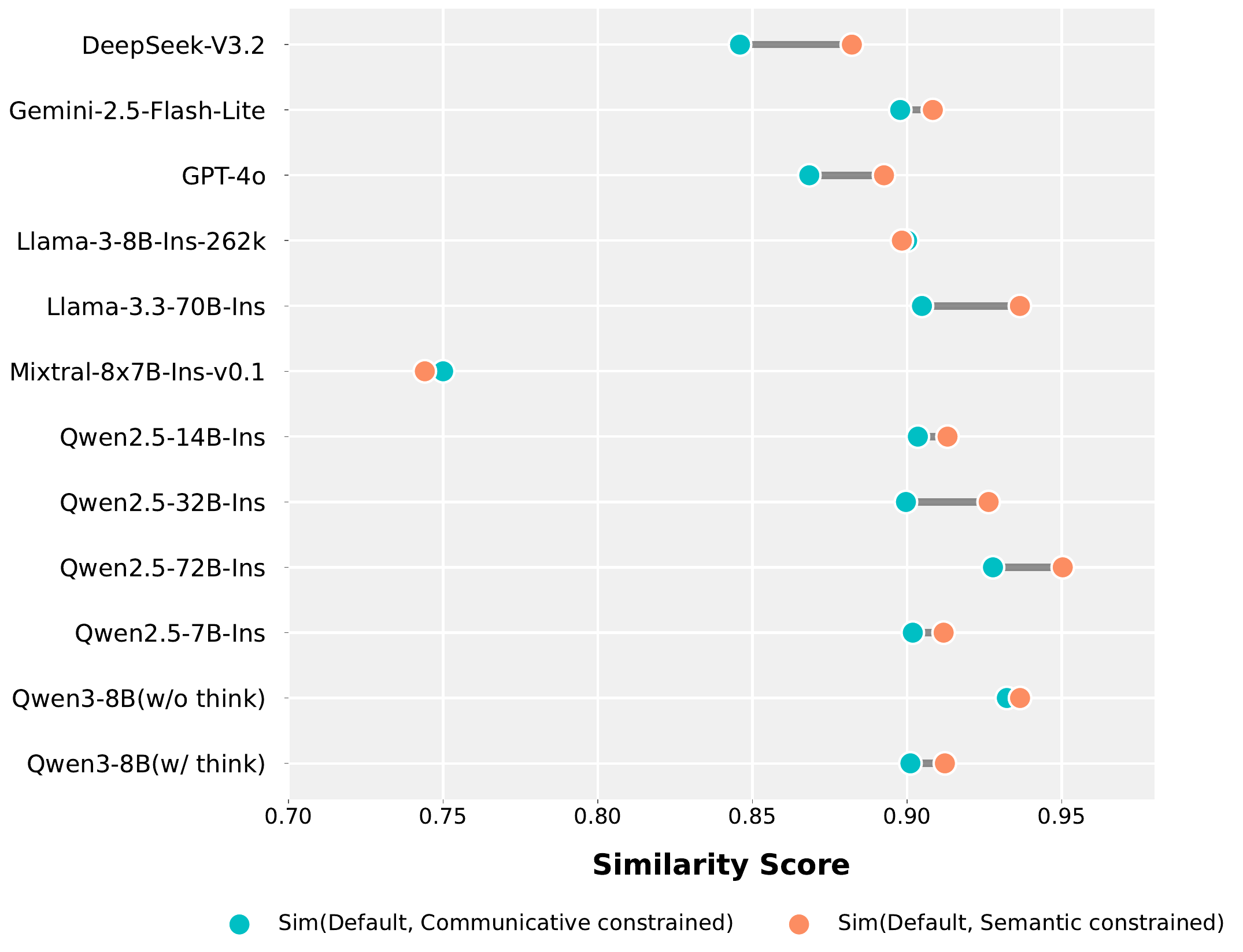}
  \caption{Similarity of default translations to semantic and communicative-constrained translations.}
  \label{fig:sim_compute}
\end{figure}

To characterize default translation behavior of LLMs, we measure cosine similarity between translations under the unconstrained setting and under semantic or communicative constraints.

As shown in Figure~\ref{fig:sim_compute}, default translations consistently exhibit higher similarity to semantic-constrained translations across most models, indicating that in the absence of explicit constraints, models default to a semantic translation mode. This pattern is consistent with Table~\ref{tab:strategy_comparison}, which shows that communicative constraints induce larger performance fluctuations. Additional experiments under varying temperature settings further confirm the stability of this observation (see Appendix~\ref{app:preference}).

\subsection{Performance on Category-wise Culture-Specific Items}
\label{sec:csi-performance}

To examine how CSI translation performance varies across types, we analyze translations with respect to Contextual Accuracy and Cultural Adaptation, which primarily reflect CSIs translation quality. The categorization framework is adapted from Newmark’s taxonomy of CSI~\citep{newmark1988textbook}. 
To maintain analytical clarity, the study is restricted to sentences that contain solely one CSI category.\footnote{Detailed category definitions and the automatic CSI classification procedure are provided in Appendix~\ref{app:csi}.}

As shown in Table~\ref{tab:csi-category-performance}, Geographic and ecological items achieve the highest scores, whereas Language symbols consistently exhibit the lowest performance, reflecting inherent differences in cultural content: geographic and ecological references typically admit direct and conventionalized correspondences across languages, facilitating accurate and adaptive translation, while language symbols are inherently abstract and non-compositional. Representative case studies illustrating these phenomena are provided in Appendix~\ref{sec:category_case_csi}.

\begin{table}[t]
\centering
\small
\begin{tabular}{lcc}
\toprule
\textbf{Cultural Category} & \textbf{CTX} & \textbf{CAD} \\
\midrule
Geographic \& Ecological   & \textbf{5.03} & \textbf{5.30} \\
Language Symbols           & 4.39          & 4.73          \\
Material Culture           & 4.64          & 4.85          \\
Organizations \& Inst.     & 4.88          & 5.05          \\
Social Culture \& Customs  & 4.58          & 4.83          \\
\bottomrule
\end{tabular}

\caption{Translation performance across CSI categories. \textbf{CTX} and \textbf{CAD} denote Contextual Accuracy and Cultural Adaptation scores respectively.}
\label{tab:csi-category-performance}
\end{table}

\subsection{Cultural Translation Knowledge Analysis}
\label{sec:knowledge-analyze}
Furthermore, to investigate the relationship between CSIs translation knowledge and their translation quality, we conduct a probing analysis of cultural translation knowledge, focusing on models’ ability to identify the appropriate rendering of the CSI. In our experiments, we use GPT-4o and construct single-choice questions for CSI translations based on reference translations, allowing the model to answer these questions to assess whether it has correctly mastered the translations.\footnote{The experimental details are presented in the Appendix~\ref{app:generation-of-questions}.}

As shown in Table~\ref{tab:Performance_on_csi_correct}, models exhibit a consistent gain for which all questions are answered correctly compared to sentences where they are not, highlighting the role of cultural translation knowledge in achieving high-quality CSI translation.

\begin{figure}[t]
  \includegraphics[width=\columnwidth]{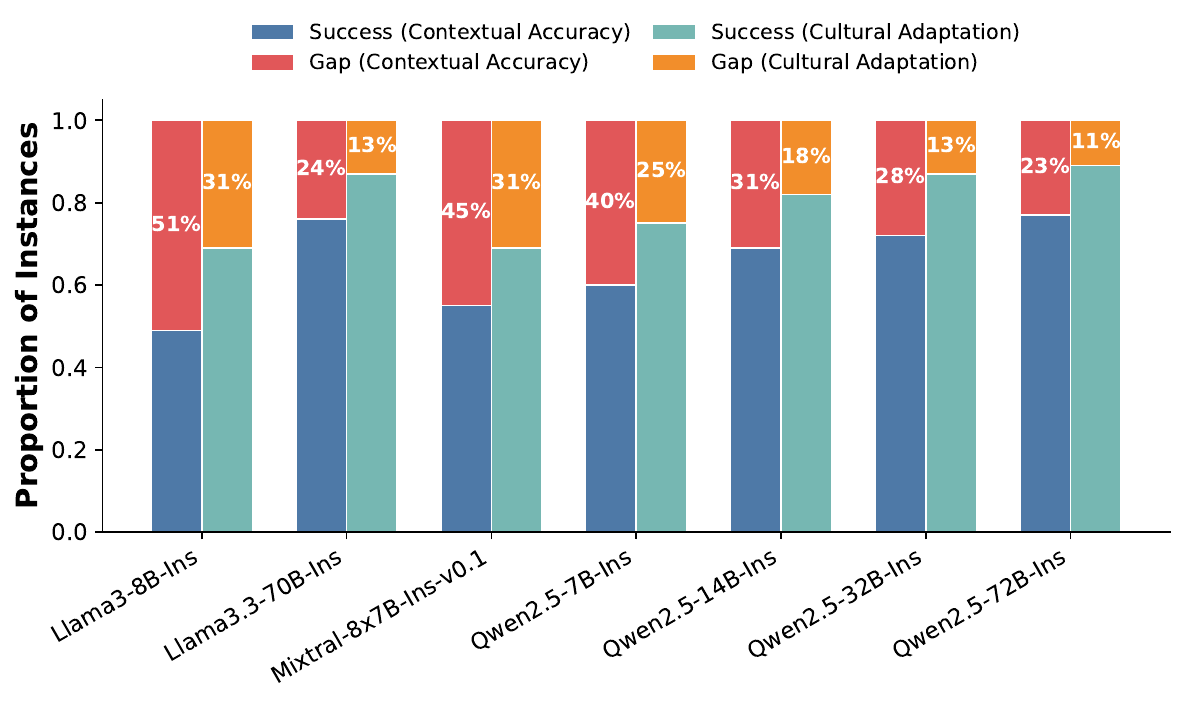}
  \caption{Distribution of CSI translation scores within the knowledge-correct subset. The lower segments indicate scores $\geq$ 4 in the corresponding dimension.}

  \label{fig:knowledge_gap}
\end{figure}

However, possessing knowledge does not inherently guarantee its faithful application. Figure~\ref{fig:knowledge_gap} illustrates the distribution of scores specifically within the knowledge-correct subset, which we define as the collection of instances where the model successfully passes all corresponding probing questions for a given CSI. We identify a persistent "knowledge-application gap": a non-negligible proportion of translations still result in low-quality outputs despite the models "knowing" the correct rendering in the probing task. In Appendix~\ref{app:Cultural Knowledge Probing}, we provide a case to illustrate this phenomenon and explore a simple two-stage inference strategy to partially mitigate this gap.

\begin{table}[t]
\centering
\scriptsize
\setlength{\tabcolsep}{2pt}
\begin{tabular}{lcccc}
\toprule
Model & \multicolumn{2}{c}{CTX} & \multicolumn{2}{c}{CAD} \\
& w/ Know. & w/o Know. & w/ Know. & w/o Know. \\
\midrule
Llama-3-8B-Ins-262k      & 4.01 & 3.23 & 4.44 & 3.85 \\
Llama-3.3-70B-Ins   & 5.02 & 4.34 & 5.17 & 4.64 \\
Mixtral-8x7B-Ins-v0.1    & 4.21 & 3.54 & 4.54 & 3.99 \\
Qwen2.5-7B-Ins      & 4.37 & 3.74 & 4.64 & 4.26 \\
Qwen2.5-14B-Ins     & 4.77 & 4.13 & 4.97 & 4.47 \\
Qwen2.5-32B-Ins     & 4.89 & 4.36 & 5.17 & 4.59 \\
Qwen2.5-72B-Ins     & 5.11 & 4.52 & 5.26 & 4.72 \\
\bottomrule
\end{tabular}
\caption{Performance of translation under Contextual Accuracy (CTX) and Cultural Adaptation (CAD). Scores are shown separately for cases where the model correctly answered all CSI probing questions (\textit{w/ Know}) and where it did not (\textit{w/o Know}).}

\label{tab:Performance_on_csi_correct}
\end{table}

\subsection{The Role of Reference Translation in \textit{LLM-as-a-Judge}}

Following~\citet{qian2024large}, we investigate the role of reference translations in cultural translation evaluation. Under the setting described in Section~\ref{sec:Evaluation}, we perform no-reference evaluation and compute Kendall’s $\tau$ to measure the rank correlation between automatic scores and human judgments across evaluation dimensions.

As shown in Table~\ref{tab:ref_effect}, incorporating reference translations generally improves the agreement between automatic evaluation metrics and human judgments. Our results show that reference translations play a crucial role in cultural translation evaluation. They provide reference renderings for CSIs and serve as a factual baseline for detecting fine-grained semantic errors, while also helping calibrate whether the translation style aligns with target-language norms.
Detailed qualitative analyses of specific cases, presented in Appendix~\ref{app:eval_case_studies}, further illustrate these effects. 

\begin{table}[t]
\centering
\small
\begin{tabular}{lcc}
\toprule
\textbf{Dimension} & \textbf{$\tau$ w/ Ref} & \textbf{$\tau$ w/o Ref} \\
\midrule
Contextual Accuracy     & 0.45 & 0.42 \\
Cultural Adaptation     & 0.39 & 0.36 \\
Fidelity                & 0.46 & 0.41 \\
Functional Equivalence  & 0.45 & 0.42 \\
Naturalness             & 0.47 & 0.45 \\
\bottomrule
\end{tabular}
\caption{Agreement between LLM scores and human judgments measured by Kendall’s $\tau$, with and without reference translations. }
\label{tab:ref_effect}
\end{table}

\section{Conclusion}

In this paper, we presented \textbf{CanMT}, a novel-driven benchmark for culture-aware machine translation spanning 12 directions. Under five theoretically grounded evaluation dimensions, we systematically assessed modern LLMs. Beyond establishing that scaling and reasoning improve performance, our systematic evaluation reveals that while communicative strategies significantly enhance target readers' comprehension, semantic strategies are more effective at accurately conveying the author's intent. Notably, we found that LLMs exhibit a default bias toward semantic translation in unconstrained settings. Our analysis across CSIs identifies a difficulty hierarchy, where abstract language symbols remain the most challenging category. Crucially, we discovered a persistent "knowledge-application gap", demonstrating that possessing cultural knowledge does not inherently guarantee its faithful application in translation. Finally, we showed that reference translations are indispensable for calibrating automatic metrics with human judgment. Overall, we introduce \textbf{CanMT}, providing a rigorous diagnostic framework for the community, and highlight several promising directions for improving culture-aware machine translation, including stronger inference-time reasoning, richer cultural knowledge coverage, and more effective bridging of the "knowledge-application gap".

\section*{Limitations}
\label{sec:limitations}

Despite the systematic approach taken in this study, several limitations remain. 
First, regarding data source, our benchmark relies primarily on literary fiction. While novels offer high-density cultural content, they may not fully represent the linguistic diversity found in other culturally rich domains, such as social media, historical archives, or oral history records. Additionally, the reliance on public domain or classic texts might introduce a temporal bias, potentially overlooking contemporary cultural neologisms.
Second, our analysis focuses on sentence-level translation. Cultural meaning is often constructed discursively across broader contexts (paragraph or document level). Future work should extend this evaluation to discourse-level settings to capture long-range cultural dependencies and consistency.

\section*{Ethical Considerations}
\label{sec:ethics}

We construct \textbf{CanMT} from classic novels and their professional translations. For works that may remain under copyright in some jurisdictions, we use only sentence-level aligned excerpts for non-commercial academic research, rather than redistributing full texts. To reduce redistribution risk, any release will exclude full copyrighted texts and will be restricted to processed artifacts and, when necessary, reconstruction scripts or document identifiers. The dataset is intended for research use only. Code and other artifacts created by the authors will be released under the MIT license. Third-party copyrighted texts and translations, if included in processed form, remain subject to their original copyright and access conditions and are not relicensed by us. We used Gemini to correct grammatical errors in the manuscript.

\section*{Acknowledgements}

Xiaocheng Feng is the corresponding author of this work.
We thank the anonymous reviewers for their insightful comments.
This work was supported by the National Natural Science Foundation of China (NSFC) (grant 62522603, 62276078), the Key R\&D Program of Heilongjiang via grant 2022ZX01A32, the Fundamental Research Funds for the Central Universities (XNJKKGYDJ2024013).

\bibliography{custom}

\appendix
\section*{Appendix}
\addcontentsline{toc}{section}{Appendix}

\section{Dataset Construction and Statistics}
\label{app:dataset}

\subsection{Dataset Statistics}
\label{app:dataset_stats}

To provide a more comprehensive characterization of the dataset, we present corpus-level statistics across different language pairs, along with the distribution of culture-specific item categories.

Table~\ref{tab:corpus_stats} reports the number of sentences and token-level statistics for each translation direction. Table~\ref{tab:csi_dist} shows the distribution of CSI categories across languages, reflecting a diverse range of cultural phenomena.

\begin{table*}[t]
\centering
\small
\renewcommand{\arraystretch}{1.2}
\setlength{\tabcolsep}{5pt}
\resizebox{\textwidth}{!}{
\begin{tabular}{lcccccccccccc}
\toprule
\textbf{Metric} 
& ru2zh & ja2zh & en2zh & en2ja & zh2ru & zh2es 
& en2es & es2en & zh2en & es2zh & ja2en & zh2ja \\
\midrule
\#Sent. 
& 125 & 106 & 111 & 103 & 95 & 146 
& 106 & 134 & 125 & 166 & 116 & 137 \\

Avg. Len. 
& 34.42 & 26.03 & 32.40 & 32.20 & 23.92 & 25.24 
& 33.51 & 50.71 & 25.38 & 46.33 & 26.16 & 24.92 \\

Std. 
& 18.98 & 12.97 & 20.46 & 17.40 & 11.68 & 12.96 
& 20.28 & 38.70 & 13.53 & 25.41 & 13.35 & 12.79 \\
\bottomrule
\end{tabular}
}
\caption{Corpus statistics across translation directions. Avg. Len. reports the average source-sentence length measured in tokens, and Std. denotes the standard deviation of source-sentence token lengths, reflecting the variability of sentence length within each translation direction.}
\label{tab:corpus_stats}
\end{table*}

\begin{table}[t]
\centering
\small
\renewcommand{\arraystretch}{1.2}
\setlength{\tabcolsep}{4pt}
\resizebox{\columnwidth}{!}{
\begin{tabular}{lccccc}
\toprule
\textbf{Lang} & \textbf{Geo} & \textbf{Lang Sym.} & \textbf{Mat.} & \textbf{Org.} & \textbf{Soc.} \\
\midrule
en & 36  & 95  & 58  & 14  & 22  \\
es & 166 & 355 & 279 & 80  & 308 \\
ja & 94  & 145 & 188 & 31  & 206 \\
ru & 92  & 244 & 309 & 196 & 288 \\
zh & 17  & 128 & 81  & 39  & 108 \\
\bottomrule
\end{tabular}
}
\caption{Distribution of CSI categories across languages. Geo: geographic and ecological items; Lang Sym.: language-related symbols; Mat.: material culture; Org.: organizations and institutions; Soc.: social customs.}
\label{tab:csi_dist}
\end{table}

\subsection{Human Filtering Instruction}
\label{app:human_filter}

\begin{figure}[t]
  \centering
  \includegraphics[width=\columnwidth]{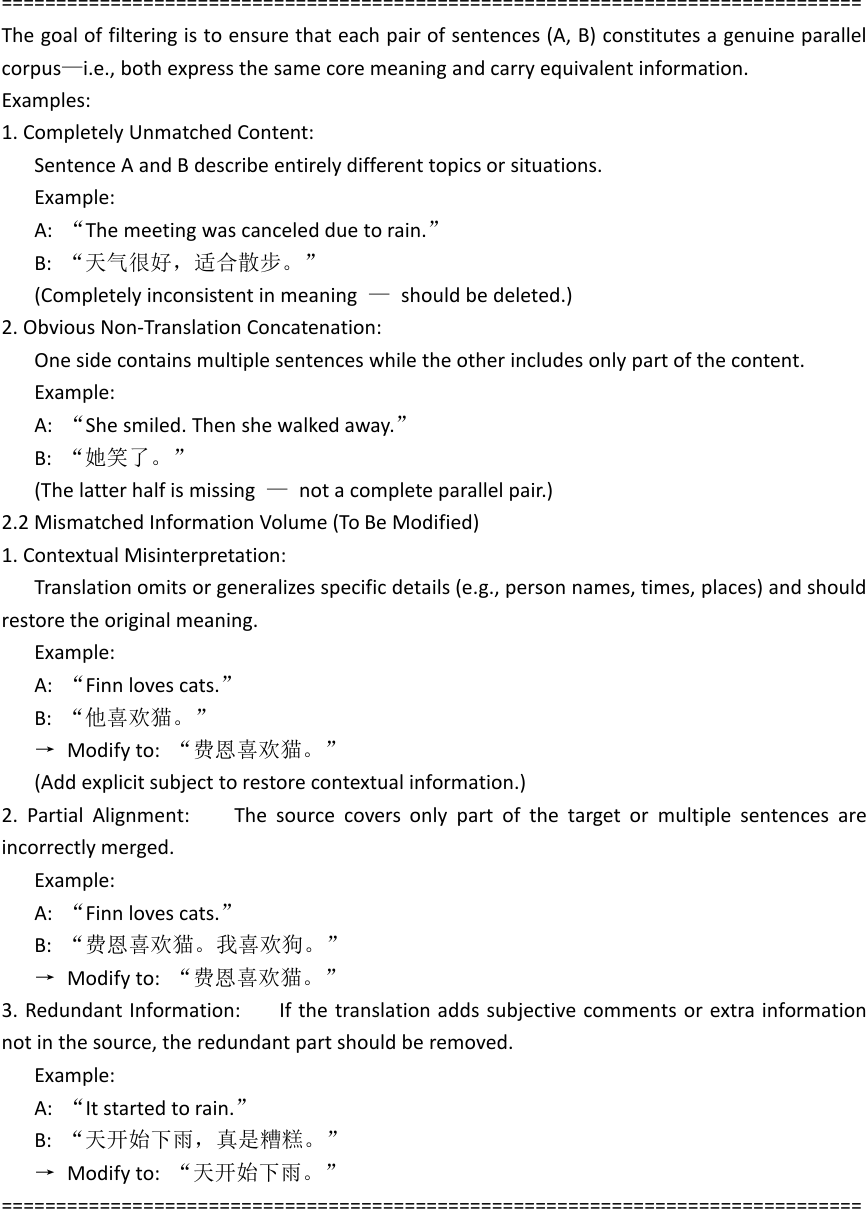}
  \caption{Instruction for human data filtering.}
  \label{fig:human_filter}
\end{figure}

During our annotation process, all human annotators were compensated appropriately. The manual data filtering instruction is illustrated in Figure~\ref{fig:human_filter}.

\section{Evaluation via Multi-Dimensions}

\subsection{Human Evaluation}
\label{app:human_eval}

\begin{figure}[t]
  \centering
  \includegraphics[width=\columnwidth]{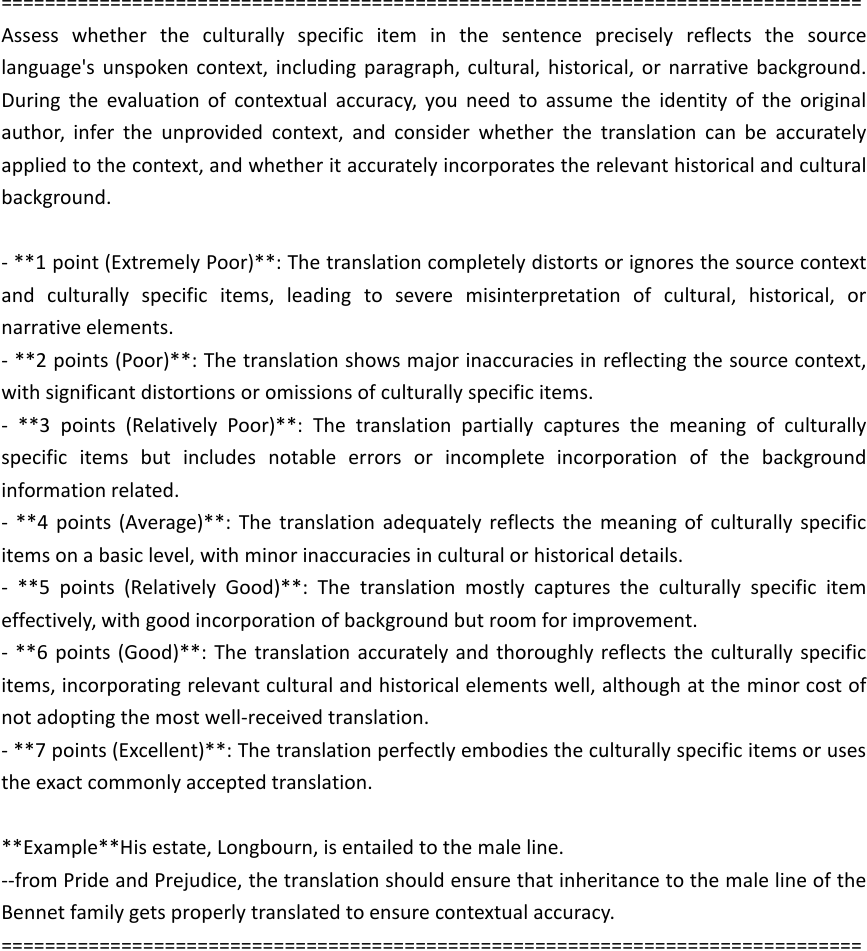}
  \caption{Instruction for human eval Contextual Accuracy.}
  \label{fig:human_ctx_prompt}
\end{figure}

\begin{figure}[t]
  \centering
  \includegraphics[width=\columnwidth]{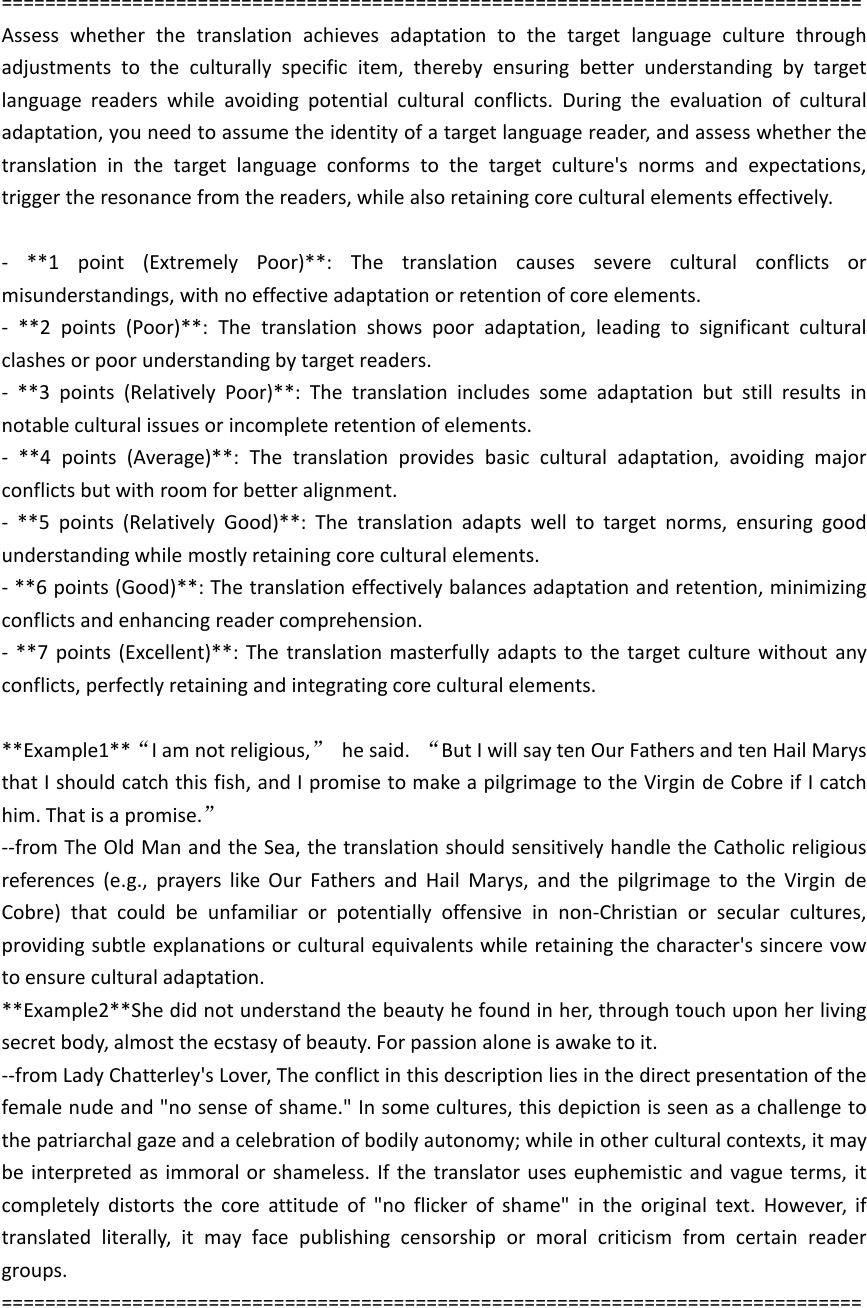}
  \caption{Instruction for human eval Cultural Adaptation.}
  \label{fig:human_cul_prompt}
\end{figure}

\begin{figure}[t]
  \centering
  \includegraphics[width=\columnwidth]{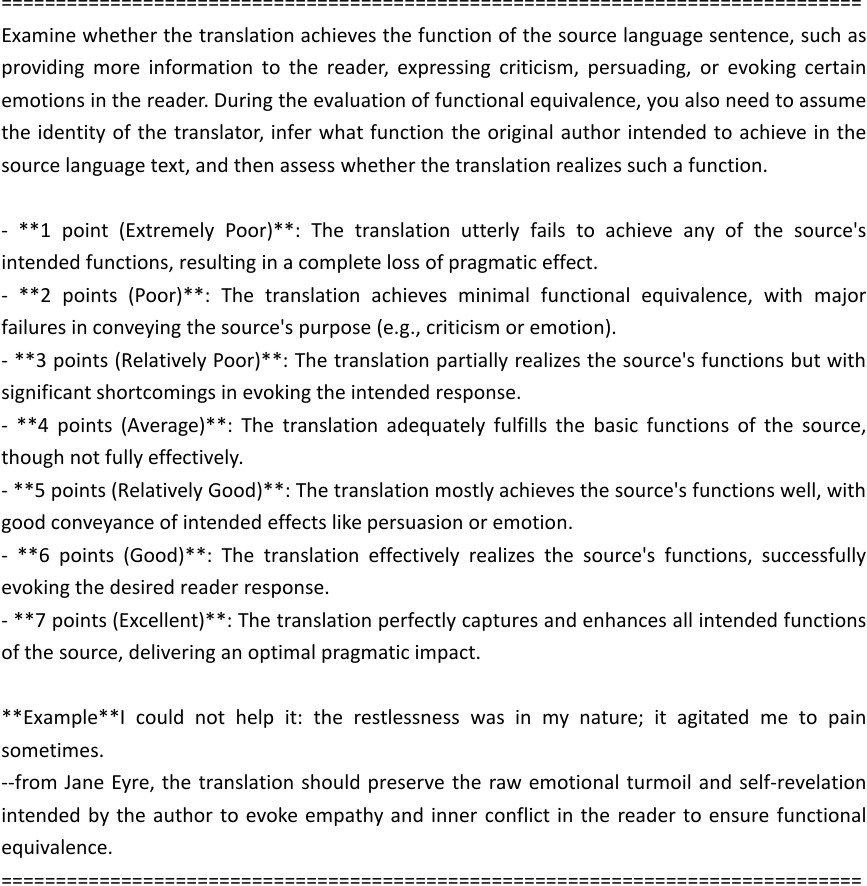}
  \caption{Instruction for human eval Functional Equivalence.}
  \label{fig:human_fun_prompt}
\end{figure}

\begin{figure}[t]
  \centering
  \includegraphics[width=\columnwidth]{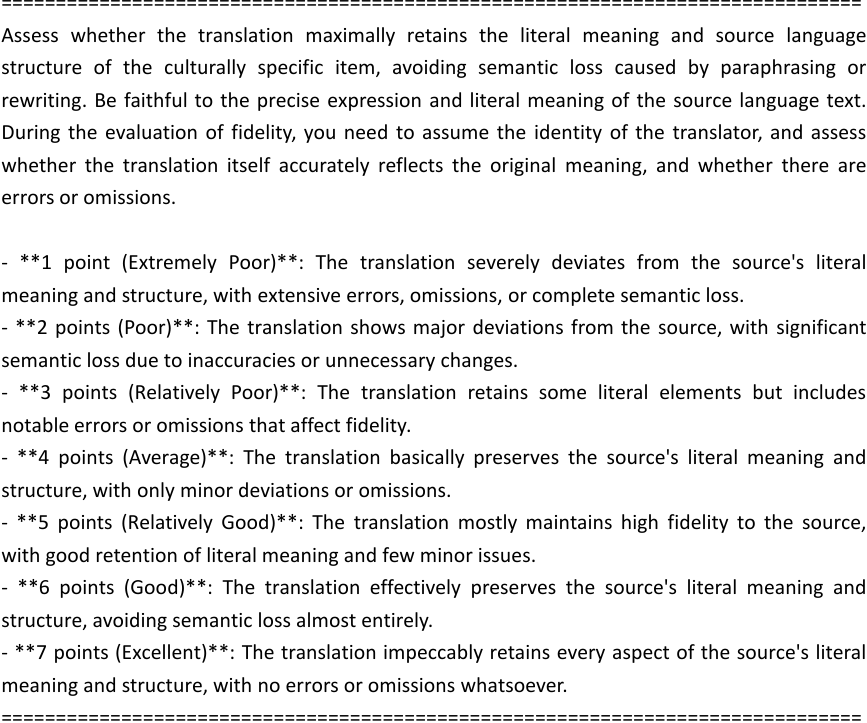}
  \caption{Instruction for human eval Fidelity.}
  \label{fig:human_fid_prompt}
\end{figure}

\begin{figure}[t]
  \centering
  \includegraphics[width=\columnwidth]{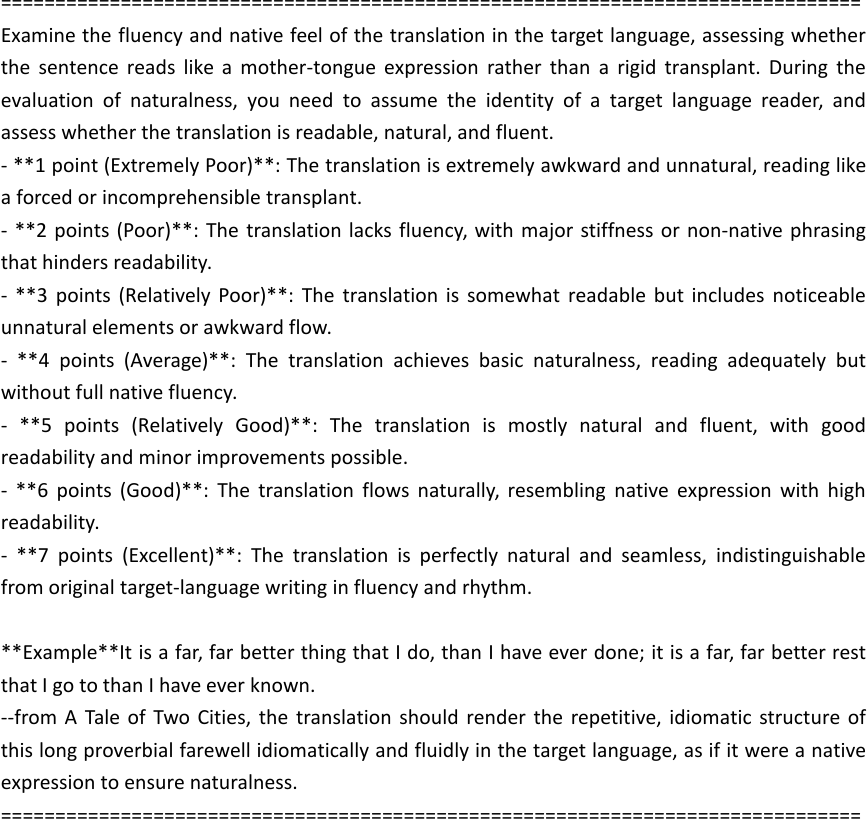}
  \caption{Instruction for human eval Naturalness.}
  \label{fig:human_nat_prompt}
\end{figure}

During our annotation process, all human annotators were compensated appropriately. The manual evaluation procedure is illustrated in Figure~\ref{fig:human_ctx_prompt}--\ref{fig:human_nat_prompt}.

\subsection{Scoring Rubrics for Evaluation Dimensions}
\label{app:scoring_rubrics}

To ensure consistency and interpretability of the multi-dimensional evaluation, we provide detailed scoring rubrics for each evaluation dimension. All dimensions are rated on a 7-point Likert scale, where higher scores indicate better translation quality.

\subsubsection{Contextual Accuracy}

Contextual Accuracy measures whether the translation of the CSI correctly reflects the meaning intended by the source text within its discourse context.

\begin{itemize}
    \item \textbf{1 point (Extremely Poor)}: The translation completely distorts or ignores the source context and culturally specific items, leading to severe misinterpretation of cultural, historical, or narrative elements.
    \item \textbf{2 points (Poor)}: The translation shows major inaccuracies in reflecting the source context, with significant distortions or omissions of culturally specific items.
    \item \textbf{3 points (Relatively Poor)}: The translation partially captures the meaning of culturally specific items but includes notable errors or incomplete incorporation of the background information related.
    \item \textbf{4 points (Average)}: The translation adequately reflects the meaning of culturally specific items on a basic level, with minor inaccuracies in cultural or historical details.
    \item \textbf{5 points (Relatively Good)}: The translation mostly captures the culturally specific item effectively, with good incorporation of background but room for improvement.
    \item \textbf{6 points (Good)}: The translation accurately and thoroughly reflects the culturally specific items, incorporating relevant cultural and historical elements well, although at the minor cost of not adopting the most well-received translation.
    \item \textbf{7 points (Excellent)}: The translation uses the exact commonly accepted translation. Or The translation perfectly embodies the culturally specific items. 
\end{itemize}

\subsubsection{Cultural Adaptation}

Cultural Adaptation assesses whether the translation appropriately adapts the CSI for target-language readers, ensuring cultural intelligibility and avoiding cultural conflict or confusion.

\begin{itemize}
    \item \textbf{1 point (Extremely Poor)}: The translation of CSIs causes severe cultural conflicts, with no effective adaptation or retention of core elements.
    \item \textbf{2 points (Poor)}: The translation of CSIs shows poor adaptation, leading to significant cultural clashes or poor understanding by target readers.
    \item \textbf{3 points (Relatively Poor)}: The translation of CSIs includes some adaptation but still results in notable cultural issues or incomplete retention of elements.
    \item \textbf{4 points (Average)}: The translation of CSIs provides basic cultural adaptation, avoiding major conflicts but with room for better alignment.
    \item \textbf{5 points (Relatively Good)}: The translation of CSIs adapts well to target norms, ensuring good understanding while mostly retaining core cultural elements.
    \item \textbf{6 points (Good)}: The translation of CSIs effectively balances adaptation and retention, minimizing conflicts and enhancing reader comprehension.
    \item \textbf{7 points (Excellent)}: The translation of CSIs masterfully adapts to the target culture without any conflicts, perfectly retaining and integrating core cultural elements. If the translation is widely recognized, it can be awarded with 7 points.
\end{itemize}

\subsubsection{Functional Equivalence}

Functional Equivalence measures whether the translation fulfills the communicative function of the source text, such as informing, persuading, or expressing attitude.

\begin{itemize}
    \item \textbf{1 point (Extremely Poor)}: The translation utterly fails to achieve any of the source's intended functions, resulting in a complete loss of pragmatic effect.
    \item \textbf{2 points (Poor)}: The translation achieves minimal functional equivalence, with major failures in conveying the source's purpose (e.g., criticism or emotion).
    \item \textbf{3 points (Relatively Poor)}: The translation partially realizes the source's functions but with significant shortcomings in evoking the intended response.
    \item \textbf{4 points (Average)}: The translation adequately fulfills the basic functions of the source, though not fully effectively.
    \item \textbf{5 points (Relatively Good)}: The translation mostly achieves the source's functions well, with good conveyance of intended effects like the offering of information or expression emotion.
    \item \textbf{6 points (Good)}: The translation effectively realizes the source's functions, successfully evoking the desired reader's emotional response or informing them more about the story.
    \item \textbf{7 points (Excellent)}: The translation perfectly captures and enhances all intended functions of the source, delivering an optimal pragmatic impact.
\end{itemize}

\subsubsection{Fidelity}

Fidelity evaluates the extent to which the translation preserves the literal meaning and core informational content of the source text.

\begin{itemize}
    \item \textbf{1 point (Extremely Poor)}: The translation severely deviates from the source's literal meaning and structure, with extensive errors, omissions, or complete semantic loss.
    \item \textbf{2 points (Poor)}: The translation shows major deviations from the source, with significant semantic loss due to inaccuracies or unnecessary changes.
    \item \textbf{3 points (Relatively Poor)}: The translation retains some literal elements but includes notable errors or omissions that affect fidelity.
    \item \textbf{4 points (Average)}: The translation basically preserves the source's literal meaning and structure, with only minor deviations or omissions.
    \item \textbf{5 points (Relatively Good)}: The translation mostly maintains high fidelity to the source, with good retention of literal meaning and few minor issues.
    \item \textbf{6 points (Good)}: The translation effectively preserves the source's literal meaning and structure, avoiding semantic loss almost entirely.
    \item \textbf{7 points (Excellent)}: The translation impeccably retains every aspect of the source's literal meaning and structure, with no errors or omissions whatsoever.
\end{itemize}

\subsubsection{Naturalness}

Naturalness assesses the fluency and idiomaticity of the translation in the target language.

\begin{itemize}
    \item \textbf{1 point (Extremely Poor)}: The translation is extremely awkward and unnatural, reading like a forced or incomprehensible transplant.
    \item \textbf{2 points (Poor)}: The translation lacks fluency, with major stiffness or non-native phrasing that hinders readability.
    \item \textbf{3 points (Relatively Poor)}: The translation is somewhat readable but includes noticeable unnatural elements or awkward flow.
    \item \textbf{4 points (Average)}: The translation achieves basic naturalness, reading adequately but without full native fluency.
    \item \textbf{5 points (Relatively Good)}: The translation is mostly natural and fluent, with good readability and minor improvements possible.
    \item \textbf{6 points (Good)}: The translation flows naturally, resembling native expression with high readability.
    \item \textbf{7 points (Excellent)}: The translation is perfectly natural and seamless, indistinguishable from original target-language writing in fluency and rhythm.
\end{itemize}

\subsection{Prompts for LLM Evaluation}
\label{app:eval_prompts}

\begin{figure}[t]
  \centering
  \includegraphics[width=\columnwidth]{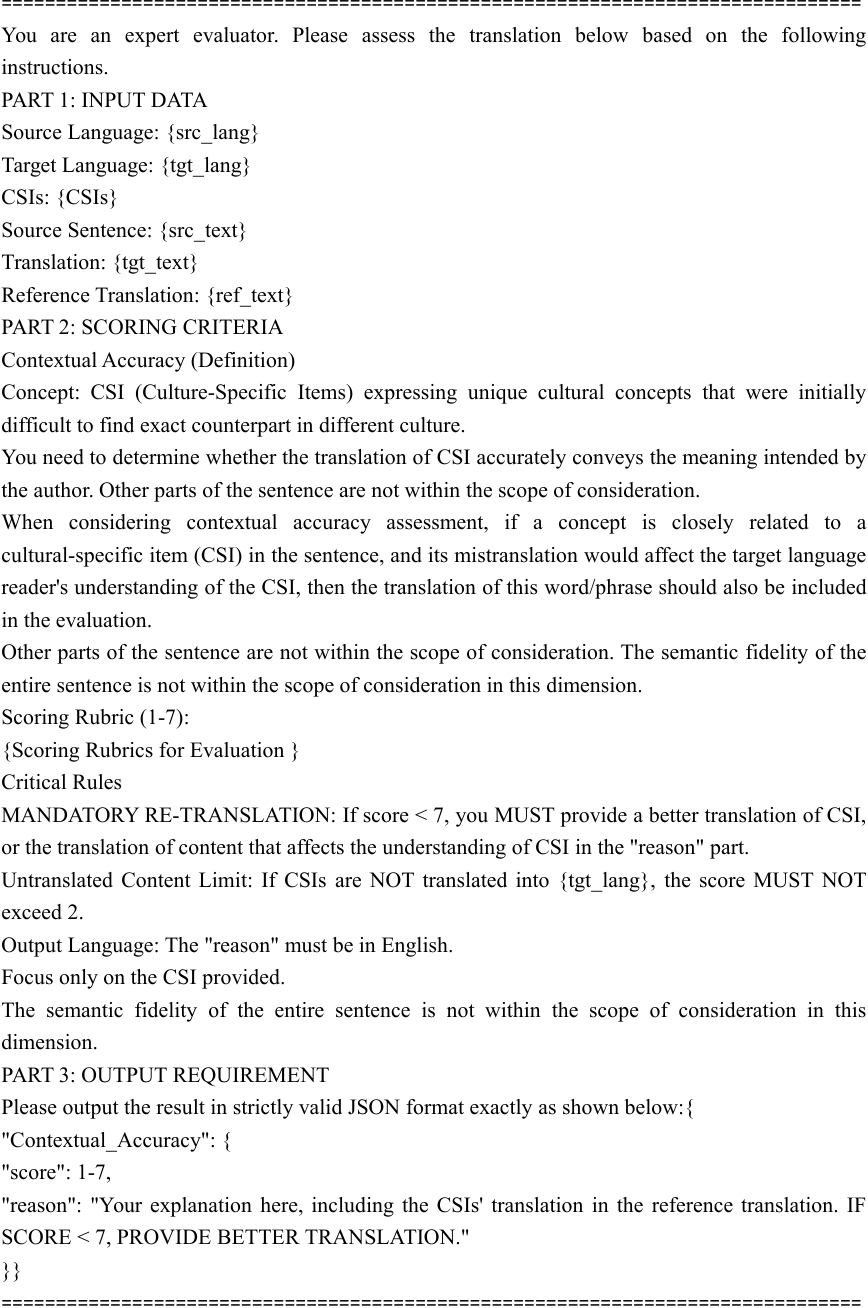}
  \caption{Prompt for eval Contextual Accuracy.}
  \label{fig:ctx_prompt}
\end{figure}

\begin{figure}[t]
  \centering
  \includegraphics[width=\columnwidth]{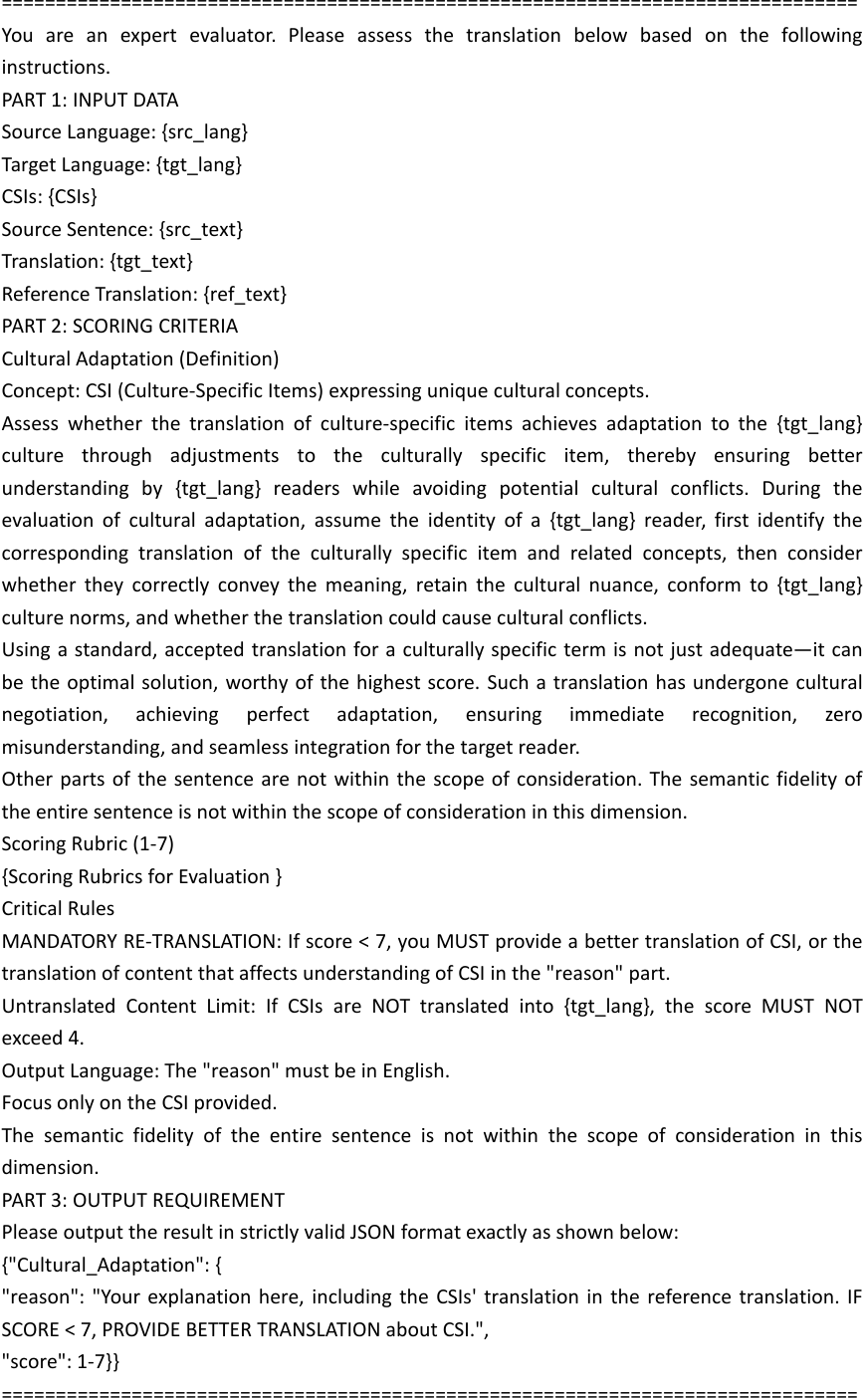}
  \caption{Prompt for eval Cultural Adaptation.}
  \label{fig:/cultural_adapt_prompt}
\end{figure}

\begin{figure}[t]
  \centering
  \includegraphics[width=\columnwidth]{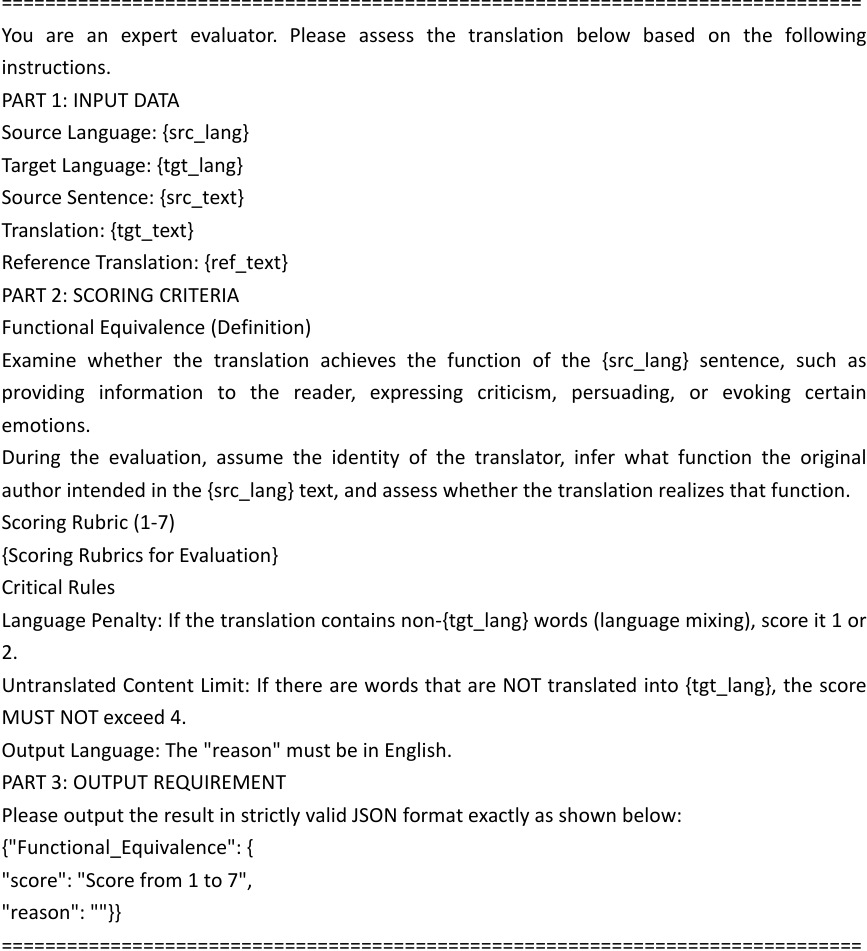}
  \caption{Prompt for eval Functional Equivalence.}
  \label{fig:Fun_Eq_prompt}
\end{figure}

\begin{figure}[t]
  \centering
  \includegraphics[width=\columnwidth]{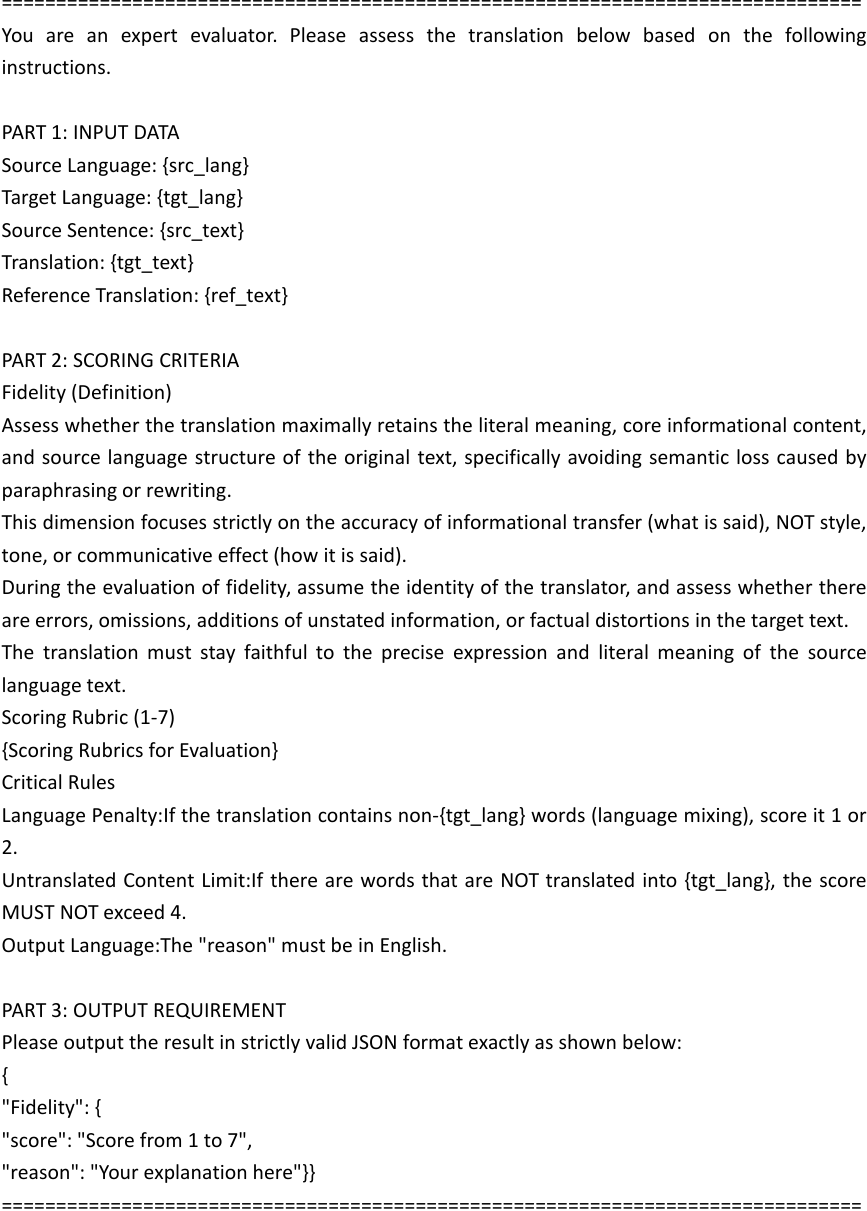}
  \caption{Prompt for eval Fidelity.}
  \label{fig:fid_prompt}
\end{figure}

\begin{figure}[t]
  \centering
  \includegraphics[width=\columnwidth]{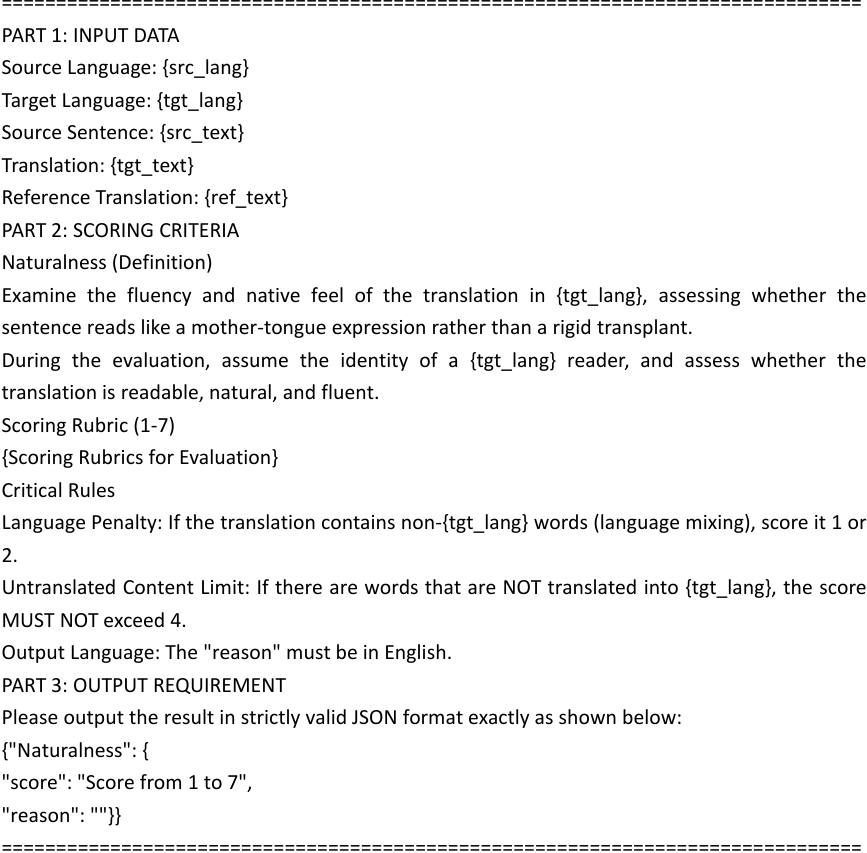}
  \caption{Prompt for eval Naturalness.}
  \label{fig:nat_prompt}
\end{figure}

For reproducibility, we provide the exact prompts used in our evaluation for each dimension, as shown in Figure~\ref{fig:ctx_prompt}--\ref{fig:nat_prompt}.

\subsection{Robustness of LLM-based Evaluation}
\label{app:judge_robustness}

To evaluate the robustness of the LLM-based scoring procedure, we conduct additional experiments using multiple independent LLM judges.

Specifically, we sample approximately 1,400 translation instances and re-score them using GPT-5-nano, DeepSeek-V3.2, and Grok-4.1. For each evaluation dimension, we compute Kendall’s $\tau$ between the scores produced by each judge and the GPT-5-nano scores used in the main experiments.

\begin{table}[t]
\centering
\small
\resizebox{\columnwidth}{!}{
\begin{tabular}{lccc}
\toprule
\textbf{Dimension} & \textbf{GPT-5-nano} & \textbf{DeepSeek-V3.2} & \textbf{Grok-4.1} \\
\midrule
Ctx. Acc.   & 0.638 & 0.487 & 0.578 \\
Cul. Adapt. & 0.615 & 0.490 & 0.575 \\
Fidelity    & 0.592 & 0.378 & 0.541 \\
Func. Eq.   & 0.562 & 0.414 & 0.553 \\
Naturalness & 0.586 & 0.405 & 0.544 \\
\bottomrule
\end{tabular}
}
\caption{Kendall’s $\tau$ correlations across different LLM judges.}
\label{tab:judge_consistency}
\end{table}

As shown in Table~\ref{tab:judge_consistency}, the scoring results exhibit moderate to strong correlations across different LLM judges, indicating consistent ranking behavior.

These findings suggest that our evaluation results are not sensitive to the choice of a specific LLM evaluator, and that the observed performance trends are robust across different judging models.

\section{Experimental Settings}
\label{app:Experimental Setting}

\subsection{Translation Prompts}

We use different prompts to implement default, communicative, and semantic translation strategies. The full prompt templates are summarized in Table~\ref{tab:translation_prompts}. For specialized MT models, we adopt their recommended prompt templates.

\begin{table*}[t]
\centering
\small
\renewcommand{\arraystretch}{1.4} 
\resizebox{\textwidth}{!}{
\begin{tabular}{p{3cm} p{10cm}}
\toprule
\textbf{Type} & \textbf{Prompt} \\
\midrule
Default &
You are a professional translation expert. Your task is to translate text from \{src\_lang\} to \{tgt\_lang\}. Output only the translated text, without any explanation. \\

Communicative &
You are an award-winning literary translator. Translate the source text from \{src\_lang\} to \{tgt\_lang\}, applying the strategy of \textbf{communicative translation}. This strategy prioritizes naturalness, engagement, and cultural appropriateness. Adapt phrasing, idioms, and culturally specific items to ensure fluency, while preserving the original meaning. Rewording is acceptable for clarity, but do not omit content. Output only the translated text. \\

Semantic &
You are an award-winning literary translator. Translate the source text from \{src\_lang\} to \{tgt\_lang\}, applying the strategy of \textbf{semantic translation}. This strategy prioritizes fidelity to the original meaning, structure, tone, and cultural context. Stay close to the source wording and avoid unnecessary adaptation. Do not add or omit content. Output only the translated text. \\
\bottomrule
\end{tabular}
}
\caption{Translation prompts used for different strategies.}
\label{tab:translation_prompts}
\end{table*}

\subsection{Decoding Settings}

Open-source models and machine translation systems are decoded using greedy decoding, except for the DeepSeek series, which follows its default configuration.

API-based models are evaluated using their default decoding settings, reflecting their standard usage.

To ensure comparability across systems, we additionally rerun experiments with temperature fixed to $T=0$ for all API-based models (See Table~\ref{tab:rerun_model_translation_results}).

\begin{table*}[t]
\centering
\small
\setlength{\tabcolsep}{4pt}
\resizebox{\textwidth}{!}{%
\begin{tabular}{lccccccccccccc}
\toprule
Model & en2es & en2ja & en2zh & es2en & es2zh & ja2en & ja2zh & ru2zh & zh2en & zh2es & zh2ja & zh2ru & Overall\_Avg \\
\midrule
deepseek-r1            & 4.94 & 5.26 & 5.29 & 5.31 & 4.92 & 5.08 & 5.18 & 4.99 & 5.33 & 5.08 & 5.30 & 5.18 & 5.15 \\
deepseek-v3.2          & 5.09 & 5.35 & 5.47 & 5.22 & 5.20 & 5.06 & 5.28 & 5.26 & 5.48 & 5.26 & 5.47 & 4.73 & 5.24 \\
gemini-2.5-flash-lite  & 4.96 & 5.19 & 5.32 & 5.03 & 4.81 & 4.96 & 4.99 & 4.94 & 5.43 & 5.07 & 5.33 & 5.06 & 5.09 \\
gpt-4                  & 4.86 & 5.17 & 5.25 & 5.20 & 4.83 & 4.97 & 5.12 & 4.91 & 5.42 & 5.20 & 5.23 & 5.17 & 5.11 \\
gpt-4o                 & 4.94 & 5.24 & 5.27 & 5.23 & 4.80 & 4.95 & 5.22 & 4.88 & 5.47 & 5.15 & 5.27 & 4.92 & 5.11 \\
grok-4.1               & 4.97 & 5.33 & 5.55 & 5.18 & 5.10 & 4.88 & 5.17 & 5.12 & 5.57 & 5.23 & 5.14 & 5.37 & 5.22 \\
\bottomrule
\end{tabular}%
}
\caption{Translation results across models and language directions with temperature fixed to $T=0$.}
\label{tab:rerun_model_translation_results}
\end{table*}

\section{Detailed Evaluation Results}

\subsection{Automatic Evaluation with COMET}
\label{app:comet_results}

To provide a more comprehensive evaluation of translation quality, we additionally report results using an automatic metric. Specifically, we employ COMET (\texttt{wmt22-comet-da}) to evaluate the translations presented in Table~\ref{tab:comet_results}.

\begin{table*}[t]
\centering
\small
\renewcommand{\arraystretch}{1.15}
\setlength{\tabcolsep}{3pt}
\resizebox{\textwidth}{!}{
\begin{tabular}{lccccccccccccc}
\toprule
\textbf{Model} 
& \textbf{En$\rightarrow$Es} & \textbf{En$\rightarrow$Ja} & \textbf{En$\rightarrow$Zh} 
& \textbf{Es$\rightarrow$En} & \textbf{Es$\rightarrow$Zh} & \textbf{Ja$\rightarrow$En} 
& \textbf{Ja$\rightarrow$Zh} & \textbf{Ru$\rightarrow$Zh} 
& \textbf{Zh$\rightarrow$En} & \textbf{Zh$\rightarrow$Es} 
& \textbf{Zh$\rightarrow$Ja} & \textbf{Zh$\rightarrow$Ru} & \textbf{Avg} \\
\midrule

\rowcolor{gray!20}
\multicolumn{14}{c}{\textbf{Proprietary LLMs}} \\

GPT-4                  & 0.785 & 0.814 & 0.811 & 0.724 & 0.746 & 0.743 & 0.802 & 0.781 & 0.771 & 0.767 & 0.862 & 0.762 & 0.781 \\
GPT-4o                 & 0.772 & 0.799 & 0.799 & 0.712 & 0.718 & 0.735 & 0.789 & 0.768 & 0.759 & 0.738 & 0.851 & 0.760 & 0.767 \\
Gemini-2.5-flash-lite  & 0.786 & 0.817 & 0.798 & 0.720 & 0.729 & 0.725 & 0.781 & 0.778 & 0.756 & 0.752 & 0.850 & 0.774 & 0.772 \\
Grok-4.1               & 0.787 & 0.825 & 0.812 & 0.727 & 0.749 & 0.724 & 0.790 & 0.799 & 0.763 & 0.760 & 0.850 & 0.768 & 0.779 \\

\midrule
\rowcolor{gray!20}
\multicolumn{14}{c}{\textbf{Open-source LLMs}} \\

LLaMA-3-8B-262k        & 0.716 & 0.669 & 0.704 & 0.668 & 0.612 & 0.652 & 0.640 & 0.683 & 0.716 & 0.684 & 0.679 & 0.677 & 0.675 \\
LLaMA-3.3-70B-Inst     & 0.776 & 0.801 & 0.798 & 0.715 & 0.724 & 0.707 & 0.765 & 0.769 & 0.757 & 0.742 & 0.845 & 0.759 & 0.763 \\

Mixtral-8x7B-Inst      & 0.674 & 0.528 & 0.480 & 0.684 & 0.429 & 0.628 & 0.467 & 0.420 & 0.719 & 0.635 & 0.468 & 0.568 & 0.558 \\

Qwen2.5-7B-Inst        & 0.724 & 0.742 & 0.780 & 0.697 & 0.699 & 0.696 & 0.762 & 0.748 & 0.751 & 0.700 & 0.784 & 0.668 & 0.729 \\
Qwen2.5-14B-Inst       & 0.762 & 0.784 & 0.803 & 0.704 & 0.719 & 0.713 & 0.781 & 0.776 & 0.749 & 0.727 & 0.807 & 0.674 & 0.750 \\
Qwen2.5-32B-Inst       & 0.764 & 0.760 & 0.810 & 0.711 & 0.721 & 0.724 & 0.777 & 0.775 & 0.758 & 0.721 & 0.784 & 0.690 & 0.750 \\
Qwen2.5-72B-Inst       & 0.780 & 0.808 & 0.803 & 0.723 & 0.728 & 0.720 & 0.790 & 0.785 & 0.761 & 0.751 & 0.838 & 0.768 & 0.771 \\

Qwen3-4B (w/o think)    & 0.741 & 0.770 & 0.780 & 0.691 & 0.706 & 0.670 & 0.747 & 0.761 & 0.748 & 0.713 & 0.812 & 0.712 & 0.737 \\
Qwen3-4B (w/ think)       & 0.737 & 0.767 & 0.781 & 0.685 & 0.688 & 0.674 & 0.739 & 0.740 & 0.745 & 0.715 & 0.809 & 0.723 & 0.733 \\
Qwen3-8B (w/o think)    & 0.757 & 0.797 & 0.799 & 0.695 & 0.726 & 0.697 & 0.774 & 0.765 & 0.751 & 0.732 & 0.832 & 0.743 & 0.756 \\
Qwen3-8B (w/ think)       & 0.756 & 0.770 & 0.800 & 0.699 & 0.708 & 0.690 & 0.766 & 0.756 & 0.752 & 0.731 & 0.834 & 0.745 & 0.751 \\
Qwen3-14B (w/o think)   & 0.773 & 0.804 & 0.805 & 0.707 & 0.729 & 0.708 & 0.778 & 0.783 & 0.763 & 0.748 & 0.841 & 0.763 & 0.767 \\
Qwen3-14B (w/ think)      & 0.762 & 0.800 & 0.795 & 0.708 & 0.719 & 0.707 & 0.771 & 0.775 & 0.754 & 0.744 & 0.835 & 0.750 & 0.760 \\
Qwen3-32B (w/o think)   & 0.767 & 0.815 & 0.816 & 0.714 & 0.740 & 0.719 & 0.792 & 0.793 & 0.759 & 0.750 & 0.846 & 0.759 & 0.772 \\
Qwen3-32B (w/ think)      & 0.768 & 0.802 & 0.801 & 0.711 & 0.727 & 0.715 & 0.775 & 0.777 & 0.755 & 0.746 & 0.839 & 0.754 & 0.764 \\

DeepSeek-R1            & 0.773 & 0.825 & 0.811 & 0.719 & 0.736 & 0.728 & 0.787 & 0.788 & 0.746 & 0.748 & 0.843 & 0.776 & 0.773 \\
DeepSeek-V3.2          & 0.782 & 0.820 & 0.808 & 0.725 & 0.740 & 0.729 & 0.802 & 0.794 & 0.760 & 0.756 & 0.846 & 0.770 & 0.778 \\

\midrule
\rowcolor{gray!20}
\multicolumn{14}{c}{\textbf{Specialized MT Models}} \\

Seed-X-PPO-7B (w/ think)         & 0.781 & 0.814 & 0.712 & 0.689 & 0.687 & 0.706 & 0.785 & 0.781 & 0.675 & 0.755 & 0.847 & 0.776 & 0.751 \\
Seed-X-PPO-7B (w/o think) & 0.779 & 0.813 & 0.822 & 0.708 & 0.749 & 0.714 & 0.782 & 0.784 & 0.771 & 0.762 & 0.844 & 0.783 & 0.776 \\
NLLB-200-3.3B          & 0.711 & 0.607 & 0.592 & 0.666 & 0.567 & 0.605 & 0.583 & 0.640 & 0.663 & 0.636 & 0.668 & 0.664 & 0.633 \\
LLaMAX3-8B-Alpaca      & 0.725 & 0.746 & 0.742 & 0.686 & 0.663 & 0.684 & 0.728 & 0.723 & 0.726 & 0.713 & 0.782 & 0.734 & 0.721 \\
\midrule
\rowcolor{gray!20}
\multicolumn{14}{c}{\textbf{Production Systems}} \\

Google Translate       & 0.704 & 0.803 & 0.782 & 0.672 & 0.667 & 0.688 & 0.764 & 0.763 & 0.757 & 0.704 & 0.825 & 0.766 & 0.741 \\
Youdao Translate       & 0.660 & 0.711 & 0.797 & 0.670 & 0.622 & 0.648 & 0.787 & 0.667 & 0.752 & 0.635 & 0.800 & 0.673 & 0.702 \\

\bottomrule
\end{tabular}
}
\caption{Comparison of COMET scores across different models and translation directions. Both reasoning and non-reasoning variants are reported when available.}
\label{tab:comet_results}
\end{table*}

Overall, the COMET scores exhibit trends broadly consistent with our multi-dimensional evaluation, with stronger models achieving higher scores across most language pairs. However, we also observe that COMET primarily reflects semantic adequacy and may not fully capture cultural adaptation, highlighting the complementary role of our proposed evaluation framework.

\subsection{Dimension-level Evaluation Results}
\label{app:Dimension-level-result}
To provide a more complete view of model behavior under our evaluation framework, this appendix reports the dimension-level scores for all evaluated models across all language directions. While the main paper focuses on overall scores for clarity and comparability, we include detailed results here to demonstrate the full evaluation coverage and to support further inspection. The complete results are presented in Tables~\ref{tab:fine_grained_results_1}--\ref{tab:fine_grained_results_4}.

\begin{table*}[t]
\centering
\small
\resizebox{\textwidth}{!}{
\begin{tabular}{lcccccccccccccccc}
\toprule
\multirow{2}{*}{\textbf{Model}} & \multicolumn{3}{c}{\textbf{Contextual Accuracy}} & 
\multicolumn{3}{c}{\textbf{Cultural Adaptation}} & 
\multicolumn{3}{c}{\textbf{Functional Equivalence}} & 
\multicolumn{3}{c}{\textbf{Fidelity}} & 
\multicolumn{3}{c}{\textbf{Naturalness}} \\
\cmidrule(lr){2-4} \cmidrule(lr){5-7} \cmidrule(lr){8-10} \cmidrule(lr){11-13} \cmidrule(lr){14-16}
 & \textbf{En $\rightarrow$ Es} & \textbf{En $\rightarrow$ Ja} & \textbf{En $\rightarrow$ Zh} & 
   \textbf{En $\rightarrow$ Es} & \textbf{En $\rightarrow$ Ja} & \textbf{En $\rightarrow$ Zh} & 
   \textbf{En $\rightarrow$ Es} & \textbf{En $\rightarrow$ Ja} & \textbf{En $\rightarrow$ Zh} & 
   \textbf{En $\rightarrow$ Es} & \textbf{En $\rightarrow$ Ja} & \textbf{En $\rightarrow$ Zh} & 
   \textbf{En $\rightarrow$ Es} & \textbf{En $\rightarrow$ Ja} & \textbf{En $\rightarrow$ Zh} \\

\midrule
\multicolumn{16}{l}{\textbf{Proprietary LLMs}} \\
GPT-4                       & 4.93 & 5.32 & 5.50 & 5.28 & 5.49 & 5.65 & 4.68 & 5.13 & 5.18 & 4.91 & 5.20 & 5.41 & 4.59 & 4.66 & 4.88 \\
GPT-4o                      & 4.69 & 5.31 & 5.65 & 5.11 & 5.32 & 5.57 & 4.57 & 4.84 & 5.05 & 5.00 & 5.14 & 5.15 & 4.35 & 4.35 & 4.64 \\
Gemini-2.5-Flash-Lite       & 5.00 & 5.26 & 5.39 & 5.31 & 5.34 & 5.50 & 4.75 & 4.91 & 4.97 & 5.22 & 5.25 & 4.97 & 4.59 & 4.64 & 4.70 \\
Grok-4.1                    & 5.06 & 5.62 & 5.47 & 5.28 & 5.76 & 5.44 & 4.77 & 5.24 & 5.16 & 5.26 & 5.46 & 5.36 & 4.57 & 4.60 & 4.70 \\
\midrule
\multicolumn{16}{l}{\textbf{Open-source LLMs}} \\
LLaMA-3-8B-Instruct-262k    & 4.65 & 3.42 & 4.09 & 5.01 & 3.98 & 4.48 & 4.15 & 3.21 & 3.73 & 4.25 & 3.24 & 3.57 & 3.95 & 2.95 & 3.59 \\
LLaMA-3.3-70B-Instruct      & 4.89 & 5.36 & 5.08 & 5.30 & 5.17 & 5.30 & 4.61 & 4.53 & 4.93 & 5.11 & 4.92 & 5.05 & 4.44 & 4.18 & 4.63 \\
\addlinespace
Mixtral-8x7B-Instruct-v0.1  & 4.47 & 3.26 & 3.65 & 4.95 & 3.70 & 3.86 & 4.25 & 2.69 & 2.74 & 4.45 & 2.81 & 2.76 & 4.23 & 2.53 & 2.73 \\
\addlinespace
Qwen2.5-7B-Instruct         & 4.33 & 3.79 & 4.78 & 4.68 & 4.15 & 5.23 & 3.92 & 3.22 & 4.44 & 4.06 & 3.37 & 4.54 & 3.66 & 3.07 & 4.30 \\
Qwen2.5-14B-Instruct        & 4.64 & 4.28 & 5.23 & 5.01 & 4.50 & 5.34 & 4.23 & 3.95 & 4.95 & 4.29 & 3.98 & 4.90 & 4.12 & 3.67 & 4.59 \\
Qwen2.5-32B-Instruct        & 4.77 & 4.57 & 5.30 & 5.18 & 4.93 & 5.42 & 4.52 & 4.10 & 4.95 & 4.78 & 4.50 & 5.11 & 4.28 & 3.88 & 4.51 \\
Qwen2.5-72B-Instruct        & 5.08 & 5.06 & 5.45 & 5.12 & 5.37 & 5.54 & 4.67 & 4.65 & 5.03 & 5.01 & 4.83 & 5.30 & 4.57 & 4.27 & 4.87 \\
\addlinespace
Qwen3-4B (w/o think)         & 4.08 & 3.99 & 4.78 & 4.63 & 4.21 & 4.96 & 3.75 & 3.47 & 4.43 & 3.89 & 3.70 & 4.43 & 3.70 & 3.43 & 4.26 \\
Qwen3-4B (w/ think)            & 4.11 & 4.46 & 4.94 & 4.68 & 4.52 & 5.14 & 3.79 & 3.89 & 4.54 & 3.92 & 4.00 & 4.69 & 3.76 & 3.44 & 4.32 \\
Qwen3-8B (w/o think)         & 4.78 & 4.54 & 5.05 & 5.05 & 4.73 & 5.27 & 4.08 & 4.17 & 4.80 & 4.38 & 4.11 & 4.89 & 4.08 & 3.69 & 4.53 \\
Qwen3-8B (w/ think)            & 4.78 & 4.69 & 5.14 & 5.03 & 4.71 & 5.38 & 4.47 & 4.18 & 5.01 & 4.66 & 4.21 & 5.04 & 4.34 & 3.72 & 4.67 \\
Qwen3-14B (w/o think)        & 5.02 & 4.90 & 5.14 & 5.15 & 5.05 & 5.38 & 4.46 & 4.56 & 4.95 & 4.60 & 4.62 & 5.20 & 4.23 & 4.12 & 4.68 \\
Qwen3-14B (w/ think)           & 4.89 & 4.81 & 5.06 & 5.18 & 4.83 & 5.29 & 4.59 & 4.49 & 5.06 & 4.79 & 4.64 & 5.04 & 4.34 & 3.89 & 4.64 \\
Qwen3-32B (w/o think)        & 4.71 & 4.72 & 5.47 & 5.10 & 4.66 & 5.52 & 4.48 & 4.53 & 4.95 & 4.80 & 4.44 & 4.93 & 4.21 & 4.01 & 4.71 \\
Qwen3-32B (w/ think)           & 4.99 & 4.91 & 5.31 & 5.21 & 5.01 & 5.41 & 4.60 & 4.67 & 5.01 & 4.92 & 4.82 & 5.23 & 4.39 & 4.12 & 4.57 \\
\addlinespace
DeepSeek-R1                 & 5.09 & 5.30 & 5.49 & 5.33 & 5.31 & 5.57 & 4.87 & 5.06 & 5.20 & 5.08 & 5.09 & 5.12 & 4.66 & 4.62 & 4.85 \\
DeepSeek-V3.2               & 5.23 & 5.62 & 5.55 & 5.29 & 5.55 & 5.82 & 4.91 & 5.04 & 5.15 & 5.11 & 5.09 & 5.14 & 4.74 & 4.75 & 4.87 \\
\midrule
\multicolumn{16}{l}{\textbf{Specialized MT Models}} \\
Seed-X-PPO-7B (w/o think)     & 4.79 & 5.01 & 5.15 & 5.00 & 4.98 & 5.47 & 4.49 & 4.56 & 5.05 & 5.09 & 4.74 & 5.03 & 4.53 & 4.15 & 4.98 \\
Seed-X-PPO-7B (w/ think)    & 4.86 & 5.04 & 5.61 & 5.35 & 5.15 & 5.77 & 4.72 & 4.51 & 5.42 & 4.92 & 4.80 & 5.31 & 4.52 & 4.30 & 5.24 \\
NLLB-200-3.3B                & 4.10 & 3.48 & 3.34 & 4.62 & 4.09 & 3.85 & 3.65 & 2.84 & 2.78 & 4.11 & 2.75 & 2.82 & 3.72 & 2.61 & 2.55 \\
LLaMAX3-8B-Alpaca           & 4.08 & 4.17 & 4.28 & 4.81 & 4.45 & 4.57 & 3.62 & 3.66 & 3.97 & 4.09 & 3.70 & 3.99 & 3.76 & 3.50 & 3.77 \\

\midrule
\multicolumn{16}{l}{\textbf{Production Systems}} \\
Google Translate            & 4.76 & 5.08 & 4.99 & 5.09 & 5.22 & 5.11 & 4.10 & 4.20 & 4.54 & 4.32 & 4.62 & 4.61 & 3.98 & 4.09 & 4.24 \\
Youdao Translate            & 3.83 & 3.93 & 5.10 & 4.22 & 4.38 & 5.26 & 3.38 & 3.48 & 4.72 & 3.40 & 3.54 & 4.84 & 3.14 & 3.27 & 4.52 \\
\bottomrule
\end{tabular}
}
\caption{Fine-grained evaluation of translation quality across multiple dimensions for En$\rightarrow$Es, En$\rightarrow$Ja, and En$\rightarrow$Zh directions.}
\label{tab:fine_grained_results_1}
\end{table*}

\begin{table*}[t]
\centering
\small
\resizebox{\textwidth}{!}{
\begin{tabular}{lcccccccccccccccc}
\toprule
\multirow{2}{*}{\textbf{Model}} & \multicolumn{3}{c}{\textbf{Contextual Accuracy}} & 
\multicolumn{3}{c}{\textbf{Cultural Adaptation}} & 
\multicolumn{3}{c}{\textbf{Functional Equivalence}} & 
\multicolumn{3}{c}{\textbf{Fidelity}} & 
\multicolumn{3}{c}{\textbf{Naturalness}} \\
\cmidrule(lr){2-4} \cmidrule(lr){5-7} \cmidrule(lr){8-10} \cmidrule(lr){11-13} \cmidrule(lr){14-16}
 & \textbf{Es $\rightarrow$ En} & \textbf{Es $\rightarrow$ Zh} & \textbf{Ja $\rightarrow$ En} & 
   \textbf{Es $\rightarrow$ En} & \textbf{Es $\rightarrow$ Zh} & \textbf{Ja $\rightarrow$ En} & 
   \textbf{Es $\rightarrow$ En} & \textbf{Es $\rightarrow$ Zh} & \textbf{Ja $\rightarrow$ En} & 
   \textbf{Es $\rightarrow$ En} & \textbf{Es $\rightarrow$ Zh} & \textbf{Ja $\rightarrow$ En} & 
   \textbf{Es $\rightarrow$ En} & \textbf{Es $\rightarrow$ Zh} & \textbf{Ja $\rightarrow$ En} \\

\midrule
\multicolumn{16}{l}{\textbf{Proprietary LLMs}} \\

GPT-4                       & 5.31 & 4.92 & 5.34 & 5.44 & 4.89 & 5.45 & 5.26 & 5.11 & 5.05 & 5.27 & 4.89 & 5.29 & 4.95 & 4.75 & 4.78 \\
GPT-4o                      & 5.16 & 4.72 & 5.27 & 5.36 & 4.93 & 5.26 & 5.01 & 4.67 & 4.76 & 5.16 & 4.71 & 5.13 & 4.66 & 4.27 & 4.74 \\
Gemini-2.5-Flash-Lite       & 5.28 & 4.48 & 5.21 & 5.34 & 4.88 & 5.42 & 5.02 & 4.77 & 4.67 & 4.99 & 4.51 & 4.92 & 4.64 & 4.50 & 4.55 \\
Grok-4.1                    & 5.46 & 4.97 & 5.35 & 5.31 & 5.04 & 5.53 & 5.19 & 5.28 & 4.79 & 5.34 & 5.20 & 4.91 & 4.81 & 4.66 & 4.59 \\
\midrule

\multicolumn{16}{l}{\textbf{Open-source LLMs}} \\
LLaMA-3-8B-Instruct-262k    & 4.25 & 2.98 & 3.19 & 4.71 & 3.68 & 3.80 & 4.06 & 3.23 & 3.27 & 3.87 & 2.98 & 3.23 & 4.08 & 3.10 & 3.53 \\
LLaMA-3.3-70B-Instruct      & 5.12 & 4.47 & 4.39 & 5.36 & 4.72 & 4.84 & 4.92 & 4.74 & 4.28 & 5.07 & 4.54 & 4.39 & 4.72 & 4.40 & 4.29 \\
\addlinespace
Mixtral-8x7B-Instruct-v0.1  & 5.01 & 3.01 & 3.30 & 5.18 & 3.69 & 4.03 & 4.61 & 2.68 & 3.38 & 4.83 & 2.55 & 3.18 & 4.40 & 2.59 & 3.64 \\
\addlinespace
Qwen2.5-7B-Instruct         & 4.67 & 3.86 & 3.78 & 4.93 & 4.37 & 4.11 & 4.66 & 4.07 & 3.84 & 4.54 & 3.98 & 3.59 & 4.25 & 3.91 & 3.78 \\
Qwen2.5-14B-Instruct        & 4.90 & 4.28 & 4.24 & 5.17 & 4.62 & 4.61 & 4.91 & 4.68 & 4.15 & 4.88 & 4.60 & 4.25 & 4.61 & 4.32 & 4.35 \\
Qwen2.5-32B-Instruct        & 5.02 & 4.49 & 4.73 & 5.31 & 4.81 & 4.87 & 5.05 & 4.86 & 4.42 & 4.99 & 4.65 & 4.46 & 4.57 & 4.41 & 4.52 \\
Qwen2.5-72B-Instruct        & 4.93 & 4.58 & 4.84 & 5.22 & 4.80 & 5.07 & 5.09 & 4.91 & 4.47 & 4.95 & 4.70 & 4.49 & 4.84 & 4.60 & 4.49 \\
\addlinespace
Qwen3-4B (w/o think)         & 4.54 & 4.02 & 3.39 & 4.90 & 4.27 & 3.93 & 4.34 & 4.31 & 3.51 & 4.30 & 3.98 & 3.35 & 4.10 & 3.92 & 3.44 \\
Qwen3-4B (w/ think)            & 4.60 & 4.17 & 3.45 & 4.81 & 4.63 & 3.91 & 4.30 & 4.28 & 3.47 & 4.28 & 4.19 & 3.45 & 4.03 & 3.98 & 3.47 \\
Qwen3-8B (w/o think)         & 4.72 & 4.26 & 3.80 & 4.77 & 4.70 & 4.23 & 4.49 & 4.70 & 4.02 & 4.42 & 4.37 & 3.90 & 4.13 & 4.45 & 4.15 \\
Qwen3-8B (w/ think)            & 4.63 & 4.27 & 3.97 & 4.86 & 4.66 & 4.46 & 4.59 & 4.75 & 4.18 & 4.72 & 4.64 & 4.12 & 4.28 & 4.25 & 4.13 \\
Qwen3-14B (w/o think)        & 5.04 & 4.49 & 3.94 & 5.13 & 4.87 & 4.54 & 4.80 & 4.74 & 4.28 & 4.81 & 4.54 & 4.17 & 4.48 & 4.43 & 4.28 \\
Qwen3-14B (w/ think)           & 4.81 & 4.75 & 4.41 & 5.20 & 4.78 & 4.72 & 4.55 & 5.02 & 4.26 & 4.84 & 4.81 & 4.43 & 4.44 & 4.51 & 4.12 \\
Qwen3-32B (w/o think)        & 5.05 & 4.46 & 4.23 & 5.14 & 4.87 & 4.78 & 4.92 & 4.80 & 4.39 & 4.90 & 4.54 & 4.25 & 4.60 & 4.54 & 4.29 \\
Qwen3-32B (w/ think)           & 4.99 & 4.63 & 4.37 & 5.31 & 4.87 & 4.79 & 5.04 & 4.88 & 4.23 & 4.98 & 4.60 & 4.44 & 4.69 & 4.55 & 4.35 \\
\addlinespace
DeepSeek-R1                 & 5.27 & 4.34 & 5.18 & 5.49 & 4.93 & 5.28 & 5.13 & 4.78 & 4.66 & 5.11 & 4.52 & 5.07 & 4.80 & 4.67 & 4.67 \\
DeepSeek-V3.2               & 5.24 & 4.64 & 5.19 & 5.50 & 4.90 & 5.34 & 5.01 & 5.12 & 4.51 & 5.05 & 4.69 & 4.81 & 4.72 & 4.73 & 4.57 \\
\midrule
\multicolumn{16}{l}{\textbf{Specialized MT Models}} \\
Seed-X-PPO-7B (w/o think)     & 4.81 & 4.33 & 3.97 & 5.13 & 4.81 & 4.43 & 5.00 & 4.74 & 4.22 & 4.72 & 4.36 & 4.05 & 4.63 & 4.78 & 4.41 \\
Seed-X-PPO-7B (w/ think)    & 5.14 & 4.59 & 4.17 & 5.28 & 4.98 & 4.50 & 4.95 & 5.11 & 4.22 & 4.75 & 4.55 & 4.07 & 4.64 & 5.01 & 4.40 \\
NLLB-200-3.3B                & 4.07 & 2.51 & 2.34 & 4.46 & 3.38 & 2.93 & 3.82 & 2.85 & 2.61 & 3.80 & 2.80 & 2.49 & 3.64 & 2.63 & 2.98 \\
LLaMAX3-8B-Alpaca           & 4.38 & 3.51 & 3.46 & 4.72 & 3.96 & 4.03 & 4.57 & 3.72 & 3.65 & 4.41 & 3.39 & 3.48 & 4.19 & 3.46 & 3.89 \\

\midrule
\multicolumn{16}{l}{\textbf{Production Systems}} \\
Google Translate            & 4.90 & 3.99 & 4.46 & 5.19 & 4.34 & 4.85 & 4.34 & 4.14 & 4.09 & 4.37 & 4.13 & 4.24 & 3.90 & 3.70 & 4.14 \\
Youdao Translate            & 4.30 & 3.55 & 3.81 & 4.59 & 3.91 & 4.13 & 3.94 & 3.36 & 3.40 & 4.07 & 3.31 & 3.38 & 3.64 & 3.15 & 3.34 \\
\bottomrule
\end{tabular}
}
\caption{Fine-grained evaluation of translation quality across multiple dimensions for Es$\rightarrow$En, Es$\rightarrow$Zh, and Ja$\rightarrow$En directions.}
\label{tab:fine_grained_results_2}
\end{table*}

\begin{table*}[t]
\centering
\small
\resizebox{\textwidth}{!}{
\begin{tabular}{lcccccccccccccccc}
\toprule
\multirow{2}{*}{\textbf{Model}} & \multicolumn{3}{c}{\textbf{Contextual Accuracy}} & 
\multicolumn{3}{c}{\textbf{Cultural Adaptation}} & 
\multicolumn{3}{c}{\textbf{Functional Equivalence}} & 
\multicolumn{3}{c}{\textbf{Fidelity}} & 
\multicolumn{3}{c}{\textbf{Naturalness}} \\
\cmidrule(lr){2-4} \cmidrule(lr){5-7} \cmidrule(lr){8-10} \cmidrule(lr){11-13} \cmidrule(lr){14-16}
 & \textbf{Ja $\rightarrow$ Zh} & \textbf{Ru $\rightarrow$ Zh} & \textbf{Zh $\rightarrow$ En} & 
   \textbf{Ja $\rightarrow$ Zh} & \textbf{Ru $\rightarrow$ Zh} & \textbf{Zh $\rightarrow$ En} & 
   \textbf{Ja $\rightarrow$ Zh} & \textbf{Ru $\rightarrow$ Zh} & \textbf{Zh $\rightarrow$ En} & 
   \textbf{Ja $\rightarrow$ Zh} & \textbf{Ru $\rightarrow$ Zh} & \textbf{Zh $\rightarrow$ En} & 
   \textbf{Ja $\rightarrow$ Zh} & \textbf{Ru $\rightarrow$ Zh} & \textbf{Zh $\rightarrow$ En} \\

\midrule
\multicolumn{16}{l}{\textbf{Proprietary LLMs}} \\
GPT-4                       & 5.22 & 4.92 & 5.78 & 5.42 & 5.10 & 5.65 & 5.16 & 4.70 & 5.42 & 5.47 & 4.83 & 5.59 & 4.87 & 4.38 & 5.14 \\
GPT-4o                      & 4.96 & 4.66 & 5.38 & 5.09 & 5.00 & 5.57 & 4.91 & 4.77 & 5.25 & 5.11 & 4.81 & 5.52 & 4.58 & 4.50 & 5.02 \\
Gemini-2.5-Flash-Lite       & 4.58 & 4.75 & 5.54 & 4.93 & 4.99 & 5.54 & 4.71 & 4.98 & 5.37 & 4.93 & 4.87 & 5.61 & 4.44 & 4.62 & 4.91 \\
Grok-4.1                    & 5.40 & 5.00 & 5.82 & 5.43 & 5.16 & 5.52 & 5.01 & 5.10 & 5.34 & 5.34 & 5.06 & 5.61 & 4.68 & 4.66 & 4.87 \\
\midrule

\multicolumn{16}{l}{\textbf{Open-source LLMs}} \\
LLaMA-3-8B-Instruct-262k    & 3.22 & 3.42 & 4.30 & 3.98 & 3.89 & 4.54 & 3.12 & 3.48 & 4.33 & 3.16 & 3.27 & 4.18 & 2.96 & 3.44 & 4.33 \\
LLaMA-3.3-70B-Instruct      & 4.32 & 4.51 & 5.26 & 4.77 & 4.75 & 5.23 & 4.31 & 4.64 & 5.08 & 4.48 & 4.54 & 5.30 & 4.27 & 4.61 & 4.86 \\
\addlinespace
Mixtral-8x7B-Instruct-v0.1  & 3.12 & 3.07 & 4.63 & 3.53 & 3.59 & 4.77 & 3.07 & 2.75 & 4.75 & 2.96 & 2.74 & 4.70 & 3.08 & 2.66 & 4.53 \\
\addlinespace
Qwen2.5-7B-Instruct         & 4.25 & 4.30 & 5.19 & 4.55 & 4.69 & 5.19 & 4.29 & 4.29 & 4.94 & 4.26 & 4.19 & 4.97 & 4.04 & 4.06 & 4.58 \\
Qwen2.5-14B-Instruct        & 4.42 & 4.58 & 5.47 & 4.67 & 4.86 & 5.45 & 4.79 & 4.42 & 5.23 & 4.80 & 4.54 & 5.26 & 4.60 & 4.54 & 4.90 \\
Qwen2.5-32B-Instruct        & 4.63 & 4.66 & 5.31 & 5.08 & 4.94 & 5.43 & 4.92 & 4.70 & 5.14 & 4.94 & 4.61 & 5.31 & 4.53 & 4.56 & 4.92 \\
Qwen2.5-72B-Instruct        & 4.95 & 4.73 & 5.49 & 5.07 & 5.04 & 5.58 & 4.92 & 4.90 & 5.34 & 5.03 & 4.95 & 5.34 & 4.66 & 4.65 & 5.10 \\
\addlinespace
Qwen3-4B (w/o think)         & 4.18 & 4.04 & 4.87 & 4.78 & 4.49 & 4.95 & 4.22 & 4.30 & 4.81 & 4.04 & 4.06 & 4.88 & 3.97 & 4.30 & 4.62 \\
Qwen3-4B (w/ think)            & 4.35 & 4.33 & 5.02 & 4.83 & 4.42 & 5.21 & 4.29 & 4.42 & 5.02 & 4.24 & 4.41 & 4.99 & 4.04 & 4.17 & 4.72 \\
Qwen3-8B (w/o think)         & 4.42 & 4.48 & 5.10 & 5.09 & 4.82 & 5.22 & 4.58 & 4.68 & 5.13 & 4.75 & 4.51 & 5.10 & 4.42 & 4.58 & 4.58 \\
Qwen3-8B (w/ think)            & 4.78 & 4.49 & 5.22 & 5.16 & 4.91 & 5.25 & 4.55 & 4.60 & 5.20 & 4.68 & 4.64 & 5.34 & 4.36 & 4.41 & 4.86 \\
Qwen3-14B (w/o think)        & 4.74 & 4.84 & 5.51 & 4.92 & 4.99 & 5.41 & 4.67 & 4.89 & 5.34 & 4.73 & 4.78 & 5.31 & 4.58 & 4.63 & 4.96 \\
Qwen3-14B (w/ think)           & 4.78 & 4.74 & 5.50 & 5.14 & 5.07 & 5.45 & 4.75 & 4.85 & 5.26 & 4.95 & 5.02 & 5.40 & 4.52 & 4.62 & 4.91 \\
Qwen3-32B (w/o think)        & 4.62 & 4.68 & 5.30 & 5.01 & 5.16 & 5.32 & 4.88 & 4.89 & 5.24 & 4.72 & 4.74 & 5.23 & 4.61 & 4.66 & 4.88 \\
Qwen3-32B (w/ think)           & 4.73 & 4.93 & 5.53 & 4.92 & 5.11 & 5.45 & 4.78 & 4.88 & 5.31 & 4.98 & 5.02 & 5.58 & 4.51 & 4.63 & 4.89 \\
\addlinespace
DeepSeek-R1                 & 4.78 & 4.54 & 5.45 & 5.13 & 4.82 & 5.52 & 4.99 & 4.87 & 5.23 & 4.77 & 4.57 & 5.30 & 4.54 & 4.73 & 4.90 \\
DeepSeek-V3.2               & 5.08 & 4.86 & 5.49 & 5.21 & 4.99 & 5.50 & 5.11 & 4.93 & 5.45 & 5.08 & 4.76 & 5.42 & 4.85 & 4.75 & 5.20 \\
\midrule
\multicolumn{16}{l}{\textbf{Specialized MT Models}} \\
Seed-X-PPO-7B (w/o think)     & 3.86 & 4.44 & 5.59 & 4.54 & 4.90 & 5.49 & 4.42 & 4.72 & 5.25 & 4.03 & 4.44 & 5.42 & 4.56 & 4.93 & 5.09 \\
Seed-X-PPO-7B (w/ think)    & 4.17 & 4.53 & 5.79 & 4.57 & 4.82 & 5.66 & 4.29 & 4.82 & 5.34 & 4.08 & 4.46 & 5.39 & 4.75 & 4.91 & 5.09 \\
NLLB-200-3.3B                & 2.33 & 3.06 & 3.06 & 3.04 & 3.55 & 3.57 & 2.32 & 3.32 & 3.12 & 2.38 & 3.31 & 3.04 & 2.40 & 2.98 & 3.19 \\
LLaMAX3-8B-Alpaca           & 3.75 & 3.62 & 4.38 & 4.42 & 4.06 & 4.67 & 3.92 & 3.90 & 4.48 & 3.84 & 3.66 & 4.39 & 3.77 & 3.82 & 4.38 \\

\midrule
\multicolumn{16}{l}{\textbf{Production Systems}} \\
Google Translate            & 3.86 & 4.32 & 5.23 & 4.46 & 4.58 & 5.06 & 4.09 & 4.70 & 4.94 & 3.78 & 4.71 & 5.19 & 3.90 & 4.38 & 4.55 \\
Youdao Translate            & 4.83 & 3.06 & 5.30 & 4.95 & 3.65 & 5.31 & 4.61 & 3.36 & 4.98 & 4.69 & 3.23 & 5.34 & 4.28 & 3.12 & 4.77 \\
\bottomrule
\end{tabular}
}
\caption{Fine-grained evaluation of translation quality across multiple dimensions for Ja$\rightarrow$Zh, Ru$\rightarrow$Zh, and Zh$\rightarrow$En directions.}
\label{tab:fine_grained_results_3}
\end{table*}

\begin{table*}[t]
\centering
\small
\resizebox{\textwidth}{!}{
\begin{tabular}{lcccccccccccccccc}
\toprule
\multirow{2}{*}{\textbf{Model}} & \multicolumn{3}{c}{\textbf{Contextual Accuracy}} & 
\multicolumn{3}{c}{\textbf{Cultural Adaptation}} & 
\multicolumn{3}{c}{\textbf{Functional Equivalence}} & 
\multicolumn{3}{c}{\textbf{Fidelity}} & 
\multicolumn{3}{c}{\textbf{Naturalness}} \\
\cmidrule(lr){2-4} \cmidrule(lr){5-7} \cmidrule(lr){8-10} \cmidrule(lr){11-13} \cmidrule(lr){14-16}
 & \textbf{Zh $\rightarrow$ Es} & \textbf{Zh $\rightarrow$ Ja} & \textbf{Zh $\rightarrow$ Ru} & 
   \textbf{Zh $\rightarrow$ Es} & \textbf{Zh $\rightarrow$ Ja} & \textbf{Zh $\rightarrow$ Ru} & 
   \textbf{Zh $\rightarrow$ Es} & \textbf{Zh $\rightarrow$ Ja} & \textbf{Zh $\rightarrow$ Ru} & 
   \textbf{Zh $\rightarrow$ Es} & \textbf{Zh $\rightarrow$ Ja} & \textbf{Zh $\rightarrow$ Ru} & 
   \textbf{Zh $\rightarrow$ Es} & \textbf{Zh $\rightarrow$ Ja} & \textbf{Zh $\rightarrow$ Ru} \\

\midrule
\multicolumn{16}{l}{\textbf{Proprietary LLMs}} \\

GPT-4                       & 5.29 & 5.31 & 4.99 & 5.39 & 5.35 & 4.90 & 5.15 & 5.36 & 5.16 & 5.29 & 5.59 & 5.01 & 4.90 & 5.09 & 4.36 \\
GPT-4o                      & 4.84 & 5.11 & 4.61 & 4.98 & 5.06 & 4.80 & 4.93 & 5.10 & 4.88 & 5.11 & 5.42 & 4.87 & 4.73 & 4.88 & 4.58 \\
Gemini-2.5-Flash-Lite       & 5.26 & 5.02 & 4.72 & 5.29 & 5.03 & 5.04 & 5.11 & 5.23 & 5.11 & 5.27 & 5.26 & 5.16 & 4.86 & 4.79 & 4.84 \\
Grok-4.1                    & 5.42 & 5.33 & 5.27 & 5.36 & 5.15 & 5.31 & 5.27 & 5.18 & 5.49 & 5.40 & 5.50 & 5.49 & 4.80 & 4.69 & 4.93 \\

\midrule

\multicolumn{16}{l}{\textbf{Open-source LLMs}} \\
LLaMA-3-8B-Instruct-262k    & 3.46 & 3.28 & 3.04 & 4.03 & 3.91 & 3.63 & 3.90 & 3.31 & 3.41 & 3.71 & 3.39 & 3.32 & 3.94 & 2.99 & 3.18 \\
LLaMA-3.3-70B-Instruct      & 4.75 & 4.90 & 4.68 & 4.89 & 4.97 & 4.99 & 4.79 & 4.91 & 4.87 & 4.93 & 4.90 & 4.84 & 4.63 & 4.35 & 4.60 \\
\addlinespace
Mixtral-8x7B-Instruct-v0.1  & 4.02 & 3.28 & 3.71 & 4.43 & 3.67 & 4.05 & 4.34 & 3.02 & 3.76 & 4.10 & 2.96 & 3.72 & 4.08 & 2.66 & 3.41 \\
\addlinespace
Qwen2.5-7B-Instruct         & 3.85 & 3.59 & 3.15 & 4.17 & 3.99 & 3.77 & 4.01 & 3.67 & 3.35 & 3.79 & 3.53 & 3.14 & 3.83 & 3.45 & 3.09 \\
Qwen2.5-14B-Instruct        & 4.51 & 4.42 & 3.41 & 4.64 & 4.51 & 4.02 & 4.64 & 4.11 & 3.80 & 4.49 & 4.21 & 3.61 & 4.38 & 3.68 & 3.44 \\
Qwen2.5-32B-Instruct        & 4.85 & 4.32 & 4.13 & 4.99 & 4.69 & 4.40 & 4.68 & 4.42 & 3.97 & 4.86 & 4.42 & 4.03 & 4.32 & 3.89 & 3.73 \\
Qwen2.5-72B-Instruct        & 5.02 & 4.70 & 4.60 & 5.08 & 4.79 & 4.79 & 5.05 & 4.91 & 4.74 & 5.15 & 4.92 & 4.94 & 4.82 & 4.39 & 4.44 \\
\addlinespace
Qwen3-4B (w/o think)        & 3.63 & 3.82 & 3.27 & 3.93 & 4.17 & 3.65 & 4.01 & 3.87 & 3.59 & 3.76 & 4.03 & 3.40 & 3.77 & 3.54 & 3.38 \\
Qwen3-4B (w/ think)            & 3.99 & 4.01 & 3.49 & 4.36 & 4.34 & 3.97 & 4.36 & 4.20 & 4.15 & 4.36 & 4.39 & 4.07 & 4.06 & 3.68 & 3.64 \\
Qwen3-8B (w/o think)         & 4.29 & 4.29 & 3.72 & 4.62 & 4.56 & 3.95 & 4.47 & 4.47 & 4.39 & 4.32 & 4.67 & 4.18 & 4.21 & 4.12 & 4.01 \\
Qwen3-8B (w/ think)            & 4.66 & 4.62 & 4.06 & 4.90 & 4.61 & 4.48 & 4.89 & 4.86 & 4.53 & 5.05 & 4.88 & 4.45 & 4.47 & 4.11 & 4.21 \\
Qwen3-14B (w/o think)        & 4.67 & 4.65 & 4.04 & 4.77 & 4.77 & 4.43 & 4.89 & 4.82 & 4.47 & 4.90 & 5.01 & 4.20 & 4.66 & 4.39 & 4.18 \\
Qwen3-14B (w/ think)           & 5.02 & 4.80 & 4.56 & 5.09 & 4.85 & 4.82 & 4.92 & 5.17 & 4.91 & 5.14 & 5.34 & 4.84 & 4.64 & 4.45 & 4.57 \\
Qwen3-32B (w/o think)        & 4.65 & 4.83 & 4.06 & 4.89 & 4.92 & 4.51 & 4.64 & 5.01 & 4.60 & 4.77 & 5.10 & 4.26 & 4.61 & 4.46 & 4.24 \\
Qwen3-32B (w/ think)           & 5.28 & 5.00 & 4.68 & 5.19 & 5.06 & 5.00 & 4.85 & 5.15 & 4.95 & 5.10 & 5.23 & 5.01 & 4.73 & 4.59 & 4.35 \\
\addlinespace
DeepSeek-R1                 & 5.08 & 5.09 & 5.25 & 5.14 & 5.27 & 5.42 & 5.11 & 5.16 & 5.28 & 5.09 & 5.20 & 5.28 & 4.84 & 4.63 & 4.86 \\
DeepSeek-V3.2               & 5.03 & 5.15 & 4.92 & 5.16 & 5.20 & 5.18 & 5.21 & 5.28 & 5.15 & 5.24 & 5.33 & 5.05 & 4.88 & 4.82 & 4.69 \\
\midrule
\multicolumn{16}{l}{\textbf{Specialized MT Models}} \\
Seed-X-PPO-7B (w/o think)     & 4.88 & 4.35 & 4.91 & 5.12 & 4.57 & 5.06 & 5.10 & 4.55 & 5.16 & 5.10 & 4.53 & 4.86 & 5.13 & 4.24 & 4.82 \\
Seed-X-PPO-7B (w/ think)    & 4.85 & 4.18 & 4.89 & 5.01 & 4.50 & 5.11 & 5.17 & 4.55 & 5.04 & 5.19 & 4.65 & 4.93 & 5.05 & 4.45 & 4.77 \\
NLLB-200-3.3B                & 2.62 & 2.33 & 2.52 & 3.24 & 3.22 & 3.17 & 3.00 & 2.73 & 2.92 & 2.75 & 2.67 & 2.69 & 3.04 & 2.61 & 3.02 \\
LLaMAX3-8B-Alpaca           & 3.49 & 3.60 & 3.31 & 4.03 & 4.01 & 3.71 & 4.09 & 3.87 & 4.08 & 3.73 & 3.83 & 3.67 & 4.12 & 3.55 & 3.68 \\

\midrule
\multicolumn{16}{l}{\textbf{Production Systems}} \\
Google Translate            & 4.71 & 4.31 & 4.46 & 5.05 & 4.55 & 4.95 & 4.38 & 4.37 & 4.86 & 4.63 & 4.28 & 4.93 & 4.28 & 4.12 & 4.41 \\
Youdao Translate            & 3.15 & 4.26 & 3.28 & 3.67 & 4.64 & 3.87 & 3.39 & 4.36 & 3.72 & 3.21 & 4.45 & 3.39 & 3.33 & 4.06 & 3.46 \\
\bottomrule
\end{tabular}
}
\caption{Fine-grained evaluation of translation quality across multiple dimensions for Zh$\rightarrow$Es, Zh$\rightarrow$Ja, and Zh$\rightarrow$Ru directions.}
\label{tab:fine_grained_results_4}
\end{table*}

\subsection{Language-level Results under Constraints}
\label{app:lang-level-res}
This appendix presents language-level evaluation results of different models under both semantic and communicative translation constraints across multiple language directions. While the main paper reports only averaged scores over language pairs for conciseness, we provide detailed per-language results here to offer a finer-grained view of model performance under different translation strategies. Results under semantic translation constraints are summarized in Tables~\ref{tab:semantic_results_1}--\ref{tab:semantic_results_4}, while results under communicative translation constraints are reported in Tables~\ref{tab:communicative_results_1}--\ref{tab:communicative_results_4}.

\begin{table*}[t]
\centering
\small
\resizebox{\textwidth}{!}{
\begin{tabular}{lcccccccccccccccc}
\toprule
\multirow{2}{*}{\textbf{Model}} & \multicolumn{3}{c}{\textbf{Contextual Accuracy}} & 
\multicolumn{3}{c}{\textbf{Cultural Adaptation}} & 
\multicolumn{3}{c}{\textbf{Functional Equivalence}} & 
\multicolumn{3}{c}{\textbf{Fidelity}} & 
\multicolumn{3}{c}{\textbf{Naturalness}} \\
\cmidrule(lr){2-4} \cmidrule(lr){5-7} \cmidrule(lr){8-10} \cmidrule(lr){11-13} \cmidrule(lr){14-16}
 & \textbf{En$\rightarrow$Es} & \textbf{En$\rightarrow$Ja} & \textbf{En$\rightarrow$Zh} & 
   \textbf{En$\rightarrow$Es} & \textbf{En$\rightarrow$Ja} & \textbf{En$\rightarrow$Zh} & 
   \textbf{En$\rightarrow$Es} & \textbf{En$\rightarrow$Ja} & \textbf{En$\rightarrow$Zh} & 
   \textbf{En$\rightarrow$Es} & \textbf{En$\rightarrow$Ja} & \textbf{En$\rightarrow$Zh} & 
   \textbf{En$\rightarrow$Es} & \textbf{En$\rightarrow$Ja} & \textbf{En$\rightarrow$Zh} \\
\midrule
Llama-3-8B-Instruct-262k   & 4.73 & 2.76 & 4.07 & 4.97 & 3.66 & 4.53 & 3.81 & 2.24 & 3.80 & 4.13 & 2.54 & 3.77 & 3.86 & 2.11 & 3.63 \\
Llama-3.3-70B-Instruct     & 5.08 & 5.26 & 5.20 & 5.28 & 5.29 & 5.20 & 4.79 & 4.51 & 4.96 & 4.90 & 4.83 & 5.22 & 4.42 & 4.17 & 4.52 \\
Mixtral-8x7B-Instruct-v0.1 & 4.67 & 3.14 & 3.48 & 4.91 & 3.67 & 3.90 & 3.98 & 2.54 & 2.55 & 4.10 & 2.76 & 2.72 & 3.85 & 2.46 & 2.63 \\
Qwen2.5-14B-Instruct       & 4.68 & 4.38 & 5.14 & 4.87 & 4.63 & 5.46 & 4.29 & 3.93 & 4.89 & 4.49 & 4.08 & 4.97 & 4.02 & 3.56 & 4.55 \\
Qwen2.5-32B-Instruct       & 4.84 & 4.59 & 5.45 & 4.81 & 4.99 & 5.37 & 4.61 & 4.03 & 5.10 & 4.77 & 4.12 & 5.08 & 4.14 & 3.72 & 4.65 \\
Qwen2.5-72B-Instruct       & 4.93 & 5.11 & 5.52 & 5.19 & 5.12 & 5.67 & 4.63 & 4.76 & 5.07 & 4.97 & 4.94 & 5.22 & 4.46 & 4.29 & 4.65 \\
Qwen2.5-7B-Instruct        & 4.63 & 4.06 & 4.77 & 4.88 & 4.21 & 5.06 & 3.85 & 3.38 & 4.41 & 4.16 & 3.44 & 4.55 & 3.88 & 3.09 & 4.23 \\
Qwen3-8B (w/o think)       & 4.45 & 4.57 & 5.03 & 5.06 & 4.62 & 5.32 & 3.96 & 4.14 & 4.76 & 4.25 & 4.26 & 4.71 & 3.96 & 3.67 & 4.40 \\
Qwen3-8B (w/ think)      & 4.48 & 4.34 & 5.17 & 4.92 & 4.67 & 5.37 & 4.25 & 3.91 & 4.85 & 4.45 & 4.17 & 5.01 & 4.08 & 3.58 & 4.54 \\
DeepSeek-V3.2              & 4.92 & 5.55 & 5.52 & 5.33 & 5.74 & 5.55 & 4.69 & 5.22 & 5.31 & 5.19 & 5.47 & 5.28 & 4.62 & 4.63 & 4.81 \\
Gemini-2.5-Flash-Lite      & 4.94 & 5.30 & 5.40 & 5.16 & 5.43 & 5.40 & 4.53 & 4.92 & 5.00 & 4.79 & 4.98 & 5.07 & 4.37 & 4.50 & 4.54 \\
GPT-4o                     & 4.92 & 5.30 & 5.39 & 5.31 & 5.37 & 5.40 & 4.67 & 4.89 & 5.05 & 5.10 & 5.34 & 5.10 & 4.66 & 4.39 & 4.65 \\
\bottomrule
\end{tabular}
}
\caption{Fine-grained evaluation of semantic translation quality across multiple dimensions for En$\rightarrow$Es, En$\rightarrow$Ja, and En$\rightarrow$Zh directions.}
\label{tab:semantic_results_1}
\end{table*}

\begin{table*}[t]
\centering
\small
\resizebox{\textwidth}{!}{
\begin{tabular}{lcccccccccccccccc}
\toprule
\multirow{2}{*}{\textbf{Model}} & \multicolumn{3}{c}{\textbf{Contextual Accuracy}} & 
\multicolumn{3}{c}{\textbf{Cultural Adaptation}} & 
\multicolumn{3}{c}{\textbf{Functional Equivalence}} & 
\multicolumn{3}{c}{\textbf{Fidelity}} & 
\multicolumn{3}{c}{\textbf{Naturalness}} \\
\cmidrule(lr){2-4} \cmidrule(lr){5-7} \cmidrule(lr){8-10} \cmidrule(lr){11-13} \cmidrule(lr){14-16}
 & \textbf{Es$\rightarrow$En} & \textbf{Es$\rightarrow$Zh} & \textbf{Ja$\rightarrow$En} & 
   \textbf{Es$\rightarrow$En} & \textbf{Es$\rightarrow$Zh} & \textbf{Ja$\rightarrow$En} & 
   \textbf{Es$\rightarrow$En} & \textbf{Es$\rightarrow$Zh} & \textbf{Ja$\rightarrow$En} & 
   \textbf{Es$\rightarrow$En} & \textbf{Es$\rightarrow$Zh} & \textbf{Ja$\rightarrow$En} & 
   \textbf{Es$\rightarrow$En} & \textbf{Es$\rightarrow$Zh} & \textbf{Ja$\rightarrow$En} \\
\midrule
Llama-3-8B-Instruct-262k   & 4.33 & 3.19 & 3.37 & 4.86 & 3.69 & 4.06 & 4.06 & 3.24 & 3.27 & 3.97 & 3.17 & 3.25 & 4.03 & 3.13 & 3.52 \\
Llama-3.3-70B-Instruct     & 5.21 & 4.42 & 4.59 & 5.15 & 4.63 & 4.75 & 4.76 & 4.81 & 3.89 & 4.96 & 4.64 & 4.01 & 4.34 & 4.44 & 4.07 \\
Mixtral-8x7B-Instruct-v0.1 & 4.99 & 3.13 & 3.46 & 5.13 & 3.67 & 4.12 & 4.73 & 2.52 & 3.37 & 4.74 & 2.51 & 3.33 & 4.29 & 2.48 & 3.39 \\
Qwen2.5-14B-Instruct       & 4.96 & 4.30 & 4.36 & 5.27 & 4.75 & 4.68 & 4.72 & 4.84 & 4.30 & 4.96 & 4.73 & 4.35 & 4.40 & 4.29 & 4.22 \\
Qwen2.5-32B-Instruct       & 5.09 & 4.55 & 4.59 & 5.27 & 4.72 & 4.87 & 4.88 & 4.87 & 4.06 & 4.92 & 4.75 & 4.35 & 4.36 & 4.42 & 4.29 \\
Qwen2.5-72B-Instruct       & 5.10 & 4.70 & 4.91 & 5.25 & 4.98 & 4.97 & 4.93 & 4.90 & 4.48 & 5.16 & 4.99 & 4.66 & 4.57 & 4.58 & 4.52 \\
Qwen2.5-7B-Instruct        & 4.69 & 4.04 & 3.63 & 5.00 & 4.43 & 4.11 & 4.56 & 4.23 & 3.80 & 4.41 & 4.07 & 3.68 & 4.07 & 3.81 & 3.78 \\
Qwen3-8B (w/o think)       & 4.79 & 4.44 & 3.82 & 5.11 & 4.60 & 4.10 & 4.37 & 4.75 & 3.86 & 4.48 & 4.48 & 3.76 & 3.94 & 4.34 & 3.91 \\
Qwen3-8B (w/ think)      & 4.47 & 4.30 & 4.15 & 4.82 & 4.52 & 4.43 & 4.17 & 4.58 & 3.84 & 4.51 & 4.57 & 3.99 & 3.91 & 4.10 & 3.78 \\
DeepSeek-V3.2              & 5.46 & 4.98 & 5.24 & 5.44 & 5.19 & 5.28 & 4.95 & 5.17 & 4.55 & 5.02 & 4.95 & 4.84 & 4.24 & 4.52 & 4.51 \\
Gemini-2.5-Flash-Lite      & 5.31 & 4.58 & 4.91 & 5.41 & 4.93 & 5.16 & 4.81 & 4.78 & 4.33 & 5.14 & 4.80 & 4.72 & 4.26 & 4.31 & 4.25 \\
GPT-4o                     & 5.49 & 4.80 & 5.09 & 5.46 & 4.99 & 5.28 & 5.04 & 5.01 & 4.66 & 5.26 & 4.80 & 5.03 & 4.59 & 4.37 & 4.49 \\
\bottomrule
\end{tabular}
}
\caption{Fine-grained evaluation of semantic translation quality across multiple dimensions for Es$\rightarrow$En, Es$\rightarrow$Zh, and Ja$\rightarrow$En directions.}
\label{tab:semantic_results_2}
\end{table*}

\begin{table*}[t]
\centering
\small
\resizebox{\textwidth}{!}{
\begin{tabular}{lcccccccccccccccc}
\toprule
\multirow{2}{*}{\textbf{Model}} & \multicolumn{3}{c}{\textbf{Contextual Accuracy}} & 
\multicolumn{3}{c}{\textbf{Cultural Adaptation}} & 
\multicolumn{3}{c}{\textbf{Functional Equivalence}} & 
\multicolumn{3}{c}{\textbf{Fidelity}} & 
\multicolumn{3}{c}{\textbf{Naturalness}} \\
\cmidrule(lr){2-4} \cmidrule(lr){5-7} \cmidrule(lr){8-10} \cmidrule(lr){11-13} \cmidrule(lr){14-16}
 & \textbf{Ja$\rightarrow$Zh} & \textbf{Ru$\rightarrow$Zh} & \textbf{Zh$\rightarrow$En} & 
   \textbf{Ja$\rightarrow$Zh} & \textbf{Ru$\rightarrow$Zh} & \textbf{Zh$\rightarrow$En} & 
   \textbf{Ja$\rightarrow$Zh} & \textbf{Ru$\rightarrow$Zh} & \textbf{Zh$\rightarrow$En} & 
   \textbf{Ja$\rightarrow$Zh} & \textbf{Ru$\rightarrow$Zh} & \textbf{Zh$\rightarrow$En} & 
   \textbf{Ja$\rightarrow$Zh} & \textbf{Ru$\rightarrow$Zh} & \textbf{Zh$\rightarrow$En} \\
\midrule
Llama-3-8B-Instruct-262k   & 3.35 & 3.58 & 4.34 & 3.92 & 3.98 & 4.59 & 2.98 & 3.61 & 4.38 & 3.08 & 3.41 & 4.27 & 3.12 & 3.40 & 4.37 \\
Llama-3.3-70B-Instruct     & 4.28 & 4.56 & 5.26 & 4.75 & 4.76 & 5.27 & 4.32 & 4.73 & 5.09 & 4.47 & 4.75 & 5.26 & 4.11 & 4.45 & 4.58 \\
Mixtral-8x7B-Instruct-v0.1 & 3.19 & 2.87 & 4.58 & 3.77 & 3.60 & 4.60 & 2.89 & 2.62 & 4.65 & 2.92 & 2.57 & 4.74 & 2.89 & 2.60 & 4.18 \\
Qwen2.5-14B-Instruct       & 4.49 & 4.50 & 5.44 & 4.84 & 4.91 & 5.46 & 4.61 & 4.62 & 5.31 & 4.75 & 4.50 & 5.39 & 4.32 & 4.30 & 4.72 \\
Qwen2.5-32B-Instruct       & 4.51 & 4.66 & 5.38 & 4.86 & 5.02 & 5.34 & 4.63 & 4.83 & 4.98 & 4.77 & 4.74 & 5.30 & 4.35 & 4.46 & 4.58 \\
Qwen2.5-72B-Instruct       & 4.97 & 4.71 & 5.61 & 5.03 & 4.95 & 5.64 & 4.71 & 4.94 & 5.32 & 5.00 & 4.92 & 5.62 & 4.42 & 4.71 & 4.95 \\
Qwen2.5-7B-Instruct        & 4.38 & 4.39 & 5.00 & 4.78 & 4.48 & 5.25 & 4.34 & 4.29 & 4.96 & 4.42 & 4.30 & 4.94 & 4.29 & 4.16 & 4.57 \\
Qwen3-8B (w/o think)       & 4.16 & 4.56 & 5.12 & 4.66 & 4.93 & 5.30 & 4.03 & 4.76 & 5.02 & 4.18 & 4.43 & 5.11 & 3.98 & 4.53 & 4.70 \\
Qwen3-8B (w/ think)      & 4.46 & 4.49 & 5.09 & 4.82 & 4.77 & 5.12 & 4.31 & 4.62 & 5.10 & 4.61 & 4.71 & 5.35 & 4.09 & 4.14 & 4.56 \\
DeepSeek-V3.2              & 4.95 & 4.91 & 5.54 & 5.20 & 5.03 & 5.42 & 4.92 & 5.02 & 5.32 & 5.00 & 4.72 & 5.54 & 4.47 & 4.58 & 4.76 \\
Gemini-2.5-Flash-Lite      & 4.58 & 4.92 & 5.58 & 4.96 & 5.10 & 5.37 & 4.51 & 4.90 & 4.97 & 4.80 & 4.98 & 5.46 & 4.25 & 4.46 & 4.58 \\
GPT-4o                     & 5.06 & 4.77 & 5.66 & 5.27 & 5.10 & 5.31 & 4.97 & 4.62 & 5.18 & 5.23 & 4.83 & 5.54 & 4.49 & 4.22 & 4.71 \\
\bottomrule
\end{tabular}
}
\caption{Fine-grained evaluation of semantic translation quality across multiple dimensions for Ja$\rightarrow$Zh, Ru$\rightarrow$Zh, and Zh$\rightarrow$En directions.}
\label{tab:semantic_results_3}
\end{table*}

\begin{table*}[t]
\centering
\small
\resizebox{\textwidth}{!}{
\begin{tabular}{lcccccccccccccccc}
\toprule
\multirow{2}{*}{\textbf{Model}} & \multicolumn{3}{c}{\textbf{Contextual Accuracy}} & 
\multicolumn{3}{c}{\textbf{Cultural Adaptation}} & 
\multicolumn{3}{c}{\textbf{Functional Equivalence}} & 
\multicolumn{3}{c}{\textbf{Fidelity}} & 
\multicolumn{3}{c}{\textbf{Naturalness}} \\
\cmidrule(lr){2-4} \cmidrule(lr){5-7} \cmidrule(lr){8-10} \cmidrule(lr){11-13} \cmidrule(lr){14-16}
 & \textbf{Zh$\rightarrow$Es} & \textbf{Zh$\rightarrow$Ja} & \textbf{Zh$\rightarrow$Ru} & 
   \textbf{Zh$\rightarrow$Es} & \textbf{Zh$\rightarrow$Ja} & \textbf{Zh$\rightarrow$Ru} & 
   \textbf{Zh$\rightarrow$Es} & \textbf{Zh$\rightarrow$Ja} & \textbf{Zh$\rightarrow$Ru} & 
   \textbf{Zh$\rightarrow$Es} & \textbf{Zh$\rightarrow$Ja} & \textbf{Zh$\rightarrow$Ru} & 
   \textbf{Zh$\rightarrow$Es} & \textbf{Zh$\rightarrow$Ja} & \textbf{Zh$\rightarrow$Ru} \\
\midrule
Llama-3-8B-Instruct-262k   & 3.35 & 3.29 & 2.99 & 3.92 & 3.85 & 3.74 & 3.91 & 3.34 & 3.67 & 3.56 & 3.20 & 3.34 & 3.72 & 2.98 & 3.36 \\
Llama-3.3-70B-Instruct     & 4.91 & 4.58 & 4.72 & 4.91 & 4.57 & 4.85 & 4.83 & 4.72 & 4.82 & 5.05 & 4.80 & 4.86 & 4.49 & 4.19 & 4.38 \\
Mixtral-8x7B-Instruct-v0.1 & 3.95 & 3.13 & 3.46 & 4.26 & 3.43 & 3.89 & 3.92 & 2.71 & 3.52 & 3.97 & 2.72 & 3.49 & 3.86 & 2.44 & 3.25 \\
Qwen2.5-14B-Instruct       & 4.34 & 4.41 & 3.67 & 4.59 & 4.45 & 4.29 & 4.55 & 4.42 & 3.84 & 4.48 & 4.34 & 3.93 & 4.26 & 3.75 & 3.36 \\
Qwen2.5-32B-Instruct       & 4.69 & 4.48 & 4.03 & 4.83 & 4.66 & 4.39 & 4.66 & 4.47 & 4.12 & 4.83 & 4.47 & 4.09 & 4.27 & 3.90 & 3.76 \\
Qwen2.5-72B-Instruct       & 5.03 & 4.73 & 4.60 & 4.92 & 4.79 & 4.79 & 5.02 & 4.85 & 4.83 & 5.19 & 4.92 & 4.93 & 4.79 & 4.37 & 4.52 \\
Qwen2.5-7B-Instruct        & 3.83 & 3.78 & 3.31 & 4.21 & 4.12 & 4.00 & 3.94 & 3.96 & 3.60 & 3.93 & 3.92 & 3.29 & 3.75 & 3.57 & 3.08 \\
Qwen3-8B (w/o think)       & 4.23 & 4.17 & 3.71 & 4.51 & 4.52 & 3.96 & 4.55 & 4.36 & 4.18 & 4.30 & 4.37 & 3.88 & 4.07 & 3.83 & 3.88 \\
Qwen3-8B (w/ think)      & 4.68 & 4.32 & 4.20 & 4.92 & 4.53 & 4.42 & 4.68 & 4.82 & 4.68 & 4.83 & 5.08 & 4.73 & 4.29 & 4.15 & 4.09 \\
DeepSeek-V3.2              & 5.21 & 5.26 & 4.89 & 5.27 & 5.19 & 5.21 & 4.93 & 5.33 & 5.01 & 5.13 & 5.58 & 4.93 & 4.76 & 4.73 & 4.35 \\
Gemini-2.5-Flash-Lite      & 5.03 & 5.41 & 5.21 & 5.00 & 5.18 & 5.27 & 4.86 & 5.18 & 5.17 & 5.23 & 5.52 & 5.29 & 4.42 & 4.68 & 4.57 \\
GPT-4o                     & 5.18 & 5.18 & 5.14 & 5.16 & 4.96 & 5.25 & 4.97 & 5.10 & 5.22 & 5.18 & 5.34 & 5.43 & 4.61 & 4.57 & 4.51 \\
\bottomrule
\end{tabular}
}
\caption{Fine-grained evaluation of semantic translation quality across multiple dimensions for Zh$\rightarrow$Es, Zh$\rightarrow$Ja, and Zh$\rightarrow$Ru directions.}
\label{tab:semantic_results_4}
\end{table*}

\begin{table*}[t]
\centering
\small
\resizebox{\textwidth}{!}{
\begin{tabular}{lcccccccccccccccc}
\toprule
\multirow{2}{*}{\textbf{Model}} & \multicolumn{3}{c}{\textbf{Contextual Accuracy}} & 
\multicolumn{3}{c}{\textbf{Cultural Adaptation}} & 
\multicolumn{3}{c}{\textbf{Functional Equivalence}} & 
\multicolumn{3}{c}{\textbf{Fidelity}} & 
\multicolumn{3}{c}{\textbf{Naturalness}} \\
\cmidrule(lr){2-4} \cmidrule(lr){5-7} \cmidrule(lr){8-10} \cmidrule(lr){11-13} \cmidrule(lr){14-16}
 & \textbf{En$\rightarrow$Es} & \textbf{En$\rightarrow$Ja} & \textbf{En$\rightarrow$Zh} & 
   \textbf{En$\rightarrow$Es} & \textbf{En$\rightarrow$Ja} & \textbf{En$\rightarrow$Zh} & 
   \textbf{En$\rightarrow$Es} & \textbf{En$\rightarrow$Ja} & \textbf{En$\rightarrow$Zh} & 
   \textbf{En$\rightarrow$Es} & \textbf{En$\rightarrow$Ja} & \textbf{En$\rightarrow$Zh} & 
   \textbf{En$\rightarrow$Es} & \textbf{En$\rightarrow$Ja} & \textbf{En$\rightarrow$Zh} \\
\midrule
Llama-3-8B-Instruct-262k   & 4.59 & 3.17 & 4.23 & 5.03 & 3.81 & 4.50 & 3.84 & 2.76 & 3.83 & 4.16 & 2.78 & 3.78 & 3.97 & 2.74 & 3.64 \\
Llama-3.3-70B-Instruct     & 4.92 & 5.17 & 4.99 & 5.16 & 5.39 & 5.20 & 4.82 & 4.76 & 4.92 & 4.80 & 4.79 & 4.93 & 4.63 & 4.30 & 4.83 \\
Mixtral-8x7B-Instruct-v0.1 & 4.69 & 3.24 & 3.69 & 5.14 & 3.74 & 3.94 & 4.41 & 2.44 & 2.58 & 4.25 & 2.58 & 2.65 & 4.17 & 2.53 & 2.57 \\
Qwen2.5-14B-Instruct       & 4.45 & 4.72 & 5.30 & 4.76 & 4.78 & 5.45 & 4.45 & 4.32 & 5.05 & 4.29 & 4.25 & 4.90 & 4.21 & 4.00 & 4.91 \\
Qwen2.5-32B-Instruct       & 4.50 & 4.65 & 5.35 & 4.95 & 4.95 & 5.32 & 4.51 & 4.37 & 5.06 & 4.42 & 4.30 & 5.02 & 4.28 & 4.13 & 4.75 \\
Qwen2.5-72B-Instruct       & 4.92 & 5.06 & 5.59 & 5.18 & 5.20 & 5.64 & 4.75 & 4.69 & 4.98 & 4.86 & 4.72 & 5.19 & 4.71 & 4.46 & 5.03 \\
Qwen2.5-7B-Instruct        & 4.29 & 4.00 & 4.78 & 4.81 & 4.31 & 5.14 & 3.92 & 3.29 & 4.49 & 4.09 & 3.35 & 4.52 & 3.82 & 3.14 & 4.30 \\
Qwen3-8B (w/o think)       & 4.73 & 4.53 & 5.16 & 5.02 & 4.73 & 5.32 & 4.26 & 4.27 & 4.85 & 4.43 & 4.44 & 4.83 & 4.16 & 3.95 & 4.49 \\
Qwen3-8B (w/ think)      & 4.75 & 4.79 & 5.33 & 5.08 & 4.92 & 5.53 & 4.50 & 4.55 & 4.93 & 4.59 & 4.23 & 4.97 & 4.28 & 3.85 & 4.77 \\
DeepSeek-V3.2              & 4.99 & 5.04 & 5.32 & 5.34 & 5.31 & 5.54 & 4.92 & 4.95 & 5.10 & 4.71 & 4.95 & 4.90 & 4.69 & 4.60 & 4.95 \\
Gemini-2.5-Flash-Lite      & 4.77 & 5.43 & 5.28 & 5.24 & 5.54 & 5.27 & 4.81 & 5.07 & 5.02 & 4.83 & 5.00 & 4.93 & 4.58 & 4.84 & 4.69 \\
GPT-4o                     & 5.09 & 5.27 & 5.63 & 5.15 & 5.44 & 5.55 & 4.59 & 5.05 & 5.23 & 4.71 & 4.93 & 5.19 & 4.66 & 4.81 & 5.02 \\
\bottomrule
\end{tabular}
}
\caption{Fine-grained evaluation of communicative translation quality across multiple dimensions for En$\rightarrow$Es, En$\rightarrow$Ja, and En$\rightarrow$Zh directions.}
\label{tab:communicative_results_1}
\end{table*}

\begin{table*}[t]
\centering
\small
\resizebox{\textwidth}{!}{
\begin{tabular}{lcccccccccccccccc}
\toprule
\multirow{2}{*}{\textbf{Model}} & \multicolumn{3}{c}{\textbf{Contextual Accuracy}} & 
\multicolumn{3}{c}{\textbf{Cultural Adaptation}} & 
\multicolumn{3}{c}{\textbf{Functional Equivalence}} & 
\multicolumn{3}{c}{\textbf{Fidelity}} & 
\multicolumn{3}{c}{\textbf{Naturalness}} \\
\cmidrule(lr){2-4} \cmidrule(lr){5-7} \cmidrule(lr){8-10} \cmidrule(lr){11-13} \cmidrule(lr){14-16}
 & \textbf{Es$\rightarrow$En} & \textbf{Es$\rightarrow$Zh} & \textbf{Ja$\rightarrow$En} & 
   \textbf{Es$\rightarrow$En} & \textbf{Es$\rightarrow$Zh} & \textbf{Ja$\rightarrow$En} & 
   \textbf{Es$\rightarrow$En} & \textbf{Es$\rightarrow$Zh} & \textbf{Ja$\rightarrow$En} & 
   \textbf{Es$\rightarrow$En} & \textbf{Es$\rightarrow$Zh} & \textbf{Ja$\rightarrow$En} & 
   \textbf{Es$\rightarrow$En} & \textbf{Es$\rightarrow$Zh} & \textbf{Ja$\rightarrow$En} \\
\midrule
Llama-3-8B-Instruct-262k   & 4.38 & 3.22 & 3.19 & 4.78 & 3.73 & 3.72 & 4.16 & 3.28 & 3.39 & 3.91 & 3.08 & 3.33 & 4.10 & 3.22 & 3.68 \\
Llama-3.3-70B-Instruct     & 4.82 & 4.36 & 4.44 & 5.20 & 4.67 & 4.88 & 5.09 & 4.72 & 4.43 & 4.51 & 4.44 & 4.40 & 4.99 & 4.44 & 4.70 \\
Mixtral-8x7B-Instruct-v0.1 & 4.74 & 3.23 & 3.35 & 5.16 & 3.66 & 4.02 & 4.81 & 2.74 & 3.57 & 4.43 & 2.66 & 3.17 & 4.73 & 2.67 & 3.97 \\
Qwen2.5-14B-Instruct       & 4.96 & 4.30 & 4.40 & 5.19 & 4.78 & 4.68 & 5.04 & 4.73 & 4.36 & 4.70 & 4.45 & 4.34 & 4.66 & 4.55 & 4.54 \\
Qwen2.5-32B-Instruct       & 4.89 & 4.39 & 4.67 & 5.20 & 4.63 & 4.99 & 5.00 & 4.76 & 4.64 & 4.60 & 4.40 & 4.53 & 4.92 & 4.63 & 4.78 \\
Qwen2.5-72B-Instruct       & 5.04 & 4.57 & 4.80 & 5.31 & 5.00 & 5.00 & 5.07 & 5.07 & 4.72 & 4.81 & 4.62 & 4.80 & 5.00 & 4.69 & 4.86 \\
Qwen2.5-7B-Instruct        & 4.77 & 3.99 & 3.61 & 5.05 & 4.33 & 4.26 & 4.60 & 4.23 & 3.91 & 4.27 & 3.98 & 3.53 & 4.46 & 4.05 & 3.92 \\
Qwen3-8B (w/o think)       & 4.63 & 4.25 & 3.82 & 5.00 & 4.67 & 4.27 & 4.49 & 4.69 & 4.04 & 4.47 & 4.35 & 3.89 & 4.38 & 4.33 & 4.18 \\
Qwen3-8B (w/ think)      & 4.88 & 4.47 & 4.33 & 5.01 & 4.55 & 4.61 & 4.65 & 4.70 & 4.38 & 4.67 & 4.48 & 4.44 & 4.47 & 4.43 & 4.36 \\
DeepSeek-V3.2              & 5.24 & 4.69 & 4.95 & 5.29 & 5.06 & 5.21 & 5.08 & 5.10 & 4.75 & 4.62 & 4.48 & 4.68 & 5.05 & 4.67 & 4.82 \\
Gemini-2.5-Flash-Lite      & 4.96 & 4.40 & 5.08 & 5.25 & 4.79 & 5.28 & 5.04 & 4.93 & 4.76 & 4.96 & 4.63 & 4.89 & 5.05 & 4.75 & 4.78 \\
GPT-4o                     & 4.91 & 4.65 & 4.97 & 5.17 & 4.95 & 5.17 & 5.15 & 4.89 & 4.89 & 4.66 & 4.58 & 5.01 & 4.98 & 4.71 & 5.00 \\
\bottomrule
\end{tabular}
}
\caption{Fine-grained evaluation of communicative translation quality across multiple dimensions for Es$\rightarrow$En, Es$\rightarrow$Zh, and Ja$\rightarrow$En directions.}
\label{tab:communicative_results_2}
\end{table*}

\begin{table*}[t]
\centering
\small
\resizebox{\textwidth}{!}{
\begin{tabular}{lcccccccccccccccc}
\toprule
\multirow{2}{*}{\textbf{Model}} & \multicolumn{3}{c}{\textbf{Contextual Accuracy}} & 
\multicolumn{3}{c}{\textbf{Cultural Adaptation}} & 
\multicolumn{3}{c}{\textbf{Functional Equivalence}} & 
\multicolumn{3}{c}{\textbf{Fidelity}} & 
\multicolumn{3}{c}{\textbf{Naturalness}} \\
\cmidrule(lr){2-4} \cmidrule(lr){5-7} \cmidrule(lr){8-10} \cmidrule(lr){11-13} \cmidrule(lr){14-16}
 & \textbf{Ja$\rightarrow$Zh} & \textbf{Ru$\rightarrow$Zh} & \textbf{Zh$\rightarrow$En} & 
   \textbf{Ja$\rightarrow$Zh} & \textbf{Ru$\rightarrow$Zh} & \textbf{Zh$\rightarrow$En} & 
   \textbf{Ja$\rightarrow$Zh} & \textbf{Ru$\rightarrow$Zh} & \textbf{Zh$\rightarrow$En} & 
   \textbf{Ja$\rightarrow$Zh} & \textbf{Ru$\rightarrow$Zh} & \textbf{Zh$\rightarrow$En} & 
   \textbf{Ja$\rightarrow$Zh} & \textbf{Ru$\rightarrow$Zh} & \textbf{Zh$\rightarrow$En} \\
\midrule
Llama-3-8B-Instruct-262k   & 3.25 & 3.26 & 4.46 & 3.74 & 3.91 & 4.58 & 3.11 & 3.55 & 4.50 & 3.30 & 3.25 & 4.18 & 3.32 & 3.64 & 4.43 \\
Llama-3.3-70B-Instruct     & 4.40 & 4.22 & 5.10 & 4.73 & 4.64 & 5.25 & 4.50 & 4.66 & 5.30 & 4.40 & 4.34 & 4.93 & 4.52 & 4.58 & 5.08 \\
Mixtral-8x7B-Instruct-v0.1 & 3.11 & 3.33 & 4.78 & 3.54 & 3.83 & 4.98 & 2.93 & 2.72 & 4.82 & 2.78 & 2.74 & 4.74 & 3.00 & 2.78 & 4.68 \\
Qwen2.5-14B-Instruct       & 4.16 & 4.39 & 5.30 & 4.72 & 4.83 & 5.41 & 4.73 & 4.74 & 5.38 & 4.76 & 4.43 & 5.08 & 4.71 & 4.70 & 5.04 \\
Qwen2.5-32B-Instruct       & 4.62 & 4.51 & 5.06 & 5.00 & 4.78 & 5.38 & 4.91 & 4.76 & 5.19 & 4.79 & 4.47 & 4.98 & 4.74 & 4.79 & 4.99 \\
Qwen2.5-72B-Instruct       & 5.08 & 4.76 & 5.33 & 5.19 & 5.00 & 5.54 & 4.95 & 4.98 & 5.34 & 5.01 & 5.03 & 5.18 & 4.86 & 4.95 & 5.07 \\
Qwen2.5-7B-Instruct        & 4.36 & 4.19 & 5.15 & 4.70 & 4.69 & 5.16 & 4.40 & 4.40 & 4.92 & 4.24 & 4.09 & 4.97 & 4.29 & 4.25 & 4.74 \\
Qwen3-8B (w/o think)       & 4.40 & 4.36 & 4.98 & 4.89 & 4.77 & 5.08 & 4.36 & 4.74 & 5.08 & 4.50 & 4.47 & 5.01 & 4.33 & 4.61 & 4.79 \\
Qwen3-8B (w/ think)      & 4.52 & 4.44 & 5.30 & 4.90 & 4.77 & 5.36 & 4.67 & 4.66 & 5.24 & 4.85 & 4.60 & 5.21 & 4.59 & 4.41 & 4.90 \\
DeepSeek-V3.2              & 5.07 & 4.64 & 5.29 & 5.20 & 5.06 & 5.38 & 5.01 & 4.99 & 5.26 & 4.72 & 4.60 & 5.06 & 4.71 & 4.69 & 5.04 \\
Gemini-2.5-Flash-Lite      & 4.92 & 4.66 & 5.63 & 5.02 & 5.15 & 5.54 & 5.02 & 4.84 & 5.32 & 4.99 & 4.73 & 5.35 & 4.75 & 4.70 & 5.16 \\
GPT-4o                     & 4.89 & 4.49 & 5.48 & 5.05 & 4.97 & 5.47 & 4.96 & 4.77 & 5.41 & 4.97 & 4.52 & 5.20 & 5.02 & 4.79 & 5.32 \\
\bottomrule
\end{tabular}
}
\caption{Fine-grained evaluation of communicative translation quality across multiple dimensions for Ja$\rightarrow$Zh, Ru$\rightarrow$Zh, and Zh$\rightarrow$En directions.}
\label{tab:communicative_results_3}
\end{table*}

\begin{table*}[t]
\centering
\small
\resizebox{\textwidth}{!}{
\begin{tabular}{lcccccccccccccccc}
\toprule
\multirow{2}{*}{\textbf{Model}} & \multicolumn{3}{c}{\textbf{Contextual Accuracy}} & 
\multicolumn{3}{c}{\textbf{Cultural Adaptation}} & 
\multicolumn{3}{c}{\textbf{Functional Equivalence}} & 
\multicolumn{3}{c}{\textbf{Fidelity}} & 
\multicolumn{3}{c}{\textbf{Naturalness}} \\
\cmidrule(lr){2-4} \cmidrule(lr){5-7} \cmidrule(lr){8-10} \cmidrule(lr){11-13} \cmidrule(lr){14-16}
 & \textbf{Zh$\rightarrow$Es} & \textbf{Zh$\rightarrow$Ja} & \textbf{Zh$\rightarrow$Ru} & 
   \textbf{Zh$\rightarrow$Es} & \textbf{Zh$\rightarrow$Ja} & \textbf{Zh$\rightarrow$Ru} & 
   \textbf{Zh$\rightarrow$Es} & \textbf{Zh$\rightarrow$Ja} & \textbf{Zh$\rightarrow$Ru} & 
   \textbf{Zh$\rightarrow$Es} & \textbf{Zh$\rightarrow$Ja} & \textbf{Zh$\rightarrow$Ru} & 
   \textbf{Zh$\rightarrow$Es} & \textbf{Zh$\rightarrow$Ja} & \textbf{Zh$\rightarrow$Ru} \\
\midrule
Llama-3-8B-Instruct-262k   & 3.23 & 3.34 & 2.85 & 3.83 & 3.88 & 3.63 & 3.80 & 3.47 & 3.47 & 3.59 & 3.36 & 3.23 & 3.90 & 3.22 & 3.44 \\
Llama-3.3-70B-Instruct     & 4.73 & 4.72 & 4.34 & 4.99 & 4.88 & 4.85 & 4.86 & 4.80 & 4.74 & 4.81 & 4.81 & 4.66 & 4.88 & 4.37 & 4.71 \\
Mixtral-8x7B-Instruct-v0.1 & 4.29 & 3.23 & 3.84 & 4.63 & 3.71 & 4.20 & 4.40 & 2.80 & 3.46 & 4.02 & 2.82 & 3.23 & 4.26 & 2.61 & 3.26 \\
Qwen2.5-14B-Instruct       & 4.33 & 4.47 & 3.58 & 4.71 & 4.72 & 4.27 & 4.83 & 4.61 & 4.12 & 4.60 & 4.57 & 3.84 & 4.52 & 4.20 & 3.88 \\
Qwen2.5-32B-Instruct       & 4.47 & 4.28 & 4.12 & 4.79 & 4.66 & 4.46 & 4.75 & 4.62 & 4.39 & 4.53 & 4.47 & 4.11 & 4.56 & 4.33 & 4.02 \\
Qwen2.5-72B-Instruct       & 4.82 & 4.57 & 4.35 & 5.11 & 4.82 & 4.65 & 5.01 & 4.88 & 4.97 & 5.01 & 4.71 & 4.79 & 4.99 & 4.63 & 4.69 \\
Qwen2.5-7B-Instruct        & 3.91 & 3.80 & 3.20 & 4.32 & 4.20 & 3.75 & 4.13 & 3.93 & 3.42 & 3.97 & 3.88 & 3.39 & 4.01 & 3.60 & 3.17 \\
Qwen3-8B (w/o think)       & 4.16 & 4.26 & 3.65 & 4.57 & 4.49 & 4.01 & 4.38 & 4.55 & 4.45 & 4.29 & 4.37 & 4.15 & 4.24 & 4.01 & 4.02 \\
Qwen3-8B (w/ think)      & 4.68 & 4.56 & 3.98 & 4.82 & 4.64 & 4.35 & 4.77 & 4.82 & 4.60 & 4.83 & 5.05 & 4.38 & 4.49 & 4.31 & 4.05 \\
DeepSeek-V3.2              & 4.79 & 5.06 & 4.37 & 5.12 & 5.22 & 4.68 & 5.01 & 5.32 & 3.93 & 4.78 & 5.07 & 4.02 & 4.65 & 4.93 & 3.69 \\
Gemini-2.5-Flash-Lite      & 5.01 & 5.12 & 4.81 & 5.18 & 5.40 & 5.23 & 5.08 & 5.43 & 5.25 & 5.09 & 5.44 & 5.11 & 4.89 & 4.99 & 4.73 \\
GPT-4o                     & 4.84 & 4.90 & 5.06 & 4.96 & 5.26 & 5.16 & 5.18 & 5.47 & 5.25 & 5.01 & 5.38 & 5.11 & 4.99 & 5.03 & 4.92 \\
\bottomrule
\end{tabular}
}
\caption{Fine-grained evaluation of communicative translation quality across multiple dimensions for Zh$\rightarrow$Es, Zh$\rightarrow$Ja, and Zh$\rightarrow$Ru directions.}
\label{tab:communicative_results_4}
\end{table*}

\subsection{Failure Analysis of Strong Models}
\label{app:hard_cases}

To investigate why strong models still fail on certain samples, we analyze the lowest-scoring cases and examine whether these failures are primarily caused by sampling variance.

Specifically, for each challenging sample, we generate two additional translations under a temperature setting of $T=0.7$, resulting in three outputs per sample. We conduct this experiment on five strong models (Grok-4.1, DeepSeek-V3.2, DeepSeek-R1, GPT-4o, and GPT-4). For each evaluation dimension, we compute the proportion of samples for which all three outputs receive scores $\leq 3$, relative to the total number of low-scoring cases.

\begin{table}[t]
\centering
\small
\resizebox{\columnwidth}{!}{
\begin{tabular}{lccccc}
\toprule
Model & Ctx. Acc. & Cul. Adapt. & Fidelity & Func. Eq. & Naturalness \\
\midrule
DeepSeek-R1 & 42.93\% & 24.05\% & 36.96\% & 26.29\% & 32.93\% \\
DeepSeek-V3.2 & 42.60\% & 27.16\% & 28.26\% & 25.85\% & 23.45\% \\
GPT-4 & 51.03\% & 35.07\% & 40.57\% & 30.26\% & 36.13\% \\
GPT-4o & 48.91\% & 35.15\% & 38.57\% & 28.89\% & 29.44\% \\
Grok-4.1 & 42.40\% & 19.20\% & 35.39\% & 35.81\% & 35.71\% \\
\bottomrule
\end{tabular}
}
\caption{Proportion of challenging samples whose multiple generations consistently receive low scores.}
\label{tab:hard_case_analysis}
\end{table}

As shown in Table~\ref{tab:hard_case_analysis}, a substantial proportion of low-scoring cases remain consistently poor across multiple generations. This suggests that these failures are not merely due to unfavorable sampling, but reflect more systematic weaknesses in model behavior.

To further understand these failures, we conduct a manual analysis on a subset of samples and identify several recurring error types, including: (1) errors in translating CSIs (e.g., loss of cultural connotation or mistranslation of culturally bound concepts), (2) translation hallucinations, (3) errors in translating units of measurement, (4) mishandling of polysemy, and (5) translationese.

Overall, these challenging cases reveal persistent limitations of strong models in handling culturally and semantically complex phenomena.

\section{Default Translation Preference under Varying Temperatures}
\label{app:preference}

To further validate the stability of the observed default translation preference, we conduct additional experiments under varying decoding temperatures.

\begin{figure*}[t]
  \centering
  \includegraphics[width=0.95\textwidth]{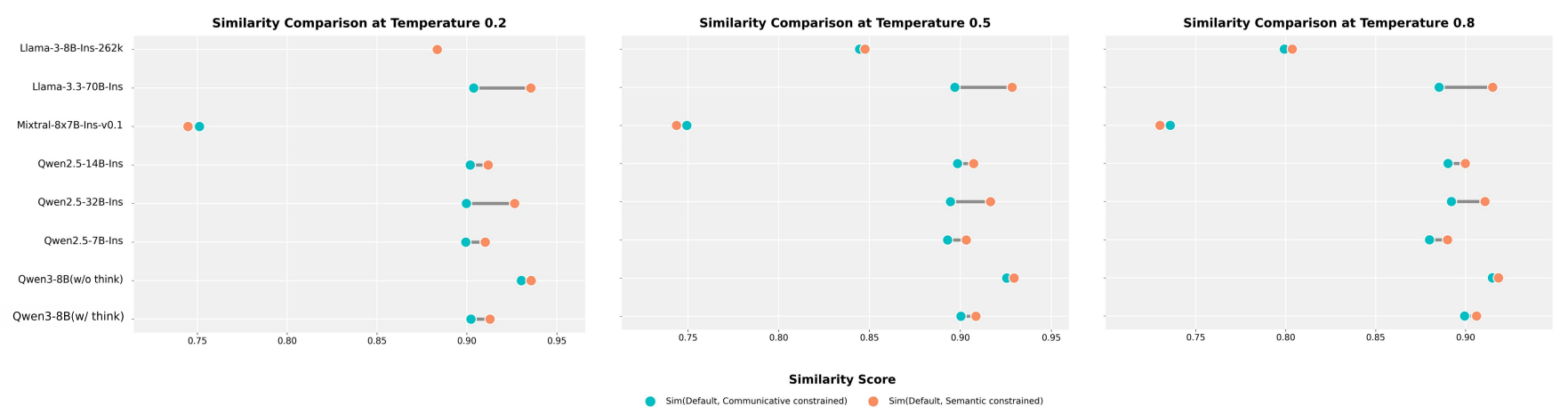}
  \caption{Similarity of default translations to semantic and communicative constrained translations under varying temperatures.}
  \label{fig:sim_vary_tmp}
\end{figure*}

Specifically, for each model, we generate default translations using multiple temperature settings (e.g., $T \in \{0.2, 0.5, 0.8\}$), and compute the cosine similarity. The results are shown in Figure~\ref{fig:sim_vary_tmp}.

Across all temperature settings, default translations consistently exhibit higher similarity to semantic-constrained translations for most models. This observation is consistent with the findings in Section~\ref{sec:Default Translation Preference Across Different Strategies}.

\section{CSI Categorization Details}
\label{app:csi}

\subsection{CSI Taxonomy}
\label{app:csi-taxonomy}
\begin{figure*}[t]
  \includegraphics[width=\textwidth]{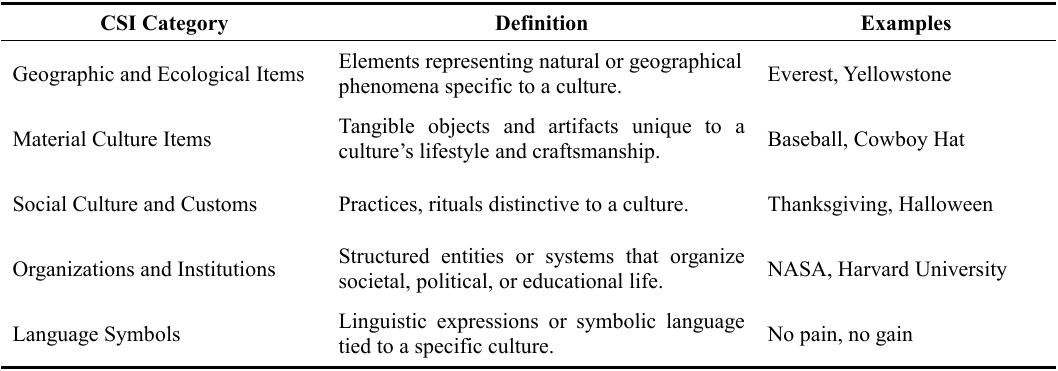}
  \caption{Definitions and representative examples of CSI categories.}
  \label{tab:CSI_Category}
\end{figure*}

We categorize culture-specific items based on Newmark’s taxonomy~\citep{newmark1988textbook}, which classifies CSIs into five main types: geographic and ecological items, material culture items, social culture and customs, organizations and institutions, and language symbols. Building on this framework, we adapt and refine the definitions to better suit the context of our translation evaluation. Figure~\ref{tab:CSI_Category} provides the resulting definitions and representative examples for each category.

\subsection{Automatic CSI Classification}
\label{app:csi-classify-prompt}
\begin{figure}[t]
  \centering
  \includegraphics[width=\columnwidth]{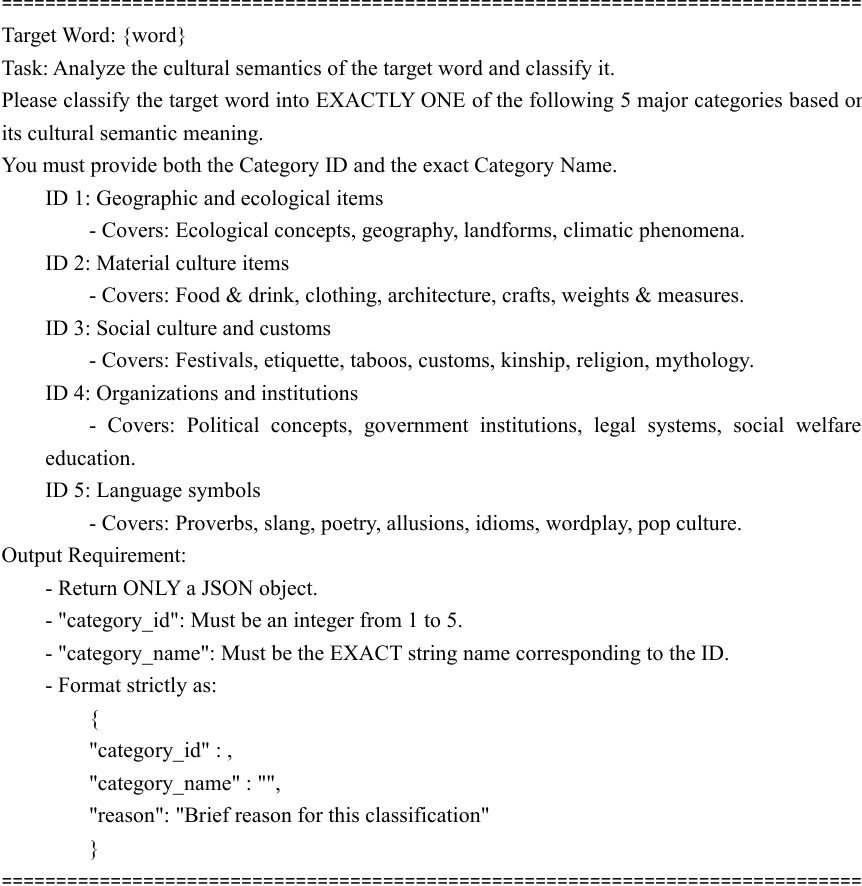}
  \caption{Prompt used for automatic classification of Culture-Specific Items.}   
  \label{fig:csi-classify-prompt}
\end{figure}

In this work, culture-specific items are automatically identified and classified using GPT-4o. Each CSI instance is assigned to one of the five CSI categories. To illustrate this process, Figure~\ref{fig:csi-classify-prompt} shows the prompt used for the automatic classification of CSIs, specifying the five categories and the expected JSON output format.

\subsection{Category-wise CSI Examples}
\label{sec:category_case_csi}

\begin{figure}[t]
  \centering
  \includegraphics[width=\columnwidth]{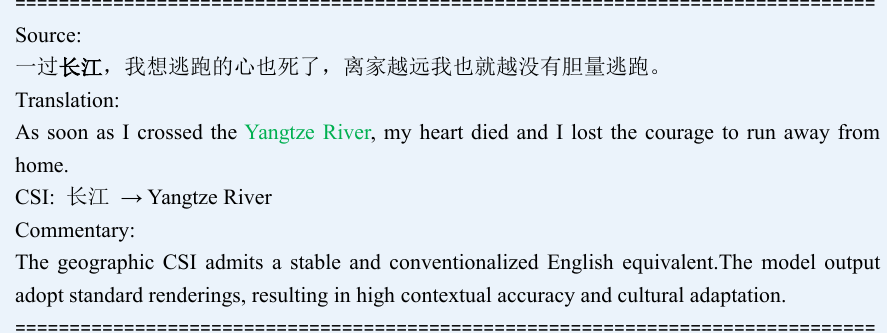}
  \caption{Translation of Geography and Ecology CSI.}
  \label{fig:example_geo_01}
\end{figure}

\begin{figure}[t]
  \centering
  \includegraphics[width=\columnwidth]{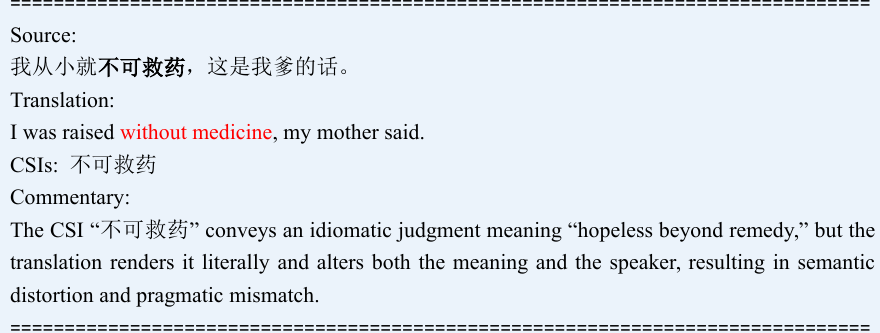}
  \caption{Translation of Language symbols CSI.}
  \label{fig:example_lang_02}
\end{figure}

As discussed in Section~\ref{sec:csi-performance}, Geographic and Ecological items tend to achieve the highest translation scores, whereas Language Symbols consistently exhibit the lowest performance. To illustrate these patterns qualitatively, we present representative examples of CSI translations across different categories. Examples of Geography and Ecology CSIs are shown in Figures~\ref{fig:example_geo_01}, while examples of Language Symbols CSIs are illustrated in Figures~\ref{fig:example_lang_02}.

\section{Cultural Knowledge Probing}
\label{app:Cultural Knowledge Probing}

\subsection{Generation of Questions}
\label{app:generation-of-questions}
\begin{figure}[t]
  \centering
  \includegraphics[width=\columnwidth]{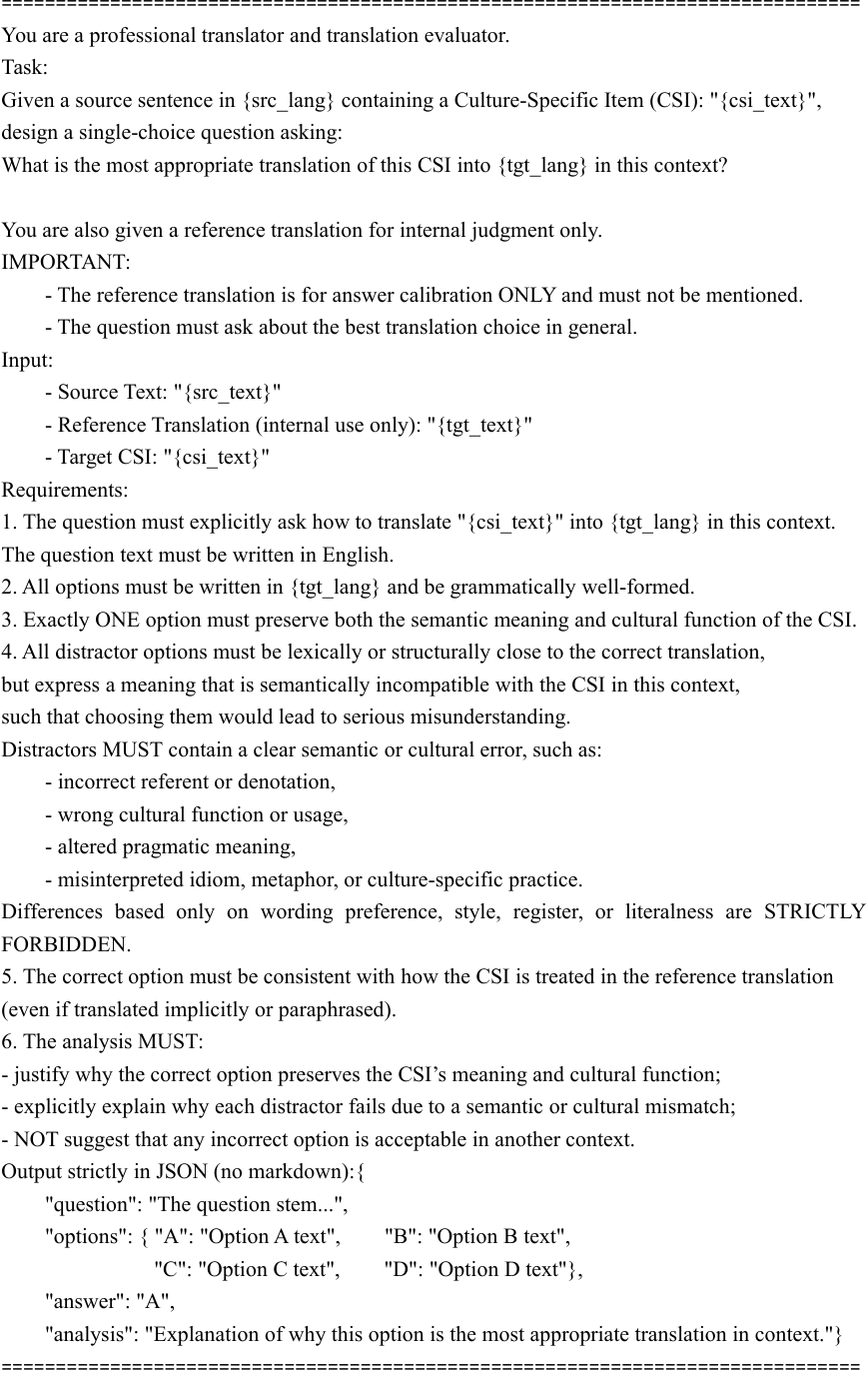}
  \caption{Prompt for Generating Culture-Specific Translation SCQs.}
  \label{fig:scq_generate_prompt}
\end{figure}

To probe models’ cultural translation knowledge, we employ GPT-4o to automatically generate single-choice questions for each CSI. For each item, the model is asked to select the most appropriate translation of the CSI given its context from four candidate options. During question construction, a reference translation is used internally to ensure the correctness of the target option. The detailed prompt template is presented in Figure~\ref{fig:scq_generate_prompt}.

\subsection{Knowledge-Application Gap Analysis}
\label{app:knowledge_gap_case}

\begin{figure}[t]
  \centering
  \includegraphics[width=\columnwidth]{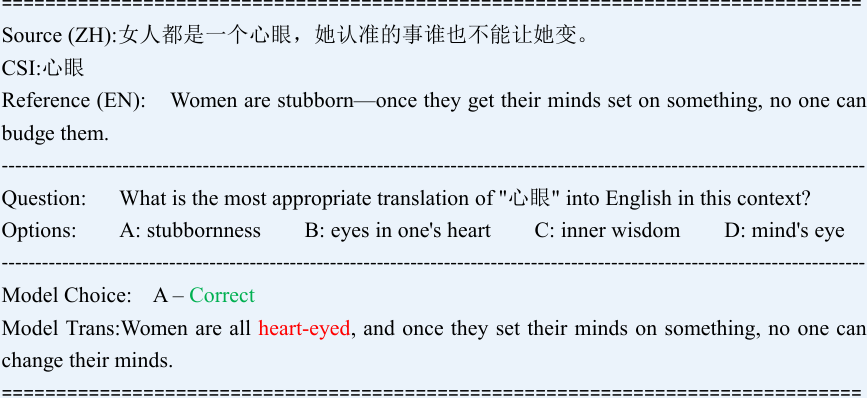}
  \caption{Case Study: Knowledge–Application Gap.}
  \label{fig:gap_example}
\end{figure}

As shown in Figure~\ref{fig:gap_example}, although the model selected the correct option for the CSI, it still failed to produce the correct translation in the final output.

This example shows that even with correct knowledge of a CSI, the model may fail to produce a contextually accurate or culturally faithful translation, highlighting the gap between knowing and applying cultural translation knowledge.

\subsection{Mitigating the Observed Gap}
\label{app:knowledge_gap_mitigation}

To further investigate whether the observed knowledge-application gap can be mitigated, we conduct a preliminary study using a simple two-stage inference strategy.

\paragraph{Stage 1: Knowledge Identification}
The model is first prompted to explicitly identify cultural-specific items in the source sentence, along with their cultural explanations and recommended translations.

\paragraph{Stage 2: Knowledge Application}
The extracted CSI information is then provided back to the model as additional context to guide the final translation.

We conduct experiments on a subset of models and evaluate their performance on CSI-related dimensions. The results are shown in Table~\ref{tab:knowledge_guided_results}.

\begin{table*}[t]
\centering
\small
\resizebox{\textwidth}{!}{
\begin{tabular}{lcccc}
\toprule
Model & Ctx. Acc. (w/ Know.) & Ctx. Acc. (w/o Know.) & Cul. Adapt. (w/ Know.) & Cul. Adapt. (w/o Know.) \\
\midrule
Llama-3-8B-Instruct-262k & 4.007 & 3.210 & 4.298 & 3.623 \\
Qwen2.5-7B-Instruct & 4.533 & 3.811 & 4.771 & 4.060 \\
Qwen2.5-14B-Instruct & 5.012 & 4.323 & 5.226 & 4.653 \\
Qwen2.5-32B-Instruct & 5.206 & 4.490 & 5.366 & 4.748 \\
\bottomrule
\end{tabular}
}
\caption{Performance comparison with and without knowledge-guided two-stage inference.}
\label{tab:knowledge_guided_results}
\end{table*}

Across multiple models, the two-stage strategy consistently improves both contextual accuracy and cultural adaptation scores.

These results suggest that the knowledge-application gap partly stems from failures in activating and utilizing relevant knowledge during generation, rather than the absence of knowledge.

\section{Impact of Reference Translations}
\label{app:eval_case_studies}

\begin{figure*}[t]
  \centering
  \includegraphics[width=\textwidth]{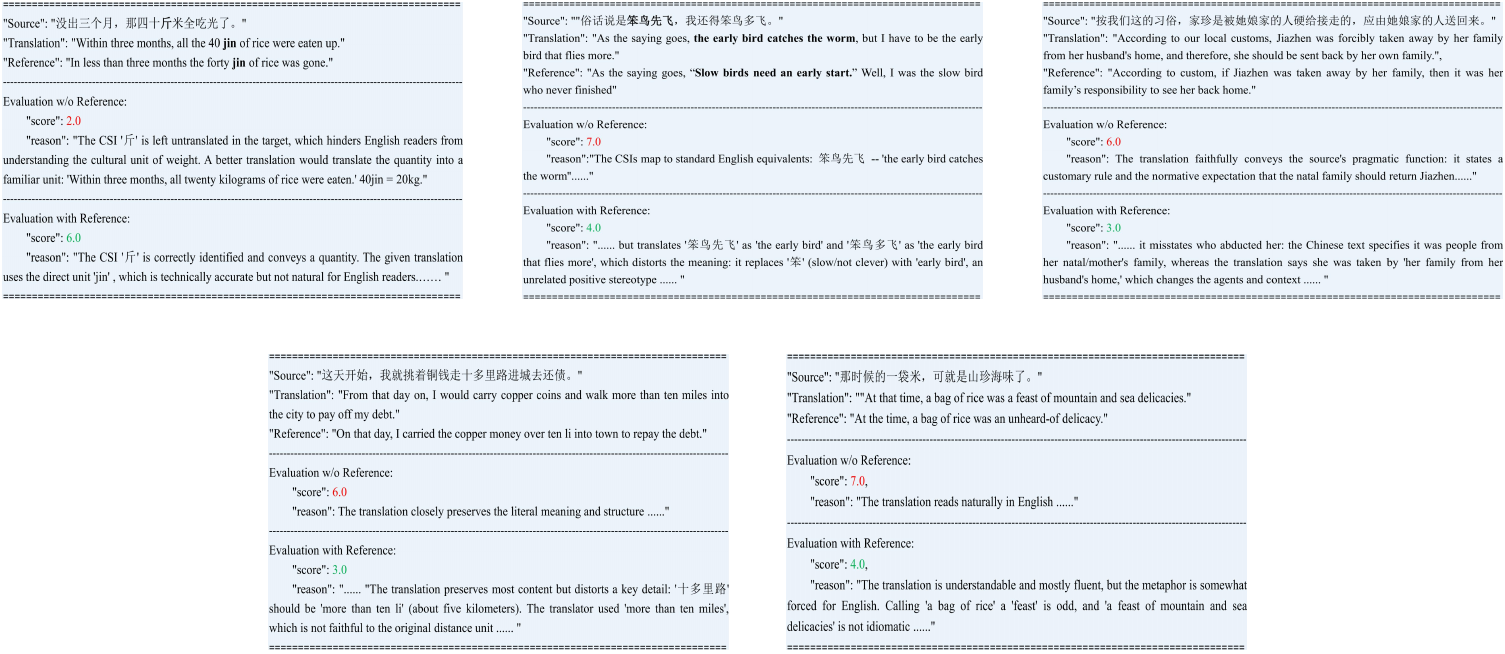}
  \caption{Representative case studies on contextual accuracy, cultural adaptation, functional equivalence, fidelity and naturalness.}
  \label{fig:eval_case}
\end{figure*}

In this appendix, we present detailed case studies to qualitatively analyze the impact of reference translations across different evaluation dimensions. For each dimension, we illustrate a representative case (See Figure~\ref{fig:eval_case}) by comparing the evaluator's reasoning and scores in reference-free and reference-based settings. These examples demonstrate how reference translations support the evaluation of cultural-specific items, facilitate the detection of fine-grained semantic errors, and help calibrate whether the translation style aligns with target-language norms.

% In this appendix, we present detailed case studies to qualitatively analyze the impact of reference translations across different evaluation dimensions. Figures~\ref{fig:eval_case_1} and~\ref{fig:eval_case_2} summarize five representative examples covering contextual accuracy, cultural adaptation, functional equivalence, fidelity, and naturalness. Below, we further discuss each case in detail by comparing the evaluator's reasoning and scores in reference-free and reference-based settings. These examples show that reference translations help validate culturally appropriate translation choices, reveal fine-grained semantic and factual errors, and calibrate whether the translation style aligns with target-language norms.

\paragraph{Contextual Accuracy.}
As shown in the upper-left panel of Figure~\ref{fig:eval_case}, the evaluator initially penalizes the retention of the specific Chinese unit ``jin'' (Score: 2), suggesting a domesticated conversion to ``kilograms'' instead. However, the provision of a reference that also retains ``jin'' validates the accuracy of the target translation, confirming that preserving the source cultural unit is the correct strategy in this context (Score: 6).

\paragraph{Cultural Adaptation.}
As shown in the upper-middle panel of Figure~\ref{fig:eval_case}, the evaluator initially praises the target for replacing the source CSI with the familiar English idiom ``the early bird'' (Score: 7). However, the reference translation reveals that this adaptation distorts the original cultural image of ``slow bird''. The reference enables the evaluator to identify the loss of the specific cultural connotation of humility, resulting in a penalized score (Score: 4).

\paragraph{Functional Equivalence.}
As shown in the upper-right panel of Figure~\ref{fig:eval_case}, without a reference, the evaluator assigns a high score for capturing the custom’s pragmatic function (Score: 6). With the reference, however, it becomes clear that the translation fails to preserve the kinship-specific meaning of niangjia, rendering it simply as her family. From the perspective of functional equivalence, this weakens the cultural norm conveyed in the source text, since the original explicitly marks the natal family as the party both performing the action and bearing the obligation. This loss of relational specificity reduces functional equivalence (Score: 3).

\paragraph{Fidelity.}
As shown in the lower-left panel of Figure~\ref{fig:eval_case}, the evaluator initially assigns a high score, assuming the translation closely preserves the literal meaning and structure of the source (Score: 6). However, the reference translation provides a precise baseline that reveals a critical fine-grained semantic error: the mistranslation of the Chinese distance unit ``li'' as ``miles''. This unit distortion, which significantly alters the factual scale of the narrative, is only identified through direct comparison with the reference, resulting in a more rigorous evaluation (Score: 3).

\paragraph{Naturalness.}
As shown in the lower-right panel of Figure~\ref{fig:eval_case}, the reference translation provides a benchmark favoring a more natural and idiomatic English expression. This comparison reveals that the target's literal translation is actually forced and unidiomatic in the given context, leading to a corrected lower score (Score: 4).

\end{document}